\documentclass{article}

\usepackage{PRIMEarxiv}

\usepackage[utf8]{inputenc} % allow utf-8 input
\usepackage[T1]{fontenc}    % use 8-bit T1 fonts
\usepackage{hyperref}       % hyperlinks
\usepackage{url}            % simple URL typesetting
\usepackage{booktabs}       % professional-quality tables
\usepackage{amsmath, amsfonts}       % blackboard math symbols
\usepackage{dsfont}
\usepackage{nicefrac}       % compact symbols for 1/2, etc.
\usepackage{microtype}      % microtypography
\usepackage{lipsum}
\usepackage{fancyhdr}       % header
\usepackage{graphicx}       % graphics
\graphicspath{{media/}}     % organize your images and other figures under media/ folder
\usepackage{tikz}
\usepackage{xspace}
\usepackage{rotating}
\usepackage[ruled,vlined]{algorithm2e}
\usetikzlibrary{positioning}
\usetikzlibrary{3d} %for including external image
\usepackage[export]{adjustbox}
\usepackage{enumitem}
\usepackage{pgfplots}
\usepackage{pgf}
\pgfplotsset{compat=1.17}
\usepackage{float}
\usepackage{subcaption}
\usepackage{multirow}
\usepackage{makecell}

\usepackage{pbox}

\usepackage{array}
\usepackage{geometry}

\definecolor{mydarkgreen}{RGB}{22, 148, 35}

\newcommand{\revision}[1]{{#1}}
\newcommand{\eg}{\textit{e.g.,}\xspace}
\newcommand{\etc}{\textit{etc.}\xspace}

\newcommand{\quotes}[1]{``#1''}
\newcommand{\imageres}{imgs_low}

\title{TreeSketchNet: From Sketch to 3D Tree Parameters Generation 
}

\author{
  Gilda Manfredi, Nicola Capece, Ugo Erra, and Monica Gruosso \\
Department of Mathematics, Computer Science, and Economics  \\
  University of Basilicata \\
  Potenza, Italy\\
  \texttt{\{gilda.manfredi, nicola.capece, ugo.erra, monica.gruosso\}@unibas.it} \\
}

\begin{document}
\maketitle

\begin{abstract}
3D modeling of non-linear objects from stylized sketches is a challenge even for experts in Computer Graphics (CG). The extrapolation of objects parameters from a stylized sketch is a very complex and cumbersome task. In the present study, we propose a broker system that mediates between the modeler and the 3D modelling software and can transform a stylized sketch of a tree into a complete 3D model. The input sketches do not need to be accurate or detailed, and only need to represent a rudimentary outline of the tree that the modeler wishes to 3D-model. Our approach is based on a well-defined Deep Neural Network (DNN) architecture, we called TreeSketchNet (TSN), based on convolutions and able to generate Weber and Penn~\cite{10.1145/218380.218427} parameters that can be interpreted by the modelling software to generate a 3D model of a tree starting from a simple sketch. The training dataset consists of Synthetically-Generated \revision{(SG)} sketches that are associated with Weber-Penn parameters generated by a dedicated Blender modelling software add-on. The accuracy of the proposed method is demonstrated by testing the TSN with both synthetic and hand-made sketches. Finally, we provide a qualitative analysis of our results, by evaluating the coherence of the predicted parameters with several distinguishing features.

\end{abstract}

% keywords can be removed

\section{Introduction}\label{sec:intro}
%Manual mesh modeling of objects characterized by non-linear complex 3D structures represents a challenge also for very CG experts. Typically, these objects such as trees are designed through procedural modelling~\cite{Xie2015}, which allows users to operate on specific parameters~\cite{deussen2005digital, huang2017} that characterize them, without the direct editing of the their geometries. However, due to the rules complexity that affect these objects, the set of their parameters is usually very large and also non-linear. Consequently, their extraction represents a very complex operation. Drawing a target 3D object is often more easy and intuitive with respect to the manual reconstruction or the parametric modeling of its geometry. 
  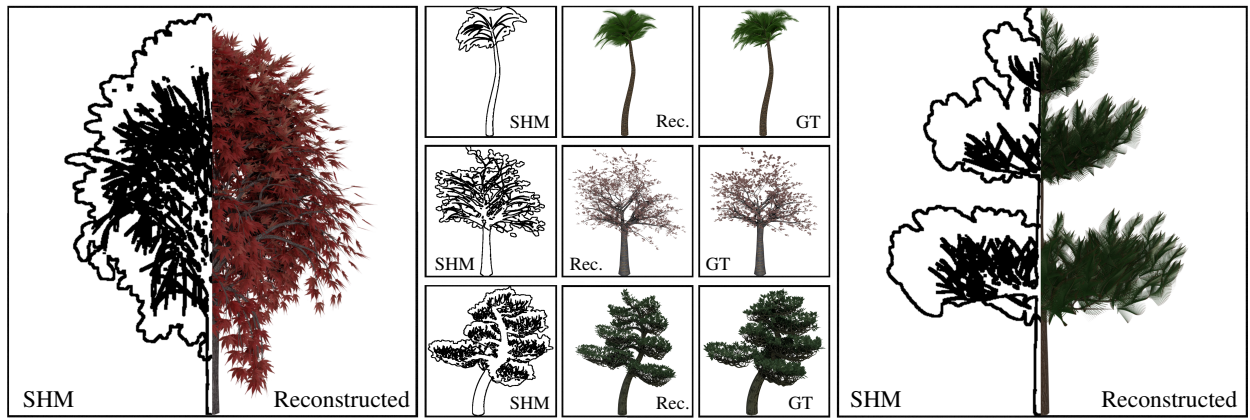
\begin{figure*}
    \begin{tikzpicture}
        \node at (0,0) {\includegraphics[width=\textwidth]{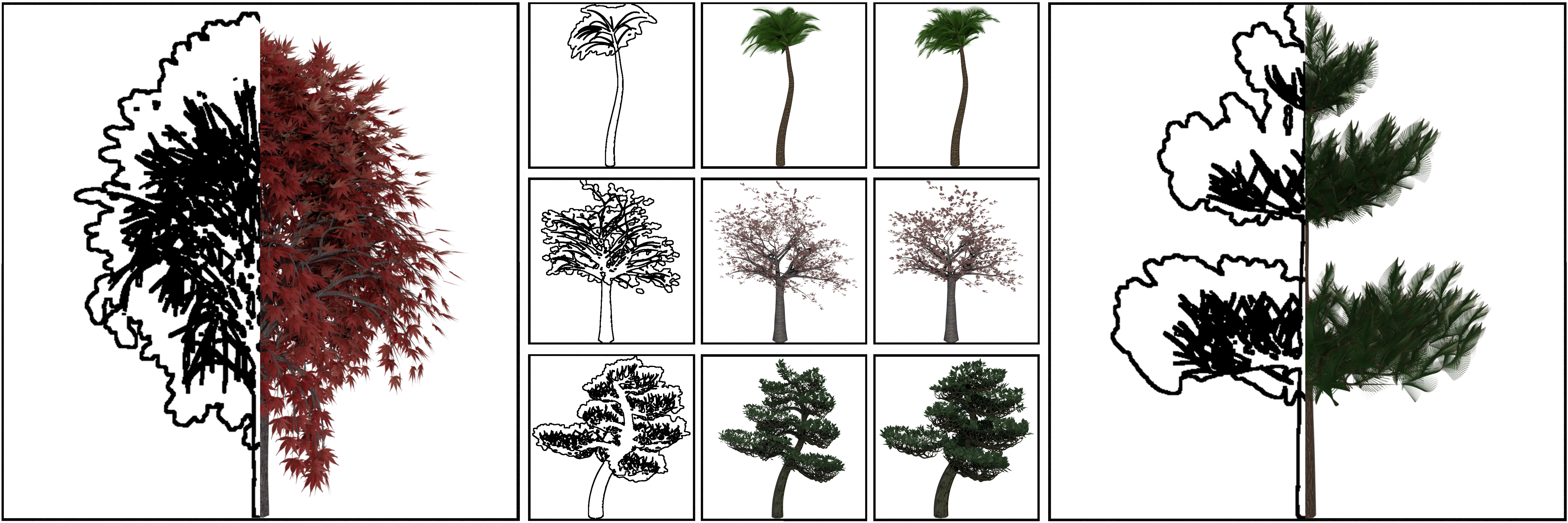}};
        \node at (-7.7,-2.5) {\small SHM};
        \node at (-3.8,-2.5) {\small Reconstructed};  
        \node at (-1.3,1.2) {\scriptsize SHM};
        \node at (0.6,1.2) {\scriptsize Rec.};
        \node at (2.4,1.2) {\scriptsize GT};
        \node at (-2.3,-0.7) {\scriptsize SHM};
        \node at (-0.55,-0.7) {\scriptsize Rec.};
        \node at (1.2,-0.7) {\scriptsize GT};
        \node at (-1.3,-2.55) {\scriptsize SHM};
        \node at (0.6,-2.55) {\scriptsize Rec.};
        \node at (2.3,-2.55) {\scriptsize GT};
        \node at (3.3,-2.5) {\small SHM};
        \node at (7.2,-2.5) {\small Reconstructed};  
    \end{tikzpicture}
    % \caption{Reconstructed 3D trees are shown as images captured from different camera angles. The image on the left shows a front-view maple tree reconstruction starting from a synthetic hand-made (SHM) input. The top row of the middle column shows input, reconstruction and Ground Truth (GT) for the left-camera view of a palm, starting from the SHM input. The middle row of the same column shows a right-camera view of a cherry tree, reconstructed from a mean adaptive filter (MAF) sketch. The bottom row of the middle column shows a right-camera view reconstruction of a bonsai, starting from SHM input. Finally, the image in the right-hand column shows a right-camera view reconstruction of a pine tree, starting from a MAF sketch.\fixme{da togliere MAF}}
        \caption{Reconstructed 3D trees are shown as images captured from different camera angles. The image on the left shows a front-view maple tree reconstruction starting from a synthetically-generated \revision{(SG)} input. The top row of the middle column shows input, reconstruction and Ground Truth (GT) for the left-camera view of a palm, starting from the \revision{SG} input. The middle row of the same column shows a front-camera view of a cherry tree, starting from the \revision{SG} input. The bottom row of the middle column shows a right-camera view reconstruction of a bonsai, starting from \revision{SG} input. Finally, the image in the right-hand column shows a left-camera view reconstruction of a pine tree, starting from the \revision{SG} input.}

    \label{teaser}
  \end{figure*}
  
Manual mesh modeling of objects that are characterized by non-linear complex 3D structures remains a challenge even for experts in CG. Typically, objects such as trees are designed using procedural modelling~\cite{Xie2015}, which allows users to manipulate the specific parameters~\cite{deussen2005digital, huang2017,nishida2016} that characterize them, avoiding direct editing of their geometries. However, the complexity of the rules that affect these objects means that the set of parameters is not only very large, but also non-linear. Drawing a target 3D object is often easier and more intuitive than a manual reconstruction, or parametric modeling of its geometry. 
There are several image-based approaches to reconstructing real objects as 3D models, \eg those based on photogrammetry~\cite{Remondino2009, Gatziolis2015} and new technologies such as Depth Cameras~\cite{Nguyen2018, Zhou2014}, LiDAR~\cite{HU20171, tachella2019real}, Laser Scanning~\cite{lu2006reconstruction} \etc Although there are several methods that use images to \revision{predict directly a coarse} 3D mesh~\cite{pix2surf_2020, wang2018pixel2mesh, pan2019deep, Kanazawa_2018_ECCV} many produce inaccurate results, with smoothed or very high poly meshes, wrongly-closed holes, \revision{non-manifold edges and vertices, and isolated vertices}. \revision{To overcome these problems several approaches aim to automatize procedural modelling using images for predicting parameters~\cite{garmentdesign_Wang_SA18,2014Smelik,liu2021deep}.}
\revision{Along the same lines as previous approaches}, we introduce a broker system that sits between 3D modelers and the specialized 3D modelling software used to build 3D trees. Users only need to provide a hand-made (HM) sketch as input; then the broker provides parameters that can be read by the 3D modelling software, and used to build a 3D tree that is as similar as possible to the sketch. Furthermore, our broker can provide the appropriate texture used to render the 3D tree. The key element in our broker is a convolution based neural network that is trained using supervised learning paradigms, and is able to learn parameter mappings and analyze the sketch based on a set of training data. As collecting a sufficiently large number of HM sketches is a very expensive operation, we have developed a specific Blender add-on. This add-on, called the Render Tree (RT) is based on the existing Blender Sapling Tree Gen~\footnote{\url{https://docs.blender.org/manual/en/latest/addons/add_curve/sapling.html}} \revision{ which adopts the Weber and Penn~\cite{10.1145/218380.218427} procedural modelling methodology to create 3D trees from a set of parameters}. The benefit of this approach is that it makes it possible to create a consistent training dataset of synthetically-generated \revision{(SG)} sketches, as reported in Figure~\ref{teaser} (\eg the maple tree). \revision{Although Weber and Penn is a fairly old model, its stability combined with the wide availability of implementation code make it a method still widely used for the parametric generation of 3D models of plants and more \cite{rutzinger2010detection,FRISKEN2022102076,RICHTER2022510,Delalieux2014}.}
Our experiment evaluated five tree \revision{species}: maple, pine, bonsai, palm, and cherry. Four camera views were gathered for each: front, back, left and right. In the final step, we generated 250 randomly-controlled versions of each \revision{species} of tree, by varying the input parameters to the RT Blender add-on.
This resulted in $5000$ \revision{SG}.
It should be noted that our dataset could be extended to include other \revision{species} of trees.

The main contributions of our paper can be summarized as:
\begin{enumerate}
    \item An approach to quickly generate a training dataset consisting of synthetic and realistic sketches of 3D trees, starting from randomly-controlled Weber-Penn parameters.
    \item A specific DNN architecture with multiple outputs, based on the set of parameter values, and the training and testing process.
    \item The RT Blender add-on and the trained DNN are available online, and can be freely used for similar predictions: \url{https://github.com/Unibas3D/TreeSketchNet}.
\end{enumerate}
\section{Related Work}\label{sec:related}
\revision{In past years, the best way to reconstruct an object in 3D relied upon the manual skills of a human modeler, especially for objects with very complex geometries, such as trees. Later, the procedural modeling technique began to take hold. This technique facilitated the modeling phase by allowing the user to manipulate the values of a parameters' series to which the object's geometry was bound. However, procedural modeling still required a user training phase to understand how the object parameters work. In recent years, thanks to the rapid growth of artificial intelligence technologies, 3D modeling has broken new ground, and has become accessible to non-expert modelers. In this section, we report the State-Of-The-Art (SOTA) in the domain of 3D mesh generation, starting from sketches or images.}
%\fixme{Until recently, the best way to reconstruct an object in 3D relied upon the manual skills of a human modeler, and modeling tools, such as well-known professional software suites that include Autodesk\textregistered 3D Studio Max and Maya, Cinema 4D (MAXON\textregistered), Houdini (SideFX\textregistered), and Blender. However, in recent years, and thanks to the rapid growth of artificial intelligence technologies, 3D modeling has broken new ground, and has become accessible to non-expert modelers. In this section, we report the state-of-the art in the domain of 3D mesh generation, starting from sketches or images.}

\subsection{3D mesh generation}\label{sub:3d_mesh}
The use of new sensors such as LiDAR or Depth Cameras is very familiar to anyone who owns a latest-generation smartphone. In this context,~\cite{cheng2013integration} propose an approach for the automatic reconstruction of 3D roofs, using airborne LiDAR data and optical multi-view aerial imagery. ~\cite{tachella2019real} propose a computational framework for real-time 3D scene reconstruction using single-photon data. The latter authors reconstructed complex outdoors scenes by acquiring LiDAR data in broad daylight from distances up to 320 meters. A LiDAR-scanned point cloud has also been used to model real-world trees, and~\cite{HU20171} present a method to model plausible trees in fine-grained detail from airborne LiDAR point clouds. The tree model is reconstructed by first segmenting a single tree point cloud, and then integrating trunk points that complete the cloud, using a connected graph nearest neighbor search for each point. The constructed branch skeletons are developed using a bottom-up greedy algorithm, then arranging the leaves. There are also other interesting approaches that uses the tree point cloud, obtained by scanning real trees, as an input for a procedural modeling algorithm (\cite{guo2020RPPM}) or for a DNN (\cite{liu2021}) to reconstruct a realistic 3D geometry of a tree.
Many other studies used a depth camera sensor. Examples include~\cite{Nguyen2018}, who propose a 3D reconstruction approach using only one depth camera and two mirrors, and \cite{Haque_2014_CVPR, Izadi2011} who use a Kinect depth camera for 3D reconstruction.
Laser scanning is another method that is used for 3D reconstruction~\cite{lu2006reconstruction}. For example, ~\cite{rutzinger2010detection} present an automated workflow that begins with tree detection from mobile laser scanner point clouds and ends with single tree modelling. The aim is to model single trees for visualization purposes in 3D city models. 

\revision{~\cite{Wang-2018} proposed a data-driven mechanism to automatically create 3D trees starting from existing ones through structure and geometrical blending. The initial models can be created using existing 3D modeling tools or through internet repositories. ~\cite{Quigley2021} tried to reconstruct trees with geometrical and topological accuracy for physical simulation starting from RGB images to create point clouds, textured meshes and skinned cylindrical articulated rigid body models.}

\revision{
With respect to these approaches, we do not use ad-hoc devices such as the laser scanner or the LiDAR for the creation of the input data. Indeed, these instruments often suffer from problems related to the real objects to be scanned, due \eg to high solar angles or huge reflections since the laser pulses depend on the reflection principle. One of the main differences concerns the amount of information that feeds our system. In fact, we use extremely raw data (sketches) that contain less information than existing multiple RGB images, point clouds, or existing polygonal meshes.}

More generally, as noted in~\cite{HU20171}, 3D tree modelling techniques can be classified into procedural~\cite{stava2014inverse}, sketch~\cite{Liu2010, okabe2006interactive}, and image-based~\cite{Tan2007, kim2014single} methods.

\revision{Approaches for generating 3D trees~\cite{Palubicki2009} are often generalized to plant simulations and vegetation modeling. ~\cite{hadrich2021} proposed a method to capture the combustion process of plants. One of their contribution is the wildfires simulation on more of $100K$ individual plants represented with detailed geometry. Similarly~\cite{palubicki2022} proposed a method to simulate ectoclimates to model tree growth interactively based on temperature, light and gradients of water. 
Plant growth simulation and visualization was faced also by~\cite{golla2020temporal} which proposed a semantic segmentation-based method to allow temporal upsampling of acquired point cloud sequences of growth process including topological changes.
}

\revision{Although these methods provided impressive results in terms of quality and correctness, some of them are based on self-organizing process to generate consistent trees and plants~\cite{Makowski2019}. Differently, our work considers SG sketches to create  training and validation datasets, and HM for the testing purposes.}

\subsection{From images to 3D reconstruction}\label{sub:picTo3D}
Digital photogrammetry using RGB images is one of the most popular techniques used to reconstruct 3D models. However, software tools such as Reality Capture require the acquisition of a considerable number of images \revision{for 3D reconstruction of a single real object}, with a well-defined methodology~\cite{kingsland2020}. Several extensive learning-based studies have tried to generate 3D objects starting from simple, single RGB views ~\cite{pix2surf_2020, yang2018learning, hane2017hierarchical, 8703434, zou20173d}. \cite{NIPS2019_8340} propose a deep implicit surface network to generate 3D meshes from 2D images, by combining local and global image features to improve the accuracy of the distance field prediction. \cite{tatarchenko2017octree} propose a volumetric 3D generating network, based on a convolutional decoder. The latter approach predicts the octree structure, and individual cell occupancy values provide high-resolution output even with limited memory. Furthermore, it can generate shapes from a single image, along with a high-level representation of objects and the overall scene. The generation of point clouds using deep learning has been widely explored. For example,\cite{fan2017point} investigate generative networks for 3D geometry based on a point cloud representation. The main focus of the latter work is a well-designed pipeline that is used to infer point positions in the 3D frame from the input image, taking into account the viewpoint. \cite{niu_cvpr18} propose a convolutional-recursive autoencoder architecture to retrieve cuboid, connectivity, and symmetry aspects of objects using a single 3D image. The \revision{encoder} is trained on a shape contour estimation task and a \revision{decoder focuses on the structure features by parsing the original image and the network}, and recursively \revision{decoding} the cuboid structure.  \cite{xie2019pix2vox} investigate a deep learning approach based on single and multiple views to reconstruct voxel 3D representations. The authors propose an encoder-decoder-based framework called Pix2Vox that can be used for 3D reconstruction from real-world and synthetic images.

\revision{Most of the mentioned 3D reconstruction methods based on images and deep learning concern 3D shapes based on voxels or point clouds. They are limited to reconstructing simple and smooth shapes, such as chairs, furniture or cars. The main difference from our approach concerns the object to be reconstructed, that is the trees, which have more complex and detailed tiny structures, such as branches and leaves. In this context, direct prediction of 3D tree models is much more difficult. Furthermore, these methods are based on single or multiple RGB images which provide more information respect to sketches.}

%One of the recent contribution is provided by \cite{LIU2021101115}, that use a cGAN to extrapolate the 3D silouhette and the skeleton of a tree, given a single tree image and some 2D strokes drawn by the user. 

%We modified the last paragraph as follow:
\revision{Like us, \cite{LIU2021101115} used a DNN to procedurally model trees in 3D. Indeed, they used a conditional Generative Adversarial Network (cGAN) to extrapolate two depth images which are useful to reconstruct the trees silhouette and skeleton as 3D point clouds using the edges and strokes obtained from an input RGB image. Finally, they created a final 3D tree model starting from the point clouds using a procedural modelling approach based on self-organization concept~\cite{Palubicki2009}. Although this approach seems similar to ours, it differs substantially in terms of methodology, input processing and final results. 
Indeed, their input consist of RGB images from which the tree edges was extracted using~\cite{canny1986} and the skeleton of the tree was drawn by the user. Both images were used to feed a cGAN and train it to create the previous mentioned depth images casting the problem into an image-to-image translation task. Differently, we used directly the sketches to feed a DNN and predict the parameters without further user interventions, keeping low the information used. In this way, our approach is more extensible with other type of sketches and trees species and it is usable without having real tree pictures.}

%and the decoder focuses on the characteristics of the structure parsing the network and the original image

\subsection{From sketches to 3D reconstruction}\label{sub:picToParameters}
As reported in~\cite{ding2016survey}, the use of sketches to reconstruct a 3D model is intuitive for a human being. In this context, deep learning approaches can be helpful in generating a 3D mesh using minimal data, such as a sketch, as input information. \cite{guillard2021sketch2mesh} provide an encoder-decoder architecture to translate a 2D sketch into a 3D mesh. The method uses latent parameters to represent and refine a 3D mesh that can match the external contours of the sketch. 
\cite{delanoy20183d} propose a data-driven \revision{learning} approach that can reconstruct 3D shapes from one or more drawings. This Convolutional Neural Network(CNN) based method predicts voxel grid occupancy from a line drawing, and outputs an initial 3D reconstruction, while users can complete the drawing of the desired shape. Unsupervised learning \revision{methods are} used \revision{also} for 3D object modeling. \cite{Wang2018} propose a learning paradigm to reconstruct 3D objects from HM sketches that does not rely on labeled HM sketch data during training. Their paradigm takes advantage of adaptation network training, notably autoencoder and adversarial loss, and combines an unpaired 2D rendered image and an HM sketch in a shared latent vector space. In a second step, nearest neighbors are retrieved from the embedded latent space, and each sketch in the training set is used in a 3D Generative Adversarial Network. An interesting approach that is not based on deep learning is proposed in~\cite{li2017}. The latter study presents a tool that can model complex free-form shapes by sketching sparse 2D strokes. The proposed framework combines multi-view inputs to model a complex shape that can be occluded. 
\revision{With respect to our approach most of these methods fails in terms of faith to the GT. Indeed, their results shown unbalanced predicted meshes also for simple structure such as chairs and cars. Furthermore, they have problems with a thin and layered structure like those of the tree trunk and branches of which no test example is also shown. Some of these approaches also require continuous user intervention to improve the faith of the prediction. Differently, we used only a single sketch as input of our pipeline and the prediction is reliable with respect to the thin and layered structures to be reconstructed.}

\subsection{Procedural modeling} 
Rule-based methods \cite{ebert2003texturing, schwarz2015} can also take advantage of deep learning, and can overcome the need for user intervention in the manipulation of a large number of parameters in rule sets.
% , especially for a non-linear complex object such as a tree. 
Thus, \cite{huang2017} present a HM sketch approach to automatically compute a set of procedural model parameters that can be used to generate 2D/3D shapes that resemble the initial input HM sketches. An interesting aspect of the latter study is that it focuses on three procedural modeling rule sets, namely 3D containers (\eg vases), 3D jewelry, and 2D trees. The authors did not consider 3D tree shape generation due to the high complexity of the task. However, several other approaches can overcome this problem by controlling the output of procedural modeling algorithms. Methods include \cite{haubenwallner2017shapegenetics}, who use a genetics-based algorithm, and \cite{stava2014inverse,ICTree}, which is based on a Monte Carlo Markov Chain. \revision{Unlike our approach, these methods require a parameter learning phase by the users to understand their functionalities. To further facilitate the modeler work,  \cite{Longay2012} developed a procedural modeling-based application for generating 3D tree models with a tablet. The modeler can define the profile of the tree and the direction of the branch's growth with a brush. Many other details of the tree need to be defined by working on the parameters related to the procedural modeling algorithms used. Although this approach is more intuitive than the previous ones, the presence of parameters implies a preliminary study by the user that is not necessary with our methodology.}
\revision{An important rule-based method is L-system, proposed by~\cite{LINDENMAYER1968280}, which is widely used for plant generation. Indeed, several algorithms based on the L-system model were proposed by~\cite{prusinkiewicz2012algorithmic, stava2010,liu2021}. In particular,} an interesting inverse procedural modeling approach \revision{was} introduced by \cite{guo2020}, that uses deep learning to detect an L-system from an input image \revision{with branching structures. Obviously, because this method can predict only parameters for generating linear branching structures, both the input image and the resulting grammar have no information about the crown of a tree}. \revision{\cite{li2021} provided an approach that can overcome this limitation. They} used a combination of deep learning and procedural modeling algorithms to extract a 3D tree model from a single \revision{photograph of a tree}. This method is based on three DNNs that, starting form an input photograph of a tree, mask out the tree, identify the tree \revision{species}, and extract the tree Radial Bounding Volume (RBV), respectively. The RBV is then used as an input for a procedural modeling algorithm that generates the 3D model of the tree. \revision{Hence it follows that this neural architecture is much more complex than our. Furthermore, their input RGB images have much more 3D hint than sketches, such as shadows and information about textures.}
\revision{Interesting are also newer approaches for generating 3D models from HM sketches. \cite{garmentdesign_Wang_SA18} use a multiple encoder-decoder network architecture to create a final draped 3D garment from a single sketch. The complexity of the network architecture is strongly linked to the necessity of predicting not only the parameters of the 2D garment pattern but also the body shape parameters, which have to be related to have a final 3D garment that resembles the input sketch. The idea of using multiple neural network to predict different set of parameters that have to be related is followed by \cite{unlu2022interactive}. Indeed, they use two neural networks to predict 2D silhouettes and 2D joints from an input mannequin sketch. These intermediate representation are then used to generate the mannequin 3D model. Unlike the two previous approaches, ours is able to generate 3D tree models with a single neural network, without using intermediate representation, although tree geometry is linked to complex and non-linear set of rules.}

\section{Generation of 3D tree parameters}\label{sec:parameters}
In this section, we discuss the details of our approach, notably: \textit{(i)} the dataset creation pipeline; \textit{(ii)} the configuration of the TSN, and;\textit{(iii)} the training procedure. The RT Blender add-on is a key element in our pipeline as it quickly allows the generation of the training dataset and the visualization of the 3D tree mesh. The latter is generated from Weber-Penn \cite{10.1145/218380.218427} parameters predicted from the TSN using \revision{SG} sketches created by the add-on as input.
\revision{To define our parameters, we adopted the current standard notation in CG vegetation simulation as used in~\cite{stava2014inverse,YANG201963,TDFGS_tang,PRADAL20091}.}

\subsection{The dataset creation pipeline}\label{sub:dataset_creation}
One of the weaknesses of supervised learning methods is finding a way to arrange a large set of labelled data, which also have to be well-structured, to obtain the expected results \cite{fredriksson2020, mohri2018foundations, yang2021survey}. Our RT Blender add-on overcomes this drawback by creating, for each \revision{species}, $250$ 3D tree meshes and storing parameters in a dedicated dictionary file for each tree (see Details in our GitHub website~\cite{treesketchnet_params_details}). 
We divide the Weber-Penn parameters into two subsets: fixed and unfixed. Fixed parameters take the same values for the same tree \revision{species}, and unfixed parameters take randomly-controlled values that vary the shape and detailed visual features of the tree. 
The TSN could be trained to understand the tree \revision{species} shown in the input sketch by adding a classification branch. However, this would increase its complexity and, consequently, degrade prediction performance. Moreover, adding a classification branch is unnecessary because it is possible to identify the tree \revision{species} associated with the input sketch using fixed parameters. This task was implemented as a robust algorithm~\ref{alg:identify}, and is discussed in Section~\ref{sec:results}.
Starting from the pre-existing Sapling Tree Generator Blender plugin, we manually define a dictionary of fixed and unfixed parameters for each of the $5$ tree \revision{species}. Here, the aim is to obtain consistent 3D tree meshes, in terms of visual features. The tree \revision{species} were chosen in order to introduce as much differentiation as possible with respect to their shape and visual features. The consistent mesh dictionaries produced by our RT plugin were used as a starting point to generate the other 3D meshes, and their respective parameter dictionaries. Unfixed parameters were randomly varied with respect to their order of magnitude, as shown in Table~\ref{tab:varying_parameters}. 

% \begin{table}[ht!]
% \caption{Unfixed parameters and their range (minimum and maximum). The Sign parameter is binary.}
%  \begin{tabular}{c c c} 
%  \hline
%  Unfixed Parameters & Min Values & Max Values \\
%  \hline
%  Sign (binary) & -1 & 1 \\
%  Tree Forks Number & 0 & fixed parameter \\ 
%  Rotation Last Angle & -360 & 360 \\
%  Curvature & -360 & 360 \\
%  Back Curvature & -360 & 360 \\
%  Curvature Variation & -360 & 360 \\
%  Split Angle & -360 & 360 \\
%  Split Angle Variation & -360 & 360\\
%  Rotate Angle & -360 & 360 \\
%  Rotate Angle Variation & -360 & 360 \\
%  Leaf Rotation & -360 & 360 \\
%  Leaf Down Angle Variation & -360 & 360 \\ 
%  Leaf Scale & 0 & $\infty$ \\ 
%  Leaf Scale Variation & 0 & 1 \\ 
%  Down Angle Variation & -360 & 360 \\ 
%  Branch Rings & 0 & $\infty$ \\
%  \hline
% \end{tabular}
% \label{tab:varying_parameters}
% \end{table}

\begin{table}[ht!]
\centering
\caption{Unfixed parameters and their range (minimum and maximum). The Sign parameter is binary.}
 \begin{tabular}{l c c} 
 \hline
 Unfixed Parameters & Min Values & Max Values \\
 \hline
 Sign (binary) & -1 & 1 \\
 \revision{Tree Forks Number} & 0 & fixed parameter \\ 
 \revision{Parent Branch Roll Angle} & -360 & 360 \\
 \revision{First Half Internodes Branching Angle} & -360 & 360 \\
 \revision{Second Half Internodes Branching Angle} & -360 & 360 \\
 \revision{Internode Branching Angle Variance} & -360 & 360 \\
 \revision{Sibling Angle} & -360 & 360 \\
 \revision{Sibling Angle Variance} & -360 & 360\\
 \revision{Branch Roll Angle} & -360 & 360 \\
 \revision{Branch Roll Angle Variance} & -360 & 360 \\
 \revision{Leaf Roll Angle} & -360 & 360 \\
 \revision{Leaf Angle Variance} & -360 & 360 \\ 
 \revision{Leaf Scaling Factor} & 0 & $\infty$ \\ 
 \revision{Leaf Scaling Factor Variance} & 0 & 1 \\ 
 \revision{Parent  Branch Angle Variance} & -360 & 360 \\ 
 \revision{Number of Branch Whorls} & 0 & $\infty$ \\
 \hline
\end{tabular}
\label{tab:varying_parameters}
\end{table}

% \begin{table}[ht!]
% \caption{Unfixed parameters and their range (minimum and maximum). The Sign parameter is binary.}
%  \begin{tabular}{l c l c} 
%  \hline
%  Unfixed Parameters & $[min_{v}, max_{v}]$ 
%  & Unfixed Parameters & $[min_{v}, max_{v}]$\\
%  \hline
%  Sign (binary) & $[-1, 1]$ &
%  \revision{Tree Forks Number} & $[0, p_{fixed}]$ \\ 
%  \revision{Parent Branch Roll Angle} & $[-360, 360]$ &
%  \revision{First Half Internodes Branching Angle} & $[-360, 360]$ \\
%  \revision{Second Half Internodes Branching Angle} & $[-360, 360]$ &
%  \revision{Internode Branching Angle Variance} & $[-360, 360]$ \\
%  \revision{Sibling Angle} & $[-360, 360]$ &
%  \revision{Sibling Angle Variance} & $[-360, 360]$\\
%  \revision{Branch Roll Angle} & $[-360, 360]$ &
%  \revision{Branch Roll Angle Variance} & $[-360, 360]$ \\
%  \revision{Leaf Roll Angle} & $[-360, 360]$ &
%  \revision{Leaf Angle Variance} & $[-360, 360]$ \\ 
%  \revision{Leaf Scaling Factor} & $[0, \infty]$ & 
%  \revision{Leaf Scaling Factor Variance} & $[0, 1]$ \\ 
%  \revision{Parent  Branch Angle Variance} & $[-360, 360]$ & 
%  \revision{Number of Branch Whorls} & $[0, \infty]$ \\
%  \hline
% \end{tabular}
% \label{tab:varying_parameters}
% \end{table}

In this way, we obtain $250$ dictionaries for each tree \revision{species}, which are used individually in our RT plugin to generate the corresponding Blender Curve ~\cite{hughes2014computer} object. This object is converted into two meshes: the skeleton (trunk and branches) of the tree, excluding the foliage, and the foliage alone. 

\begin{figure}[ht!]
    \centering
    \begin{tikzpicture}
     \node at (0,0) {\includegraphics[width=0.9\textwidth]{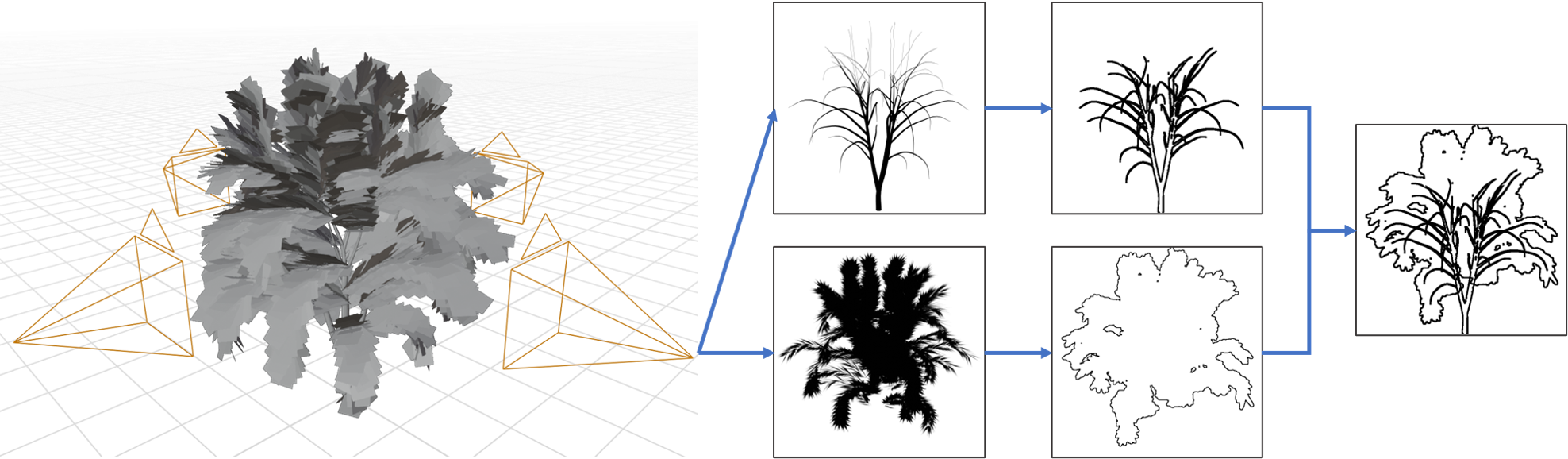}};
     \node at (0.1,0.3) {\textbf{a}};
		 \node at (2.7,0.3) {\textbf{c}};
% 		 \node at (0.1,-1.8) {\textbf{b}};
% 		 \node at (2.6,-1.8) {\textbf{d}};
        \node at (0.1,-2.0) {\textbf{b}};
		 \node at (2.7,-2.0) {\textbf{d}};
		 \node at (5.6,-0.85) {\textbf{e}};
    \end{tikzpicture}
    \caption{Generation of the \revision{SG} sketch. For each camera view, the tree’s skeleton and foliage are individually rendered after applying the black and white materials, and after a set of pre-processing operations specific to the two \revision{species} of meshes. In the final step, these renderings are mixed to obtain the final \revision{SG} sketch.}
    \label{fig:sketchPipeline}
\end{figure}
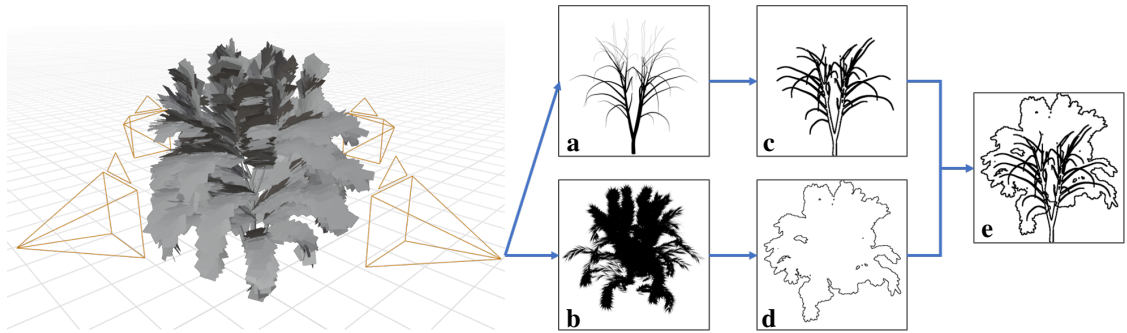

%The obtained meshes were loaded in a well-designed Blender scene integrated with $4$ cameras corresponding to the $4$ tree views: front, back, left and right as shown in Figures~\ref{fig:sketchPipeline} and \ref{fig:MAFPipeline}. In addition, our scene contains a directional light without the shadows to better illuminate the tree. The default ambient occlusion was also enabled to increase the rendering realism. 
The obtained meshes are loaded into a well-designed Blender scene integrated with $4$ cameras corresponding to the $4$ tree views: front, back, left, and right as shown in Figures~\ref{fig:sketchPipeline}. \revision{These views have been chosen because they are easy to recognize and because is also easier to make visual comparisons with the 3D model predicted by the TSN. Furthermore, each of the $4$ views is completely different from the others: a characteristic that can reduce the TSN overfitting, granting a great level of generalization, as shown in Figure \ref{fig:images_different_views}. We also tried to train our TSN with other views, but we didn’t get significantly improvement.} In addition, our scene contains a directional light, without shadows, to better-illuminate the tree. The global scene illumination consist of Blender path tracing and ambient occlusion to increase rendering realism. 
\revision{After setting the scene, the SG sketch for each view have to be rendered. The first step of the SG sketch generation pipeline consists in adding}
% The first step in generating the \revision{SG} sketch is to add 
two black materials to the tree, one for the trunk and one for the foliage, as shown in Figures~\ref{fig:sketchPipeline}a and~\ref{fig:sketchPipeline}b. The first is a simple diffuse material with zero roughness. The second consists of a simple leaf-shaped black texture, based on the tree \revision{species}.

% \begin{figure}[ht!]
%     %\centering
%     \begin{tikzpicture}
%      \node at (0,0) {\includegraphics[width=0.45\textwidth]{samples/\imageres/sketches_pipeline/syntetic_handMade_pipeline.png}};
%      \node at (-2.9,-0.7) {\textbf{a}};
% 		 \node at (2.9,-0.65) {\textbf{b}};
% 		 \node at (-2.9,-3.1) {\textbf{c}};
% 		 \node at (2.9,-3.05) {\textbf{d}};
% 		 \node at (-0.8,-4.35) {\textbf{e}};
%     \end{tikzpicture}
%     \caption{Generation of the \revision{SG} sketch. For each camera view, the tree’s skeleton and foliage are individually rendered after applying the black and white materials, and after a set of pre-processing operations specific to the two \revision{species} of meshes. In the final step, these renderings are mixed to obtain the final \revision{SG} sketch.}
%     \label{fig:sketchPipeline}
% \end{figure}

The next step is to render the skeleton and the foliage meshes of the tree individually, with the previously-defined black materials.
To obtain the skeleton edges, we define several steps as shown in Figure~\ref{fig:sketchPipeline}c: \textit{(i)} the first step is to slightly thin the skeleton mesh to remove the thinnest branches; \textit{(ii)} the second step consists of the following chain of compositing operations ~\cite{porter1984}:    
\begin{enumerate}[label=(\alph*)]
\item remove the background using a $0.5$ threshold value;
\item apply the Intel\textregistered~Open Image Denoise-based filter to delete the residual background artifact;
\item apply the Sobel filter~\cite{996} to trace rough outlines;
\item apply a color ramp that sets pixel values less than $0.9$ to $0$ and the remainder as $1$, to highlight all branch edges;
\item apply the inversion color filter to obtain a negative, with white background and dark edges;
\item apply the erosion morphological operation to delete scattered points and produce the final skeleton sketch, as can be seen in Figure~\ref{fig:sketchPipeline}c.
\end{enumerate}
Similarly, to obtain the foliage edges (see Figure~\ref{fig:sketchPipeline}d), we use a simpler compositing chain of operations consisting of:
\begin{enumerate}[label=(\alph*)]
\item apply the background removal filter using a $0.9$ threshold value;
\item apply the strong Gaussian filter to uniformalise and emphasize the shape of the foliage;
\item apply a color ramp that sets values less than $0.01$ to $0$ pixel and the remainder $1$, to include blurred foliage edges;
\item apply the Sobel filter to trace the edges;
\item apply the same inversion color filter used for the skeleton sketch to obtain the foliage sketch, as shown in Figure~\ref{fig:sketchPipeline}d. 
\end{enumerate}
The final synthetic sketch is obtained by multiplying the skeleton and foliage images (the mixed sketch), as shown in Figure~\ref{fig:sketchPipeline}e. \revision{The drawing style chosen for the SG sketch aims to resemble a hand-made sketch drawn by a typical user, as shown in \cite{sketchyscene,HDHSO_Eitz_2012}.}

\begin{figure}[ht!]
 \centering
\begin{tabular}{c c c c c c} 
 & Maple Tree & Palm Tree & Pine Tree & Cherry Tree & Bonsai Tree\\
 \begin{turn}{90} \quad\quad GT\end{turn} 
& \frame{\includegraphics[width=0.135\textwidth]{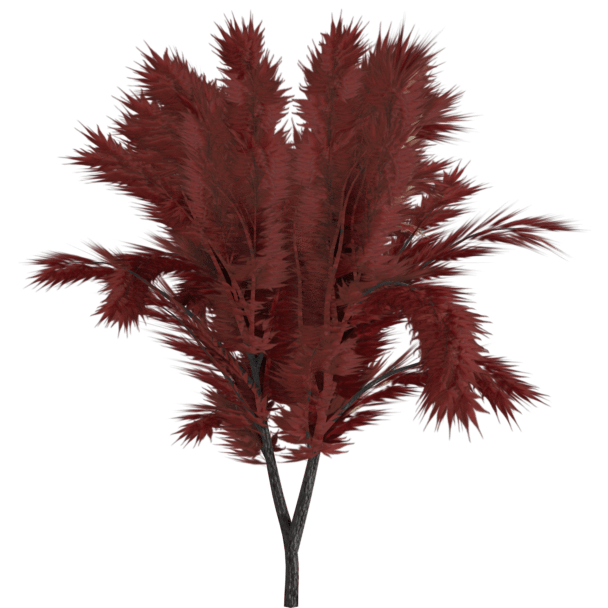}} 
& \frame{\includegraphics[width=0.135\textwidth]{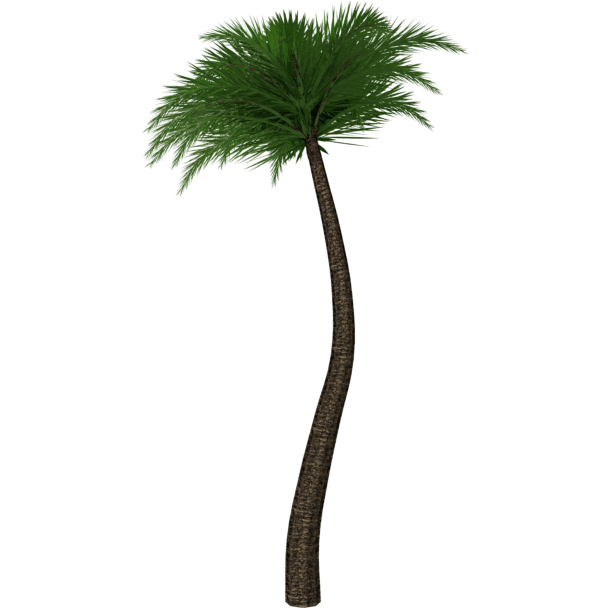}} 
& \frame{\includegraphics[width=0.135\textwidth]{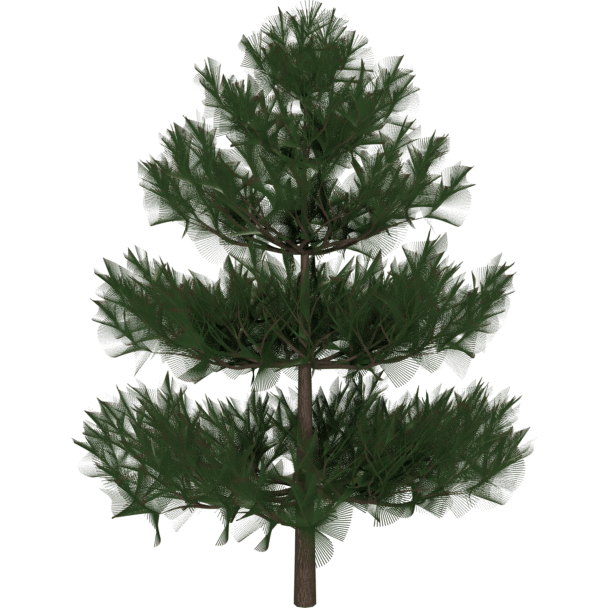}} 
& \frame{\includegraphics[width=0.135\textwidth]{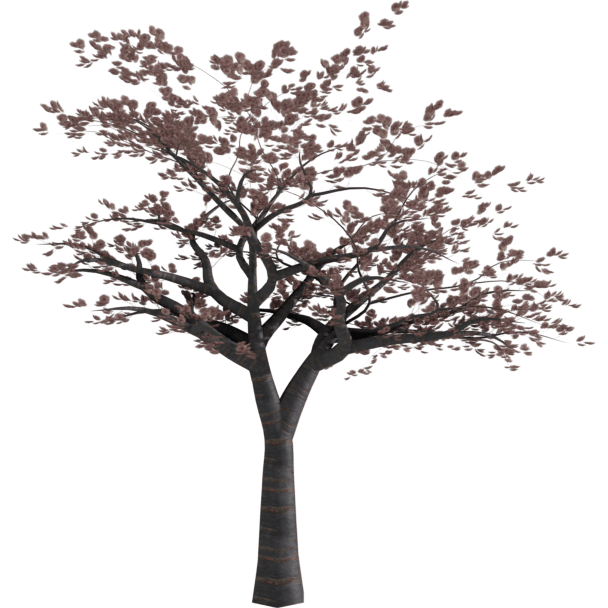}} 
& \frame{\includegraphics[width=0.135\textwidth]{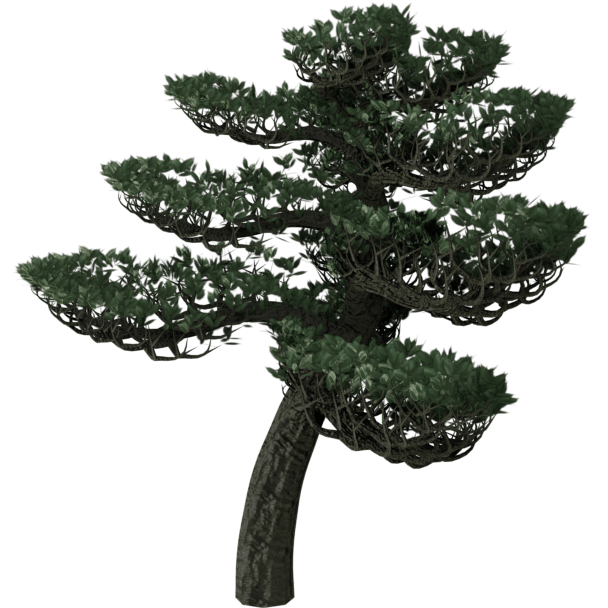}} \\
 \begin{turn}{90} \quad\quad \revision{SG}\end{turn} 
&\frame{\includegraphics[width=0.135\textwidth]{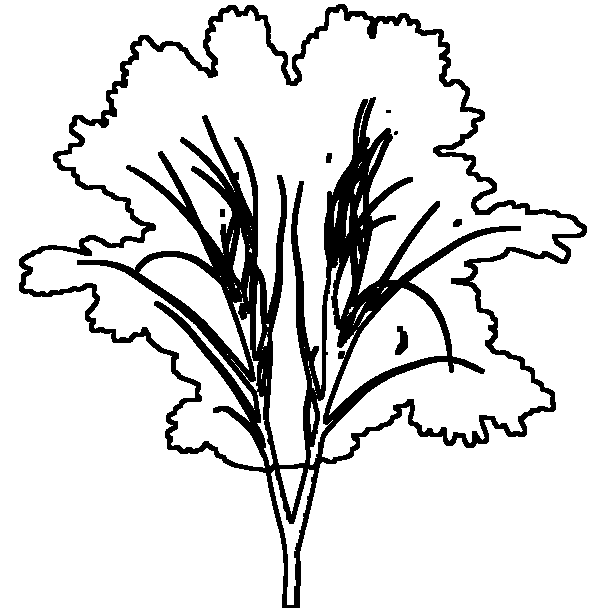}}
&\frame{\includegraphics[width=0.135\textwidth]{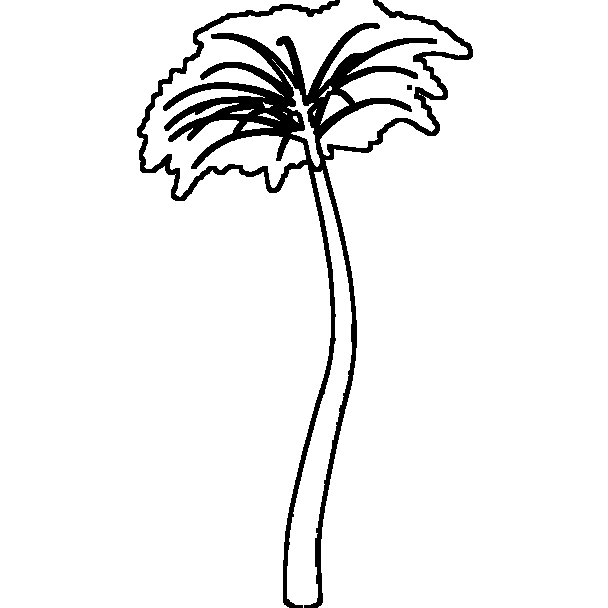}}
&\frame{\includegraphics[width=0.135\textwidth]{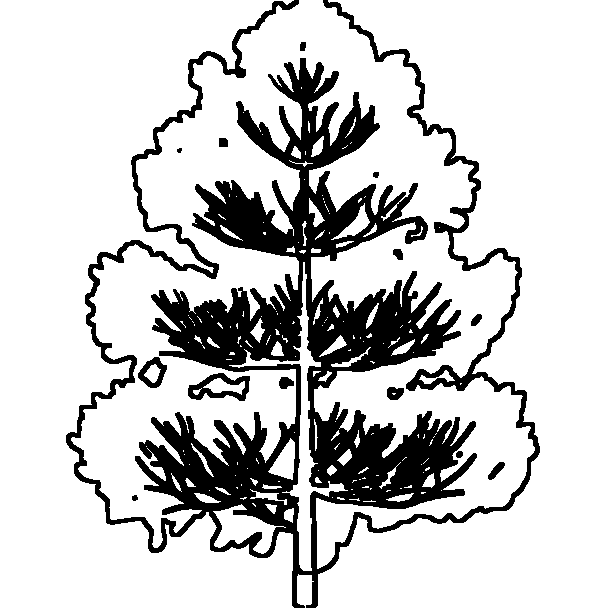}}
&\frame{\includegraphics[width=0.135\textwidth]{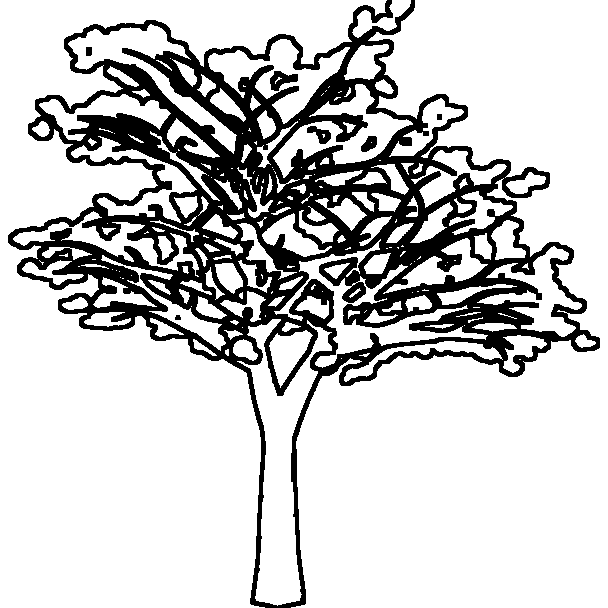}}
&\frame{\includegraphics[width=0.135\textwidth]{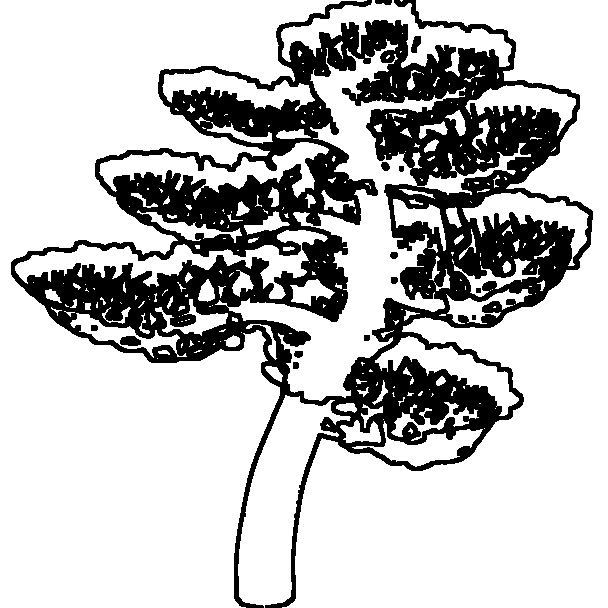}}
\\
\end{tabular}
\caption{Sample datasets for each tree \revision{species}. The first row shows a camera view of the tree reconstructed using ground truth (GT) parameters. The second row show \revision{SG} sketches corresponding to the GT views.}

\label{fig:dataSet_samples}
\end{figure}

For each camera view in the scene, we generate an \revision{SG} sketch. The final dataset therefore consists of $250$ trees $\times$ $4$ views $\times$ $5$ tree-\revision{species} $= 5000$ \revision{SG} sketches of trees ($1000$ sketches for each \revision{species}~\cite{LIU2021101115}), together with their parameter dictionaries and ground truth (GT) images. Resolution was $608 \times 608$, which is the maximum dimension supported by our tested core net (see Section~\ref{sub:backbone}). This initial resolution can be resized based on the required input of each tested TSN, including those chosen by us (see Section~\ref{sub:CNN}). The dataset is split into training and validation sets, with examples of the $5$ tree \revision{species}, together with their GT parameters, and \revision{SG} sketches, as reported in Figure~\ref{fig:dataSet_samples}. The validation set consists of a single viewpoint for each tree \revision{species}, and the training set consists of the other three viewpoints ~\cite{bishop1995neural}. A specific test dataset was created separately, as described in Section~\ref{sec:results}.

\subsection{The TSN architecture and training procedure}\label{sub:CNN}

To address the regression problem, we define a specific TSN architecture based on EfficientNet-B7~\cite{Tan2019EfficientNetRM} as the core net. This network was found to be the best core net for our architecture, after a series of empirical experiments with other well-known networks, as reported in Section~\ref{sub:backbone}. EfficientNet-B7 belongs to a family of networks created starting from an initial network architecture, called EfficientNet-B0, subsequently scaled up using the method proposed by~\cite{Tan2019EfficientNetRM}. This method aim to an improvement in performance by scaling uniformly the EfficientNet-B0 width, depth and image resolution. All the EfficientNet architectures are characterized by a convolutional layer followed by $7$ macro-blocks, each containing $L_i$ Inverted Residual Blocks, sometimes called MBConv blocks\cite{sandler2018mobilenet2}. EfficientNet-B7 is $8.4\times$ smaller, $6.1\times$ faster, and much more accurate on ImageNet than the best existing CNN.

\begin{figure*}[ht!]
\centering
\includegraphics[width=0.8\textwidth]{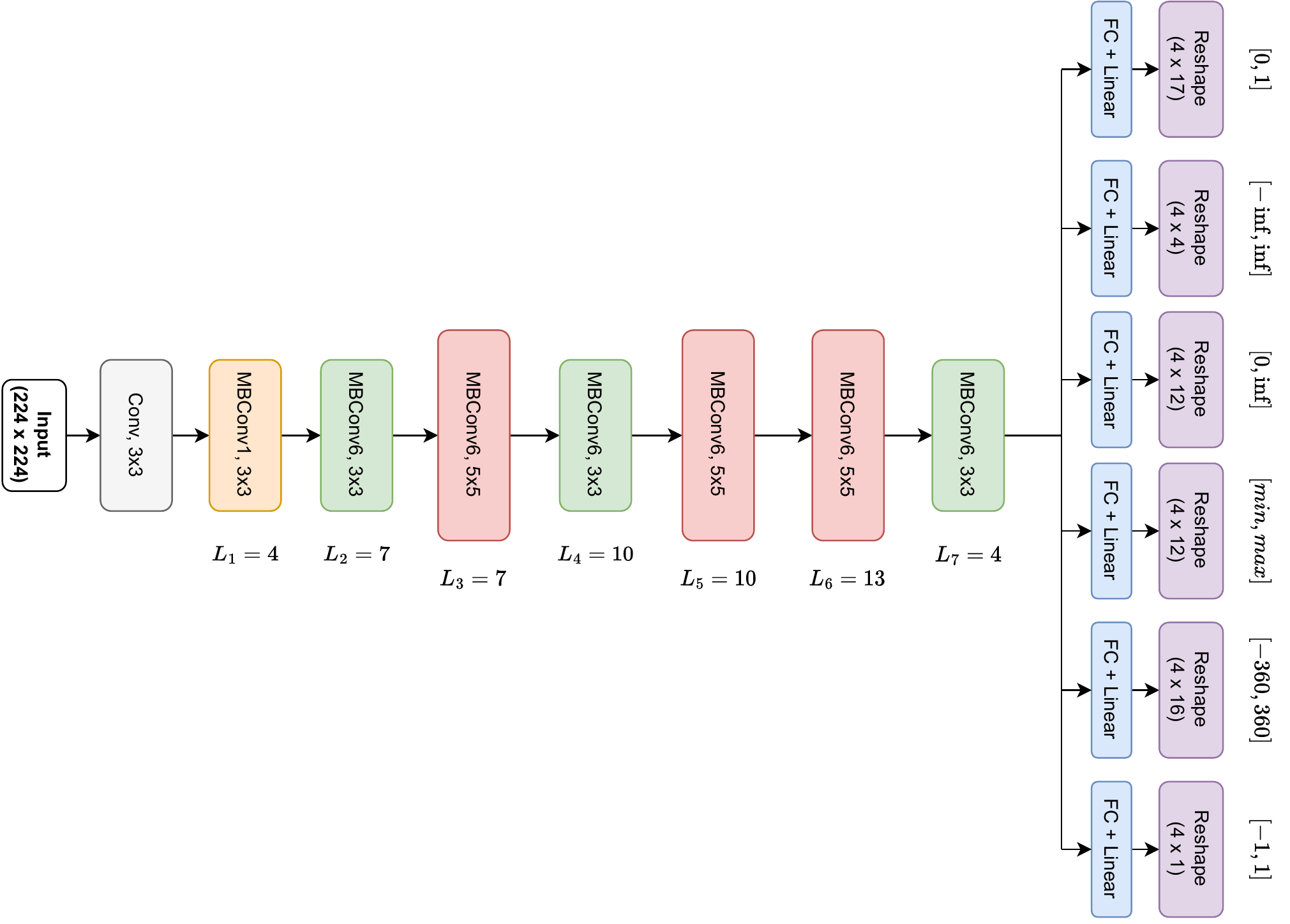}
\caption{The architecture of our TSN, with input in the form of a $224 \times 224$ sketch. The initial block is a simple convolutional layer. The following blocks contain $L_i$ MBConv blocks. The last part of the network is made up of six branches, one for each parameter order of magnitude. Each branch has a fully-connected layer with linear activation function and a reshape layer.}
\label{fig:our_net}
\end{figure*}
\revision{Our TSN input consist of $224 \times 224$ reshaped sketches (see Sec.~\ref{sub:dataset_creation}). This is because the resolution for training CNNs is generally between $64 \times 64$ and $256 \times 256$, considering that a CNN model performs better with lower input image resolutions \cite{Thambawita2021-eo}. Furthermore, \cite{Tan2019EfficientNetRM} show that the EfficientNet base model accuracy gain saturate after reaching $80\%$, also for resolutions higher than $224 \times 224$. Each sketch has no $\alpha$ channel, to stay coherent with the images contained in the dataset used to train and test EfficientNet \cite{Tan2019EfficientNetRM}. The background of each sketch is white in order to make the TSN more efficient in extracting the essential features of the image \cite{agriculture11090827}.}
As we use a supervised learning paradigm, \revision{each input image is associated with} the corresponding tree parameters, structured as a matrix that we call the target matrix. Each target matrix row represents a single parameter, and column values are parameter values. Since the number of values of each parameter item can be $4$ or $1$, our target matrix has a dimension of $4 \times n_p$, where $n_p=62$ is the number of parameters (see Details in our GitHub website~\cite{treesketchnet_params_details}). To generate a consistent matrix, 1-valued parameters are repeated in each column. The matrix contains all fixed and unfixed parameters reported in our GitHub website~\cite{treesketchnet_params_details}, without consideration of pruning, armature, and animation parameters, which are left as default values where they are mandatory. To avoid overfitting and underfitting problems (see Sec.~\ref{subsub:over_under}), based on their order of magnitude, we divide this target matrix into six sub-matrices that represent the target for each final TSN branch. The dimension of each sub-matrix is $4 \times n_t$, where $n_t \in [1;n_p]$, which represents the number of parameters for a specific order of magnitude (see details in Sec.~\ref{subsub:over_under}). Our TSN was trained using Adam~\cite{kingma2014adam}, with default parameters and $1e^{-5}$ as the initial Learning Rate and the Mean Squared Error as the loss function. To assess performance accuracy we used $(1 - RMSE) * 100$ (Root Mean Square Error) during the training phase. \revision{Because the train dataset is not so large, we used the EfficientNet-B7 layers pre-trained with} ImageNet~\cite{deng2009imagenet, Goodfellow-et-al-2016} \revision{dataset of images. This techninque is largerly used to have good Network performance with a not very large dataset \cite{Weiss2016}.}
\revision{We conducted some tests to select the best batch size value, which is equal to $8$ for our approach.} Our TSN was trained using a NVIDIA Titan XP GPU with a 12 GB G5X frame buffer. 
%\subsubsection{Overfitting and Underfitting Discussion}\label{subsub:over_under}
\subsubsection{Overfitting and underfitting}\label{subsub:over_under}
\begin{figure*}[ht!]
    \centering
    \input{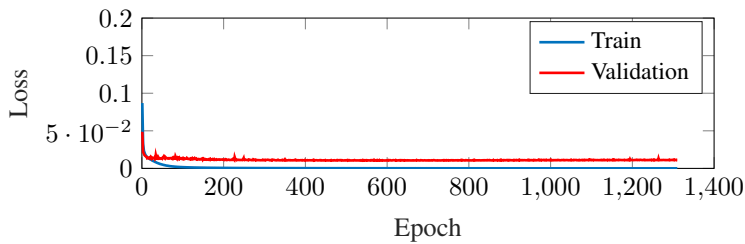}
    \caption{Training Loss vs Validation Loss of TSN. }
    \label{fig:efficientnet_loss}
\end{figure*}

The optimal dataset and the TSN configuration were identified after several attempts, in which we addressed overfitting and underfitting problems. Our first, \quotes{toy}, approach used a basic VGG-16~\cite{Simonyan15} DNN as the core net that represents $608 \times 608$ rendered (sketches and GT) images. This configuration was found to lead to general overfitting with higher intensity in some branches and very high losses. We therefore inserted dropout layers~\cite{srivastava2014dropout} before the layers of each output DNN branch with $\alpha=0.2$ for branches that were less affected by overfitting, and $\alpha=0.5$ for branches that were more affected. High loss during training could be an indication of the vanishing gradient problem and, consequently, underfitting ~\cite{kolen2001, he2015convolutional, capece2019deepflash}. To reduce this problem, we introduce a residual learning approach using skip connections~\cite{7780459} that are placed among consecutive layers in the VGG-16 core net. Although this latter approach resulted in acceptable performance, we used this experiment as a baseline for our final configuration. This configuration used a more recent version of EfficientNet
% Inception V3, 
which is able to better-manage the problems described above. 
% Two 
Four other attempts used ResNet~\cite{7780459} (variant 50), AlexNet~\cite{krizhevsky2012imagenet}, Inception V3~\cite{szegedy2016rethinking} and CoAtNet~\cite{Dai2021CoAtNetMC} as core nets. 
% The former is a 
ResNet and Inception V3 are classic neural networks~\cite{khan2020survey} that 
% is 
are used in many computer vision tasks, 
% while the latter 
AlexNet was used in comparable study with a similar sketch-to-mesh-parameter approach~\cite{huang2017}, while CoAtNet and EfficientNet represent the SOTA for the image recognition. Section~\ref{sub:backbone} reports quantitative performance comparisons between VGG-16 with skip connections, and EfficientNet-B7.
Using VGG-16 as the core net, we experimented with using the entire target matrix as the output of the DNN, with only one output branch. However, the results were very poor, and the DNN was completely underfitted. We therefore defined $6$ DNN output branches, based on the order of magnitude of parameter values. This was possible because the Weber-Penn Blender parameter dictionary consists of heterogeneous data, as reported in our GitHub website~\cite{treesketchnet_params_details}. Therefore, we were able to carry out conversions and resize many of them, as a function of their order of magnitude. Specifically, we defined several sub-matrices, one for each DNN output branch, as shown in Figure~\ref{fig:our_net}. Discrete integers and string parameter values were parsed as floats; Boolean parameter values were parsed as integers; non-numeric string parameter values were parsed as floats using the Labeled Encoding method (\eg \textit{Leaf Shape} described in our GitHub website~\cite{treesketchnet_params_details}; and binary $[-1; 1]$ parameters were converted to $[0;1]$.  
Although it is not possible to consider $[-\inf, \inf]$ values, we normalized them and $[-360, 360]$ values to a $[-1,1]$ range using the Max Abs Scaling method. This scaled the data and preserved sparsity. Maximum values were stored in normalization matrices and reused in the testing step. Tests of Inception V3 identified significant overfitting on $[-\inf, +\inf]$ and $[0, 1]$ DNN output branches, which were reduced by adding a $L2$ regularization~\cite{ng2004feature} on the first two fully-connected layers, followed by a dropout layer with $\alpha=0.2$. We obtained little overfitting on the $[-360, 360]$ DNN output branch; in this case we only used a dropout layer with $\alpha=0.5$ to reduce the problem (see Figure~\ref{fig:our_net}). \revision{EfficientNet-B7 performs good on each branch, with slight better performance on $[0, \infty]$, $[-1, 1]$, $[-360, 360]$, and $[min, max]$ branches. For the sake of simplicity, Figure \ref{fig:efficientnet_loss} shows the general training and validation loss calculated by averaging the losses of the various branches.}
\section{Results}\label{sec:results}

\renewcommand{\tabcolsep}{0.05pt}
\begin{figure}
\centering
%\begin{adjustbox}{max width=\linewidth}
    %\begin{tabular}{@{} C @{} D @{} D @{} D @{} D @{}}
		\begin{tabular}{lccccc}
    %\begin{tabular}{\columnwidth}{@{\extracolsep{0pt}}l|c|c|c|c|c}
    %\begin{tabular}{p{\dimexpr0.005\linewidth-2\tabcolsep-1.5\arrayrulewidth\relax}| *{5}{p{\dimexpr0.193\linewidth-2\tabcolsep-1.5\arrayrulewidth\relax}|}}
		\rotatebox{90}{\footnotesize \ Ground Truth \ \ Reconstructed \quad \quad Input} 
		& \includegraphics[width=0.193\textwidth]{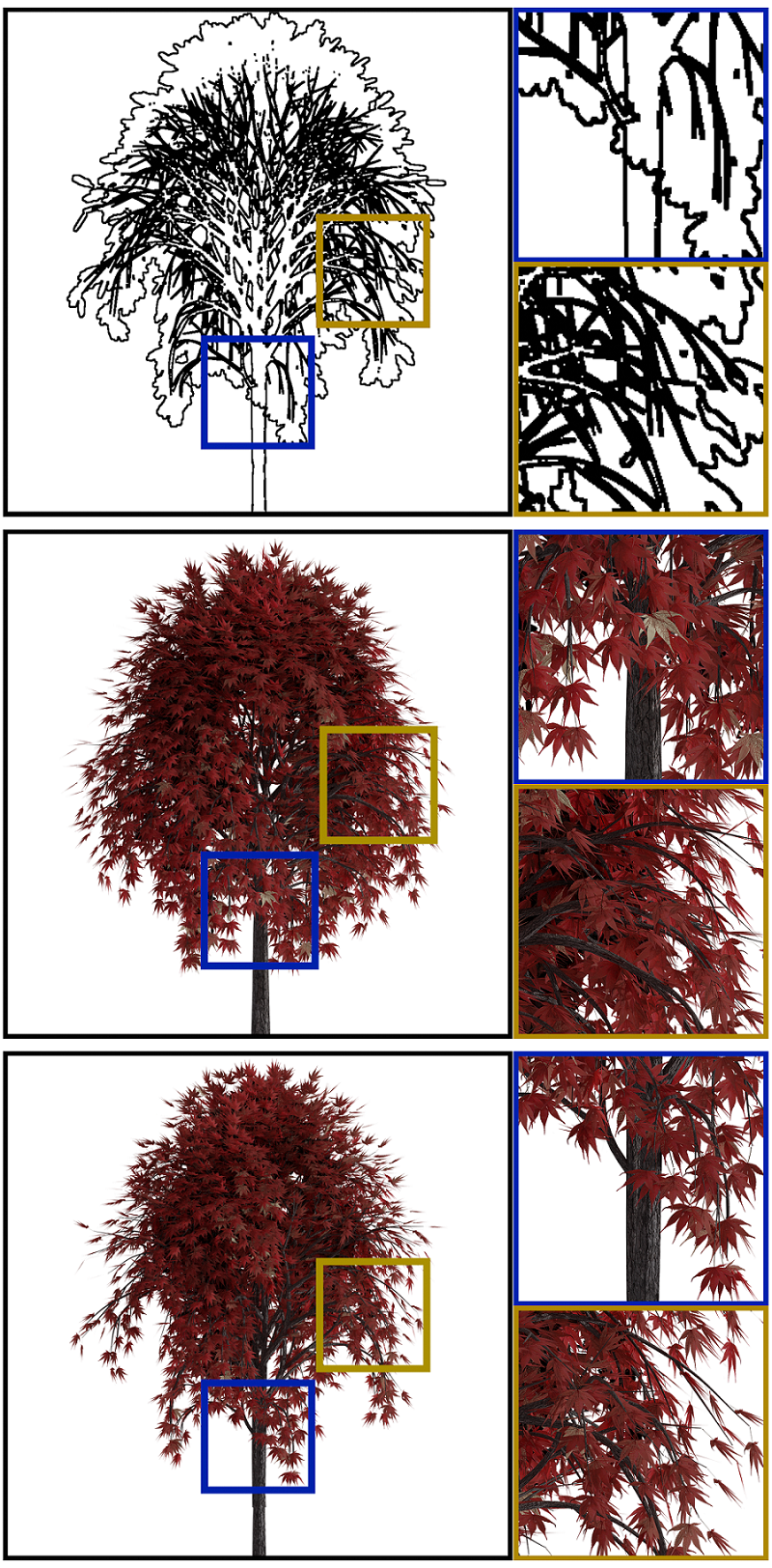}
		& \includegraphics[width=0.193\textwidth]{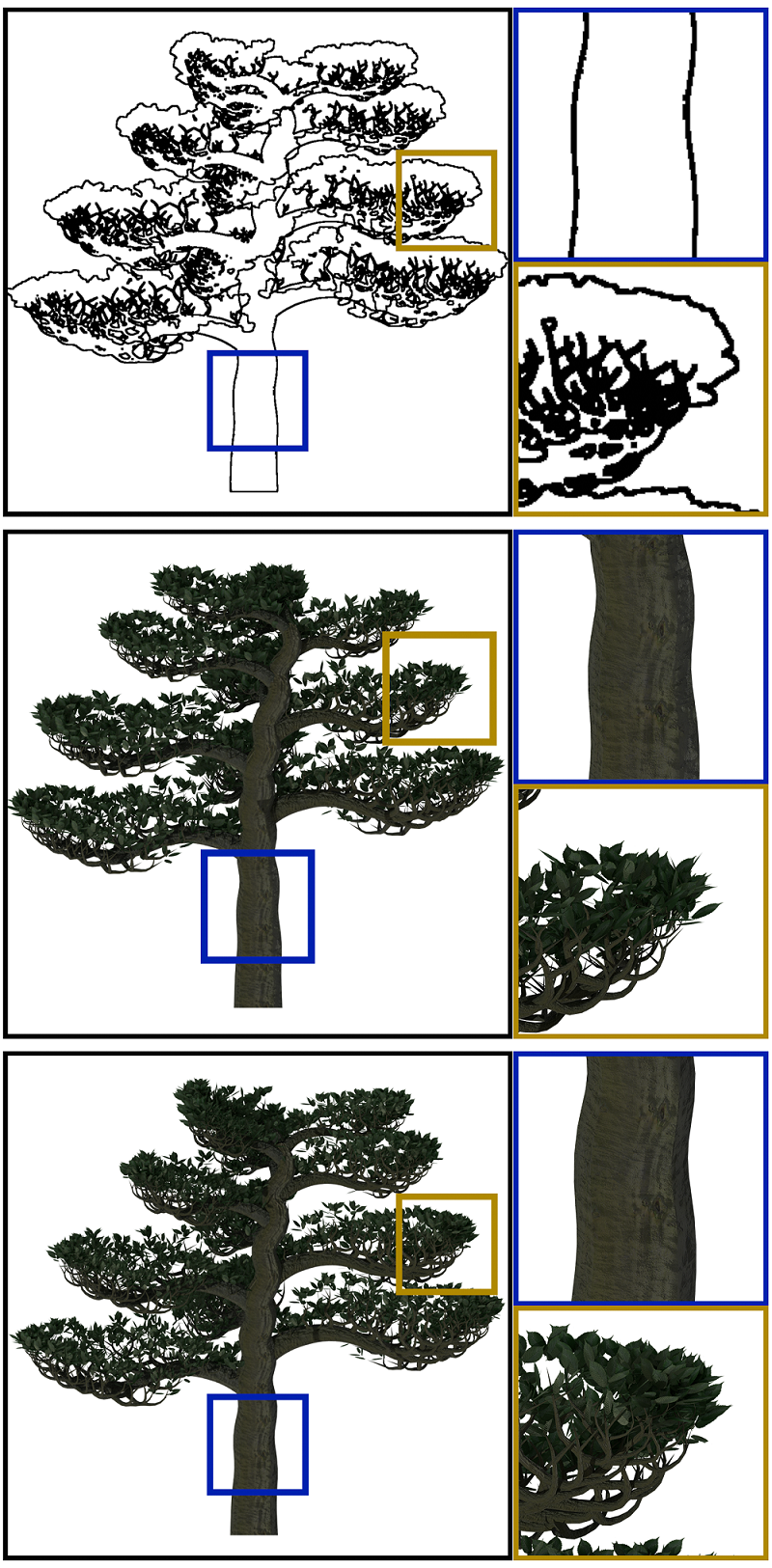}
		& \includegraphics[width=0.193\textwidth]{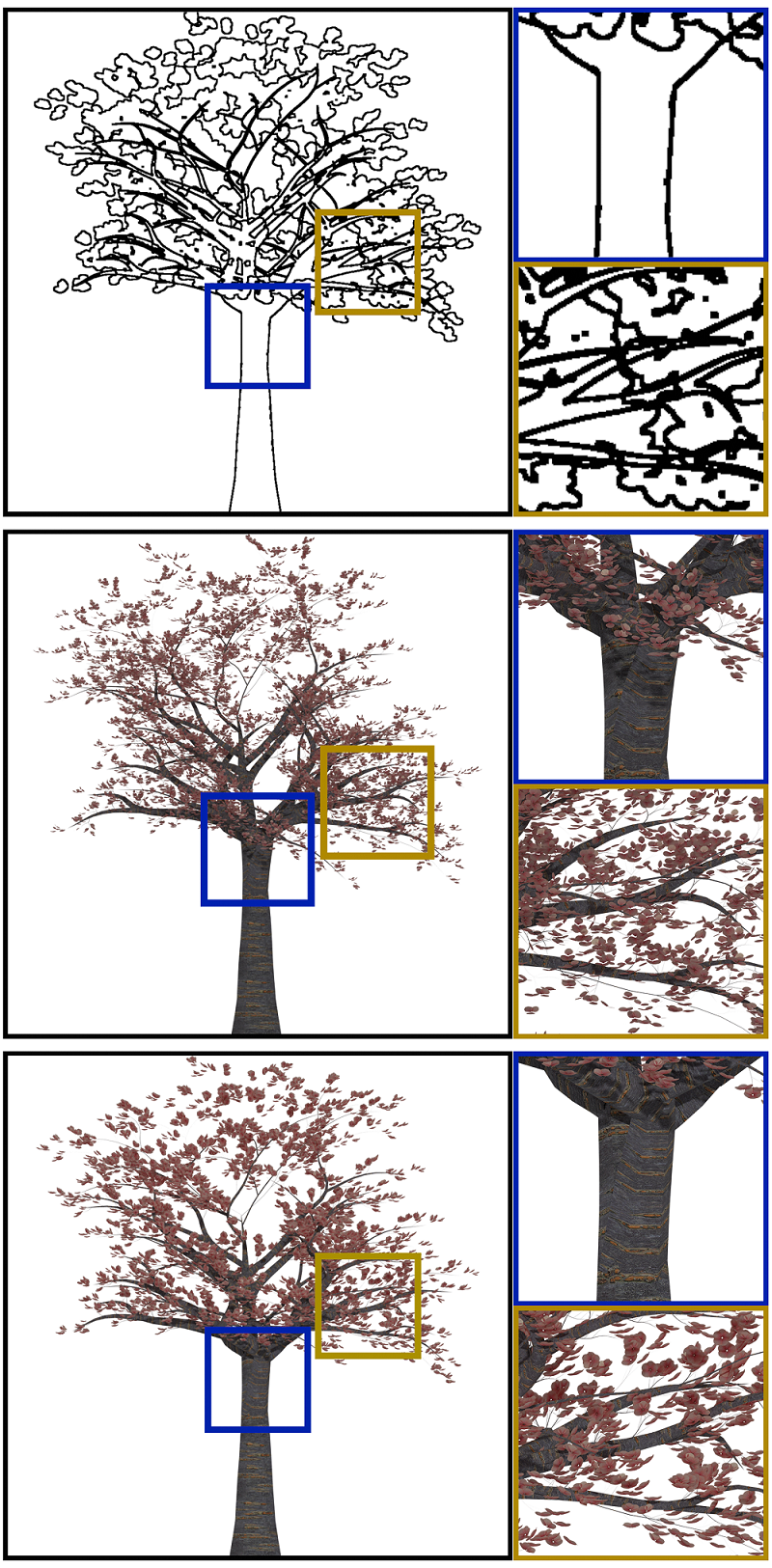}
		& \includegraphics[width=0.193\textwidth]{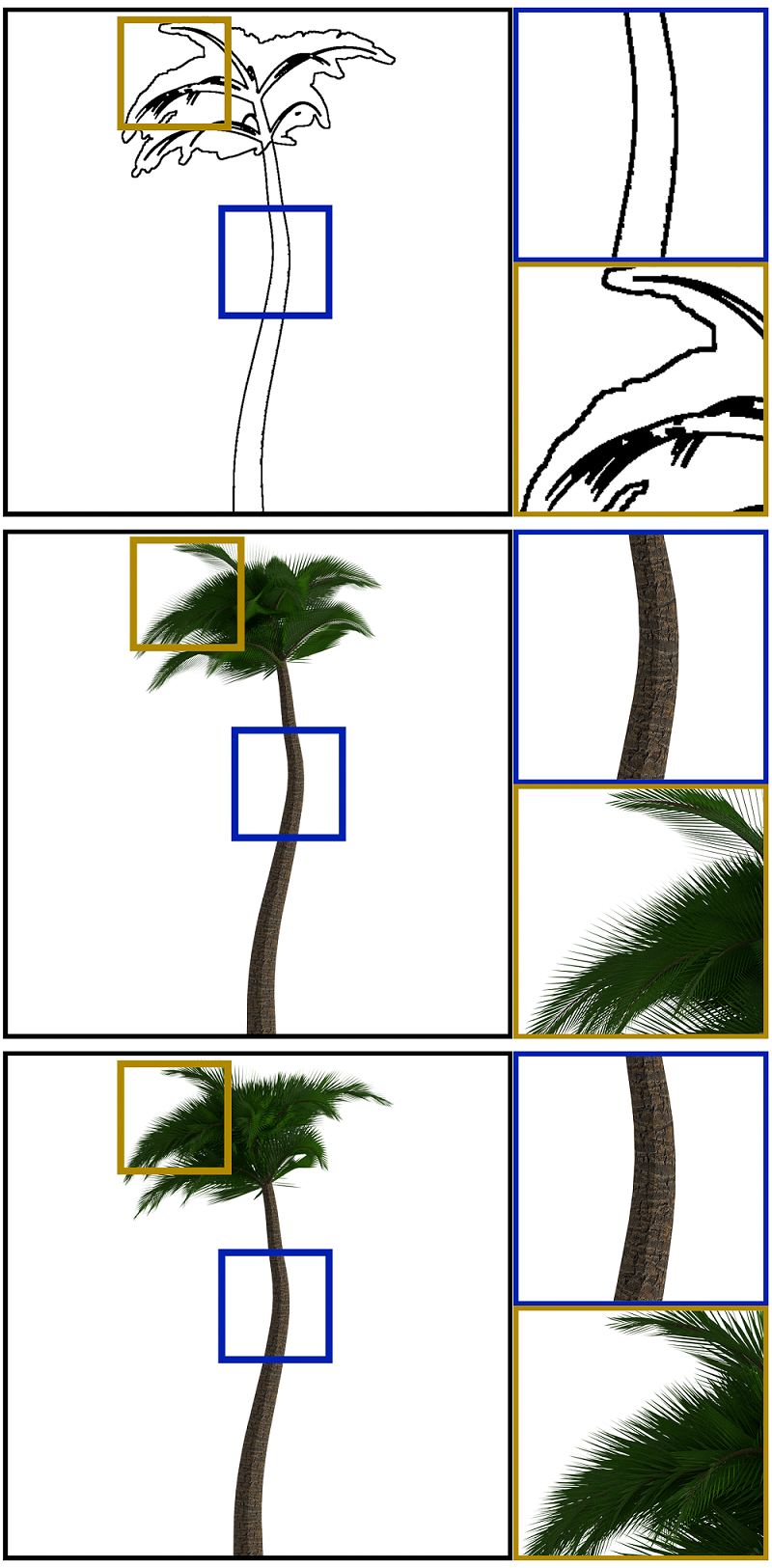}
		& \includegraphics[width=0.193\textwidth]{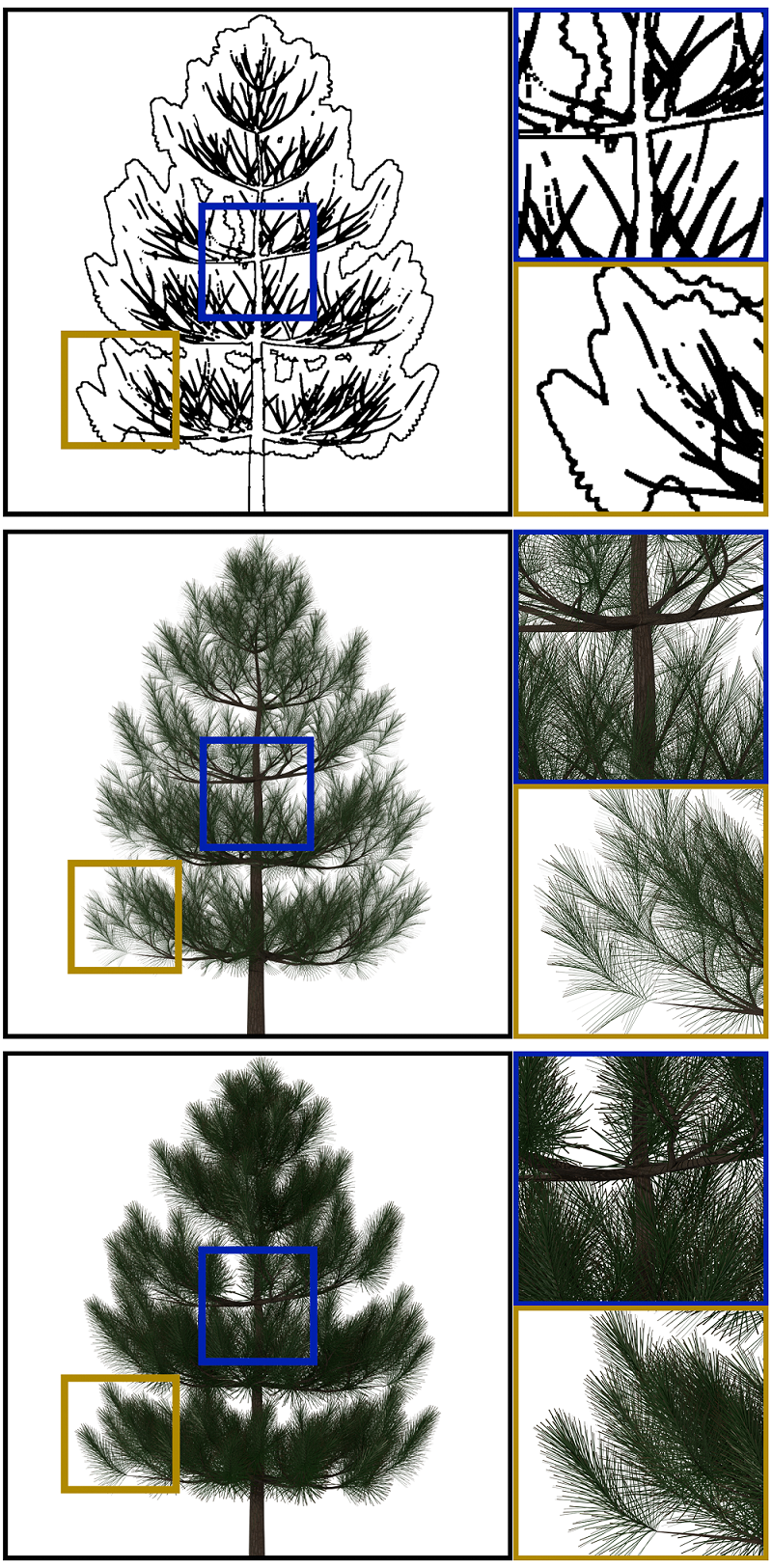}
		\end{tabular}

    \caption{The results of the \revision{SG} test set obtained from our RT add-on taken from the same camera view, for each tree-\revision{species}. The first row shows the input \revision{SG} sketches; the second row shows the reconstructed 3D meshes using the corresponding predicted parameters dictionaries; the last row shows the GT.}
    \label{fig:results_1}
\end{figure}
\renewcommand{\tabcolsep}{2pt}

\begin{figure*}[ht!]
\centering
 \begin{tabular}{c c c c} 
    \rotatebox{90}{\parbox{6em}{\centering \small Input}}
    & \frame{\includegraphics[width=0.15\textwidth]{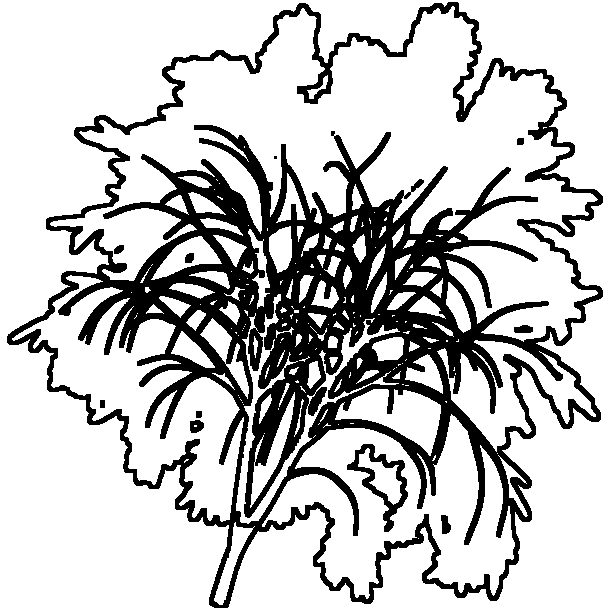}}
    & \frame{\includegraphics[width=0.15\textwidth]{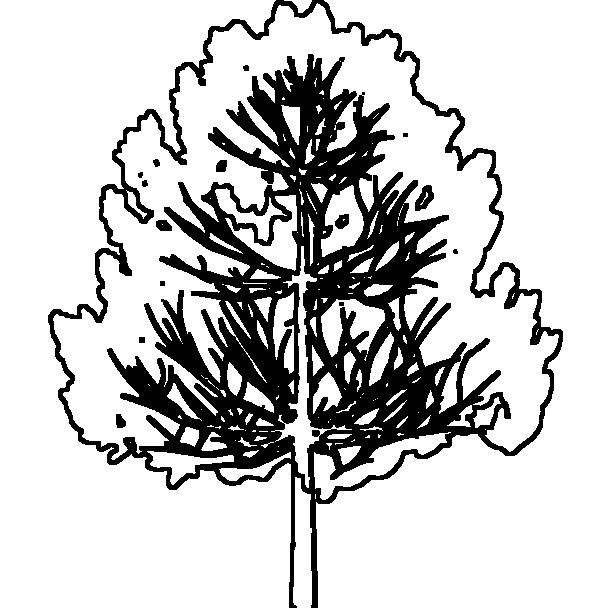}}
    & \frame{\includegraphics[width=0.15\textwidth]{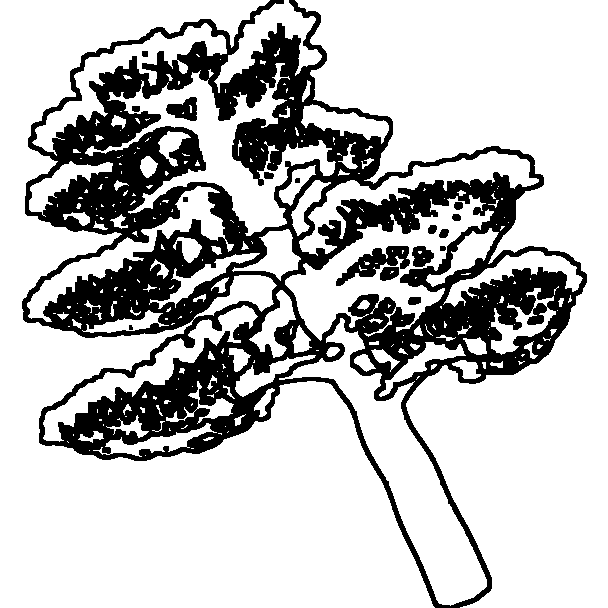}}\\ 
    \rotatebox{90}{\parbox{6em}{\centering \small Reconstructed}}
    & \frame{\includegraphics[width=0.15\textwidth]{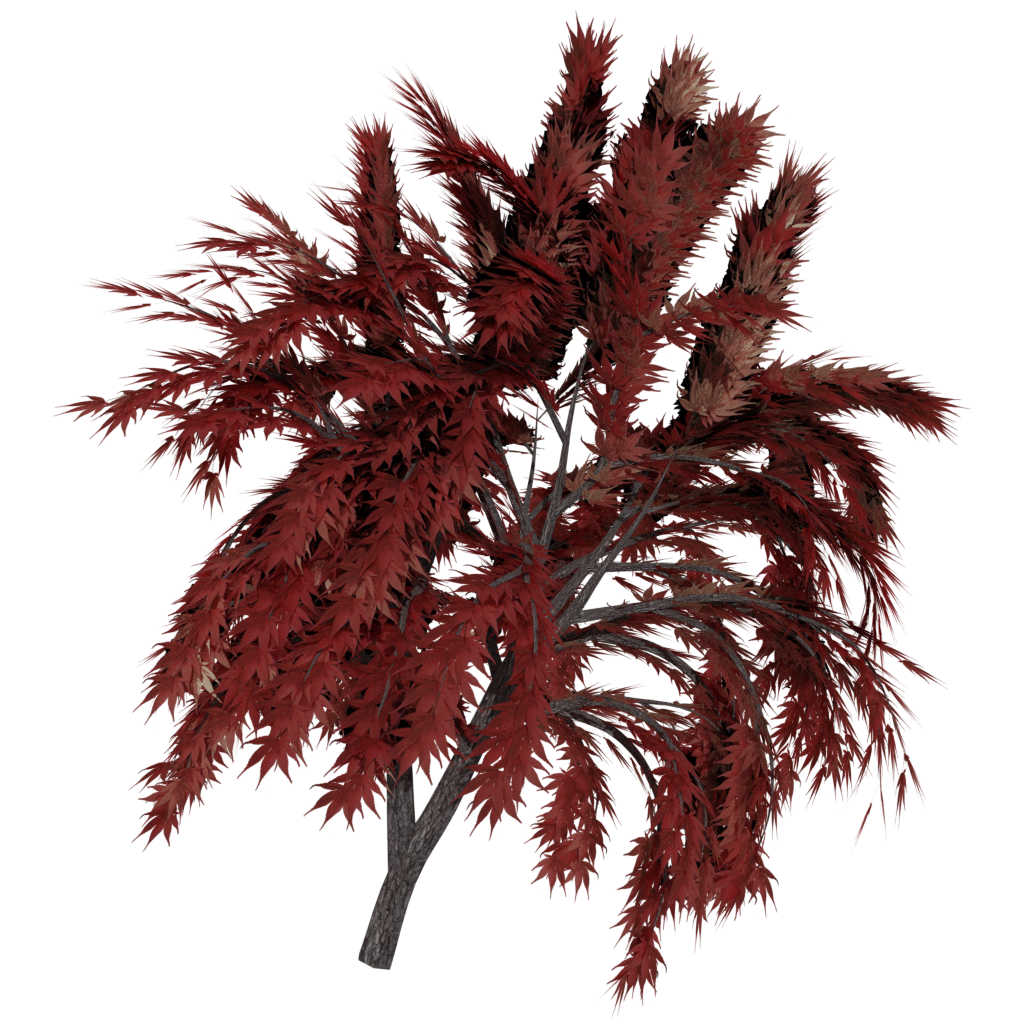}}
    & \frame{\includegraphics[width=0.15\textwidth]{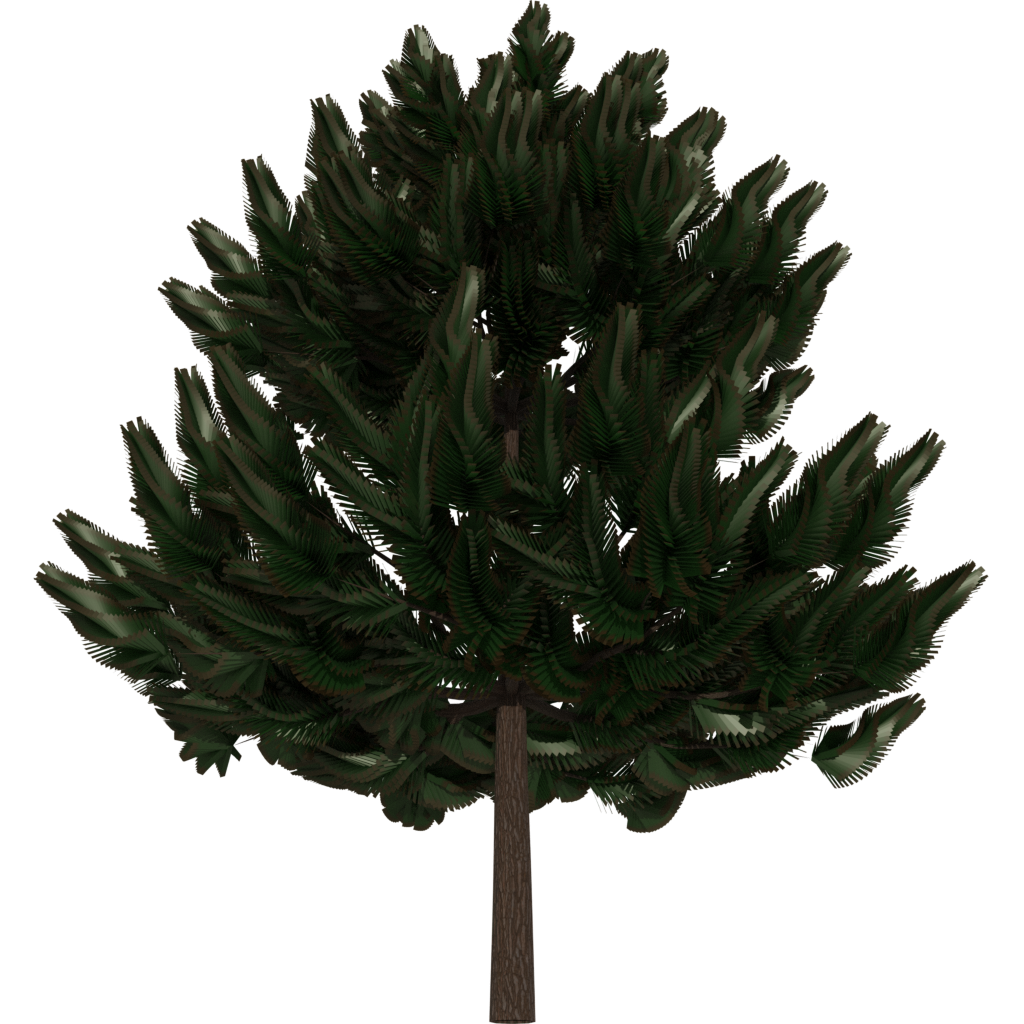}}
    & \frame{\includegraphics[width=0.15\textwidth]{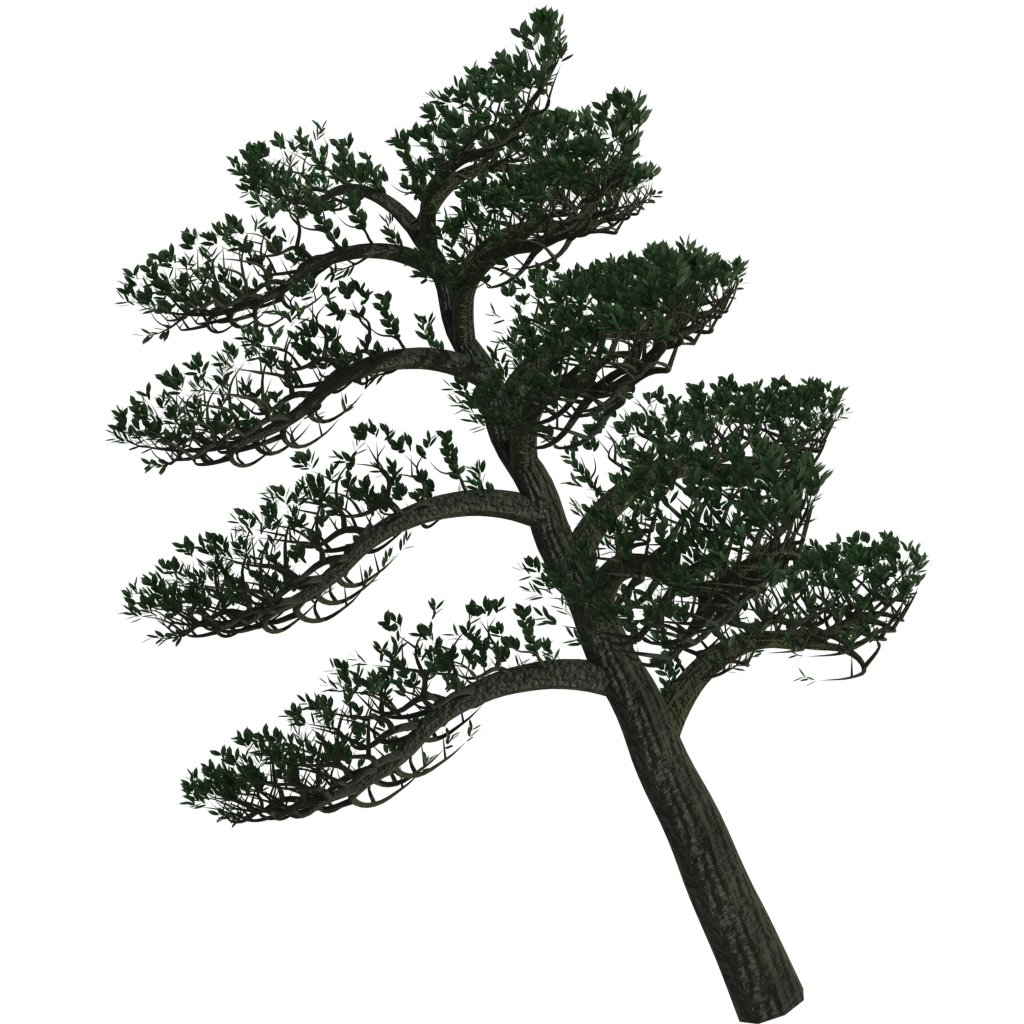}}\\ 
    \rotatebox{90}{\parbox{6em}{\centering \small Ground Truth}}
    & \frame{\includegraphics[width=0.15\textwidth]{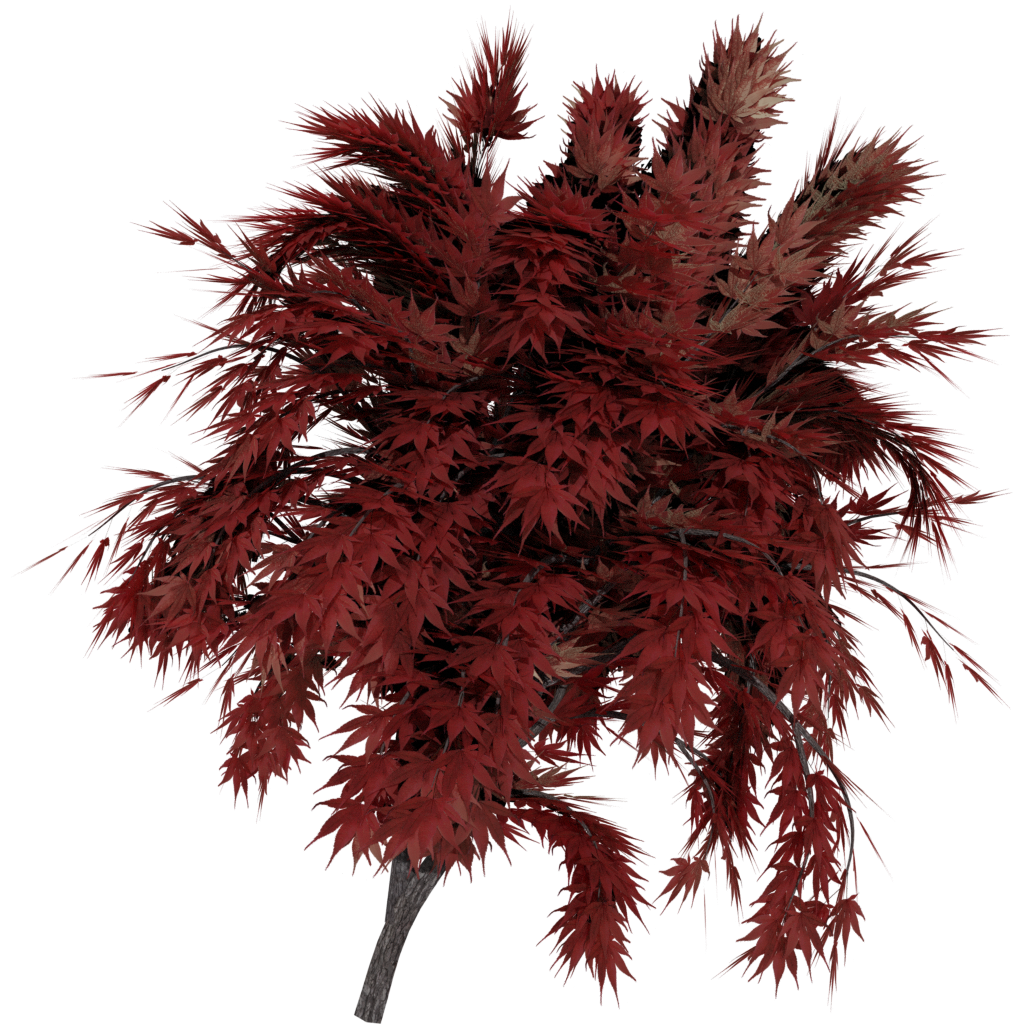}}
    & \frame{\includegraphics[width=0.15\textwidth]{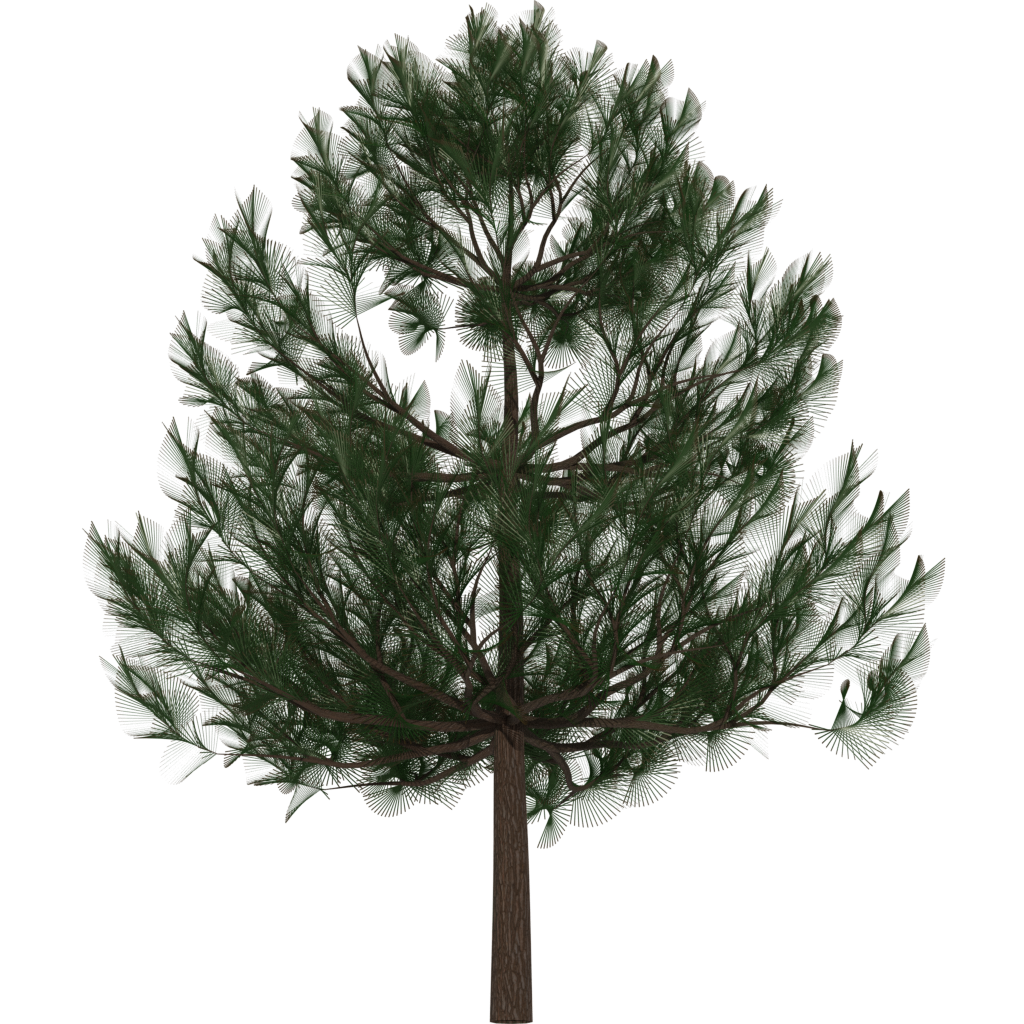}}
    & \frame{\includegraphics[width=0.15\textwidth]{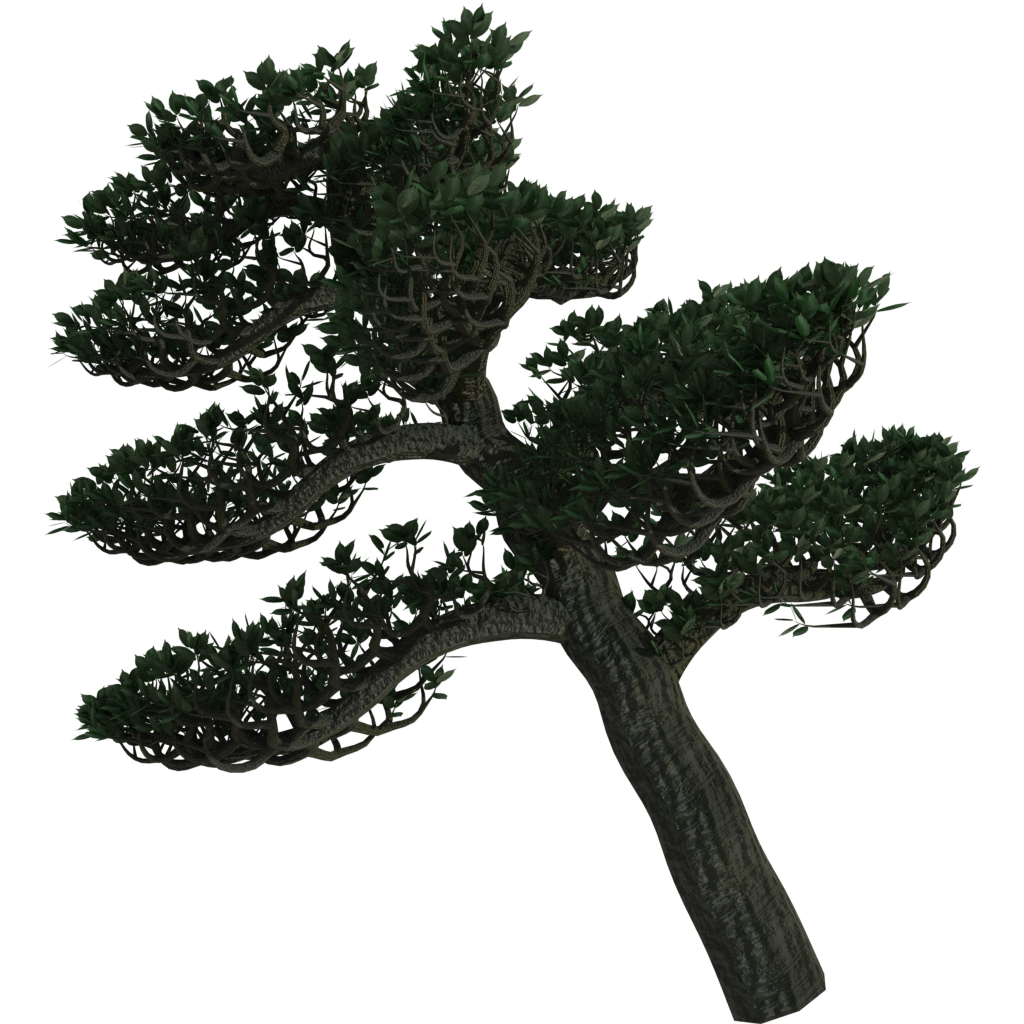}}\\
	\end{tabular}
	\caption{Results of TSN network given images with different camera views: in the left column the camera was right tilted; in the middle was bottom placed and tilted upward; in the last column the camera was left tilted.}
	\label{fig:images_different_views}
\end{figure*}

\begin{figure}
    \centering
    \begin{tikzpicture}
     \node at (-3.8,0) {\frame{\includegraphics[width=0.185\textwidth]{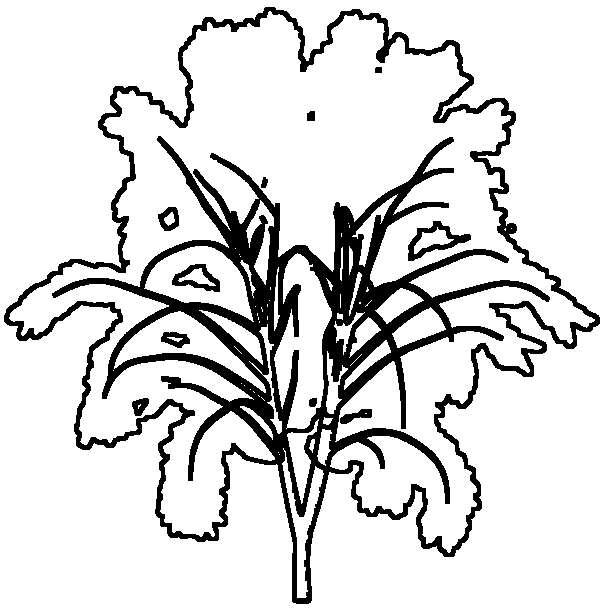}}};
     \node at (-1.45,0.78) {\frame{\includegraphics[width=0.09\textwidth]{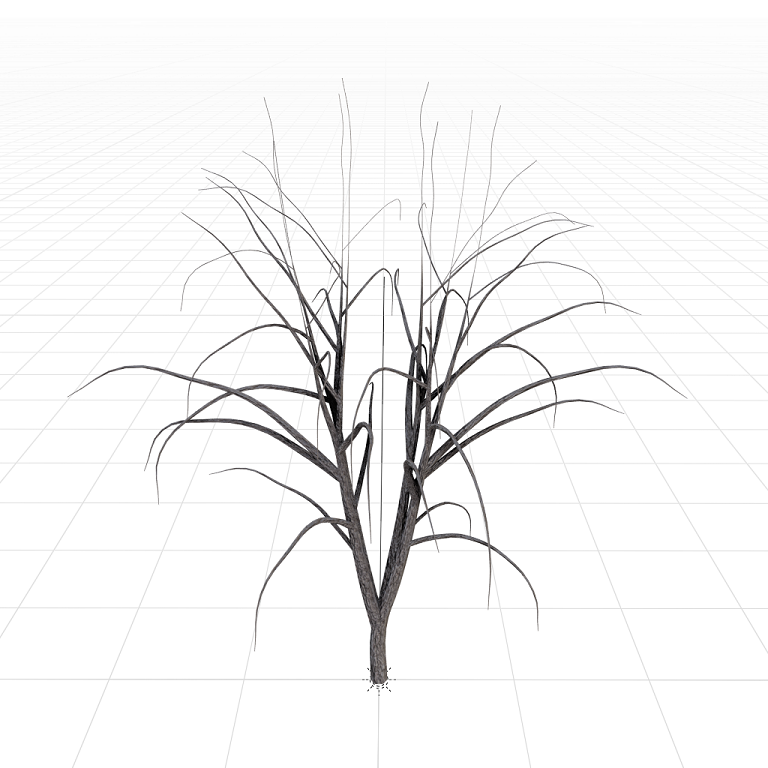}}};
     \node at (-1.45,-0.78) {\frame{\includegraphics[width=0.09\textwidth]{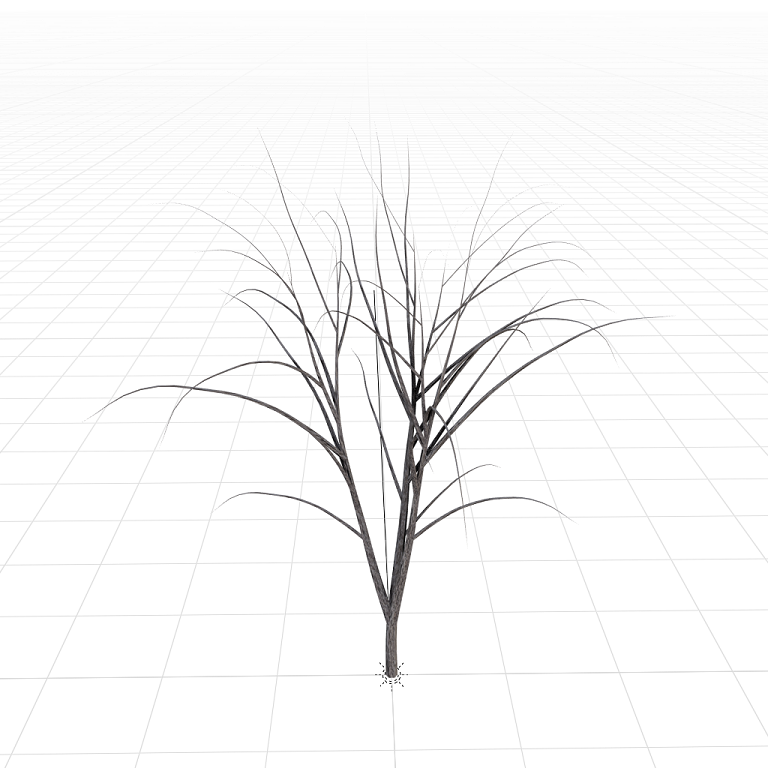}}};
     \node at (0.0,0.78) {\frame{\includegraphics[width=0.09\textwidth]{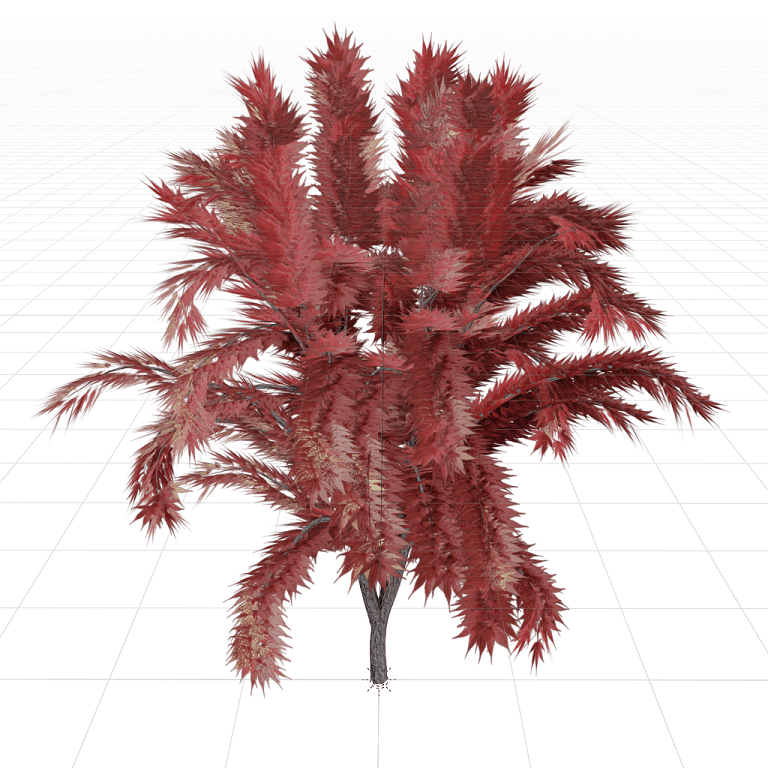}}};
     \node at (0.0,-0.78) {\frame{\includegraphics[width=0.09\textwidth]{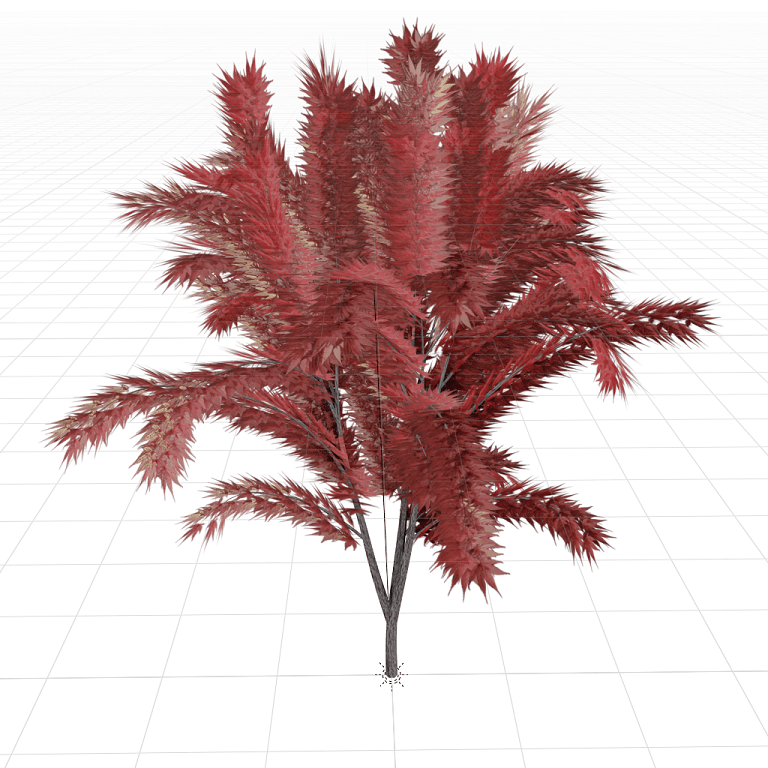}}};
     \node at (1.55,0.78) {\frame{\includegraphics[width=0.09\textwidth]{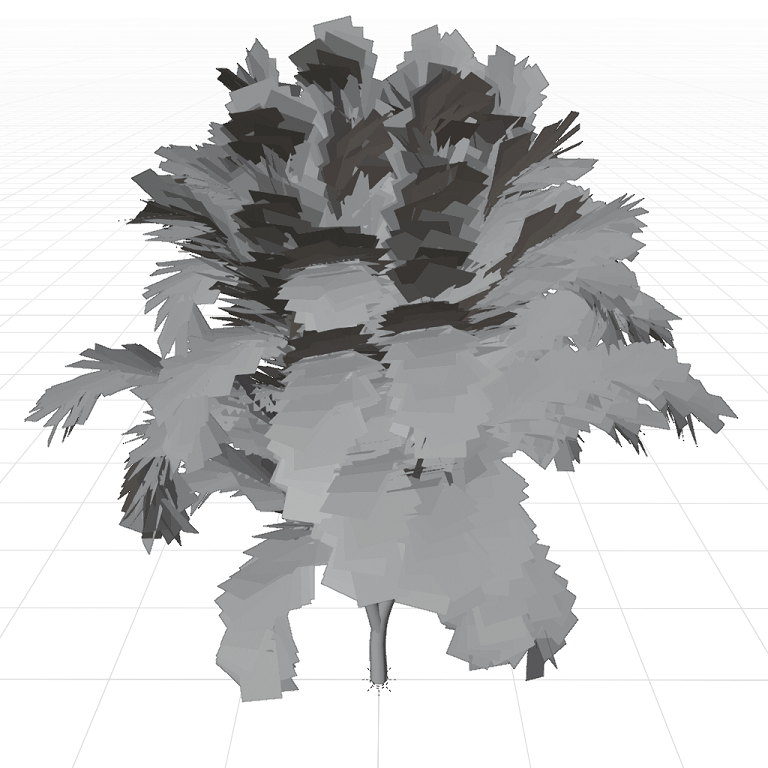}}};
     \node at (1.55,-0.78) {\frame{\includegraphics[width=0.09\textwidth]{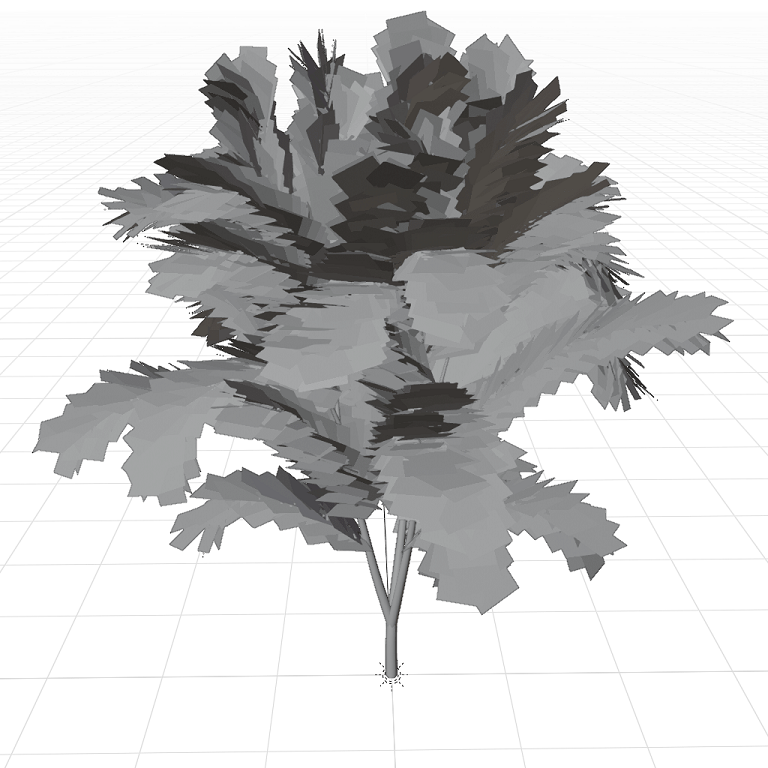}}};
    \node[text=black] at (-3.8,1.7) {\scriptsize Sketch};
    \node[text=black] at (-1.45,1.7) {\scriptsize Skeleton};
    \node[text=black] at (0.0,1.7) {\scriptsize Material};
    \node[text=black] at (1.55,1.7) {\scriptsize Mesh};

    \end{tikzpicture}
    \begin{tikzpicture}
     \node at (-1.97,0) {\frame{\includegraphics[width=0.185\textwidth]{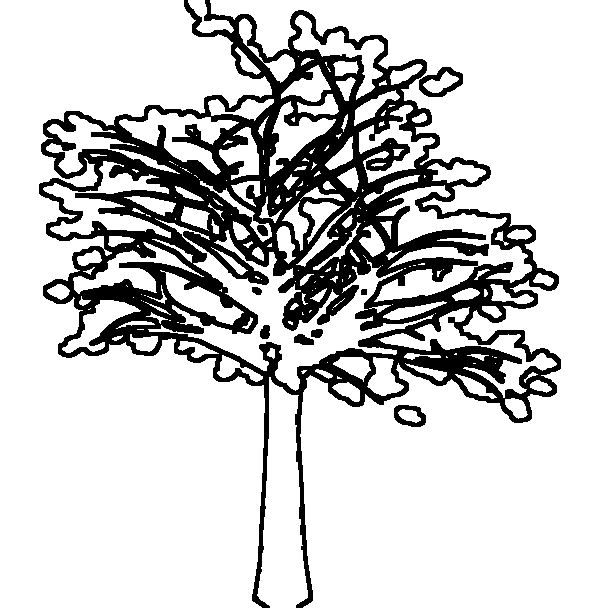}}};
     \node at (0.4,0.78) {\frame{\includegraphics[width=0.09\textwidth]{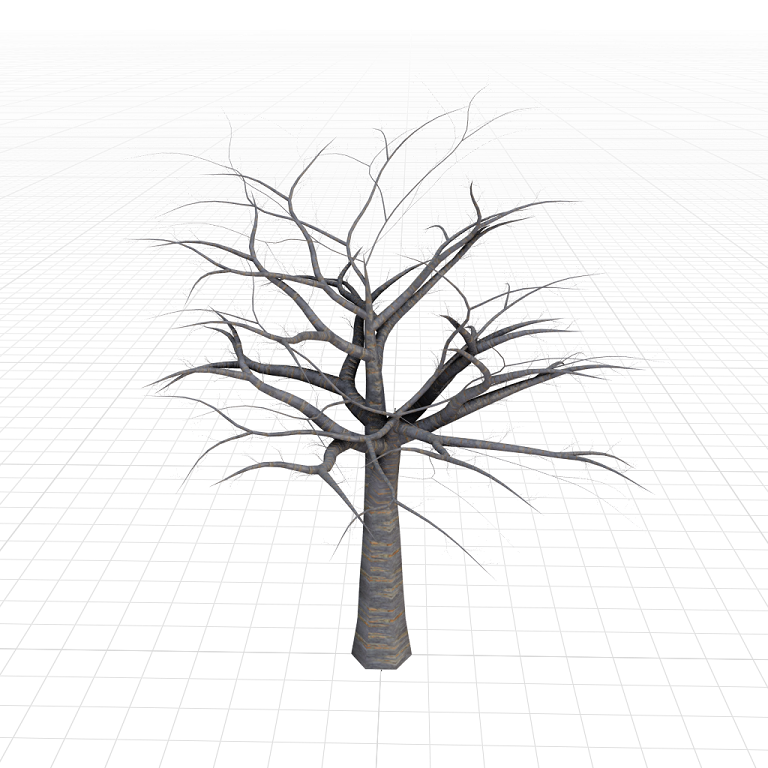}}};
     \node at (0.4,-0.78) {\frame{\includegraphics[width=0.09\textwidth]{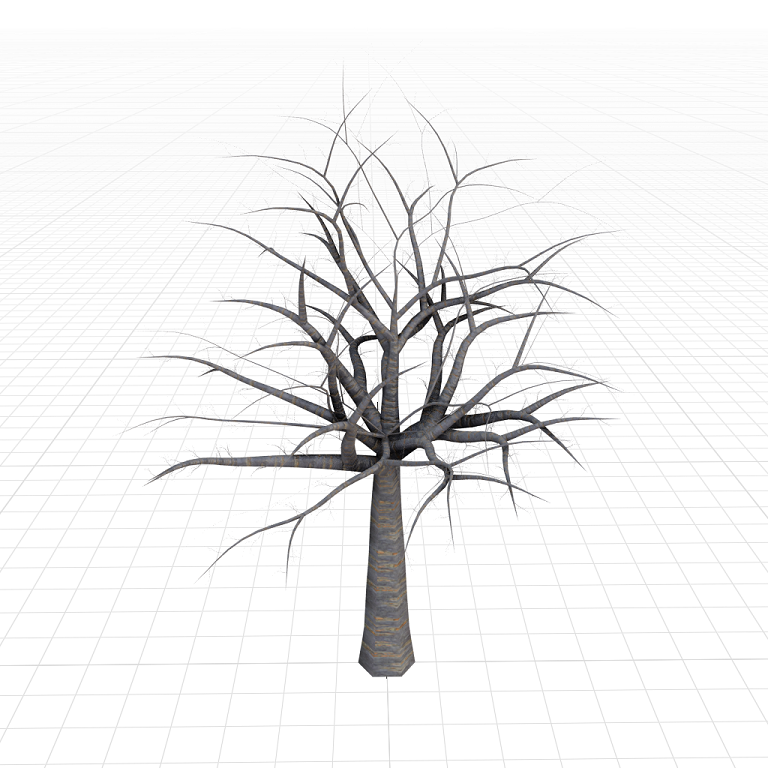}}};
     \node at (1.95,0.78) {\frame{\includegraphics[width=0.09\textwidth]{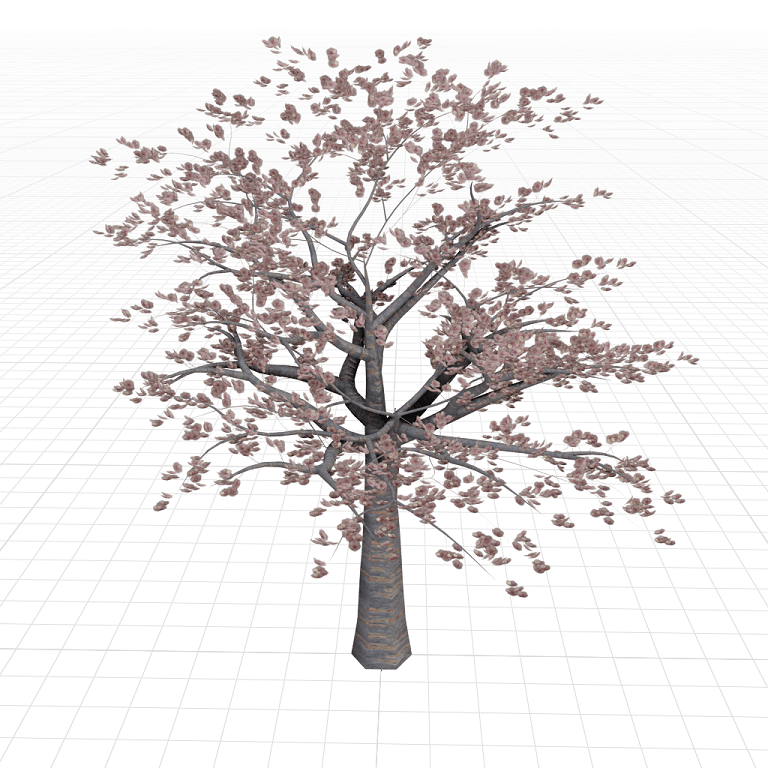}}};
     \node at (1.95,-0.78) {\frame{\includegraphics[width=0.09\textwidth]{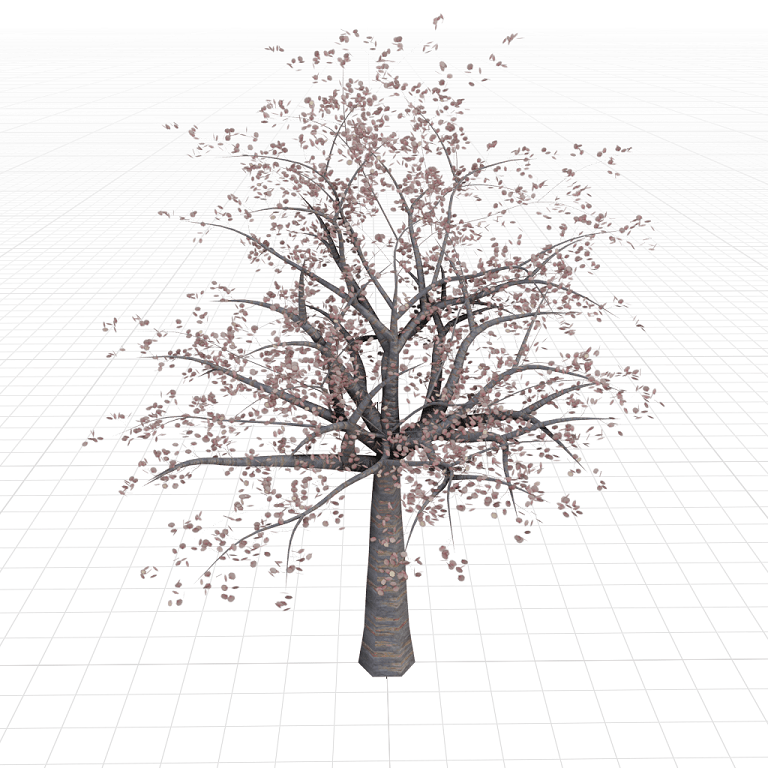}}};
     \node at (3.5,0.78) {\frame{\includegraphics[width=0.09\textwidth]{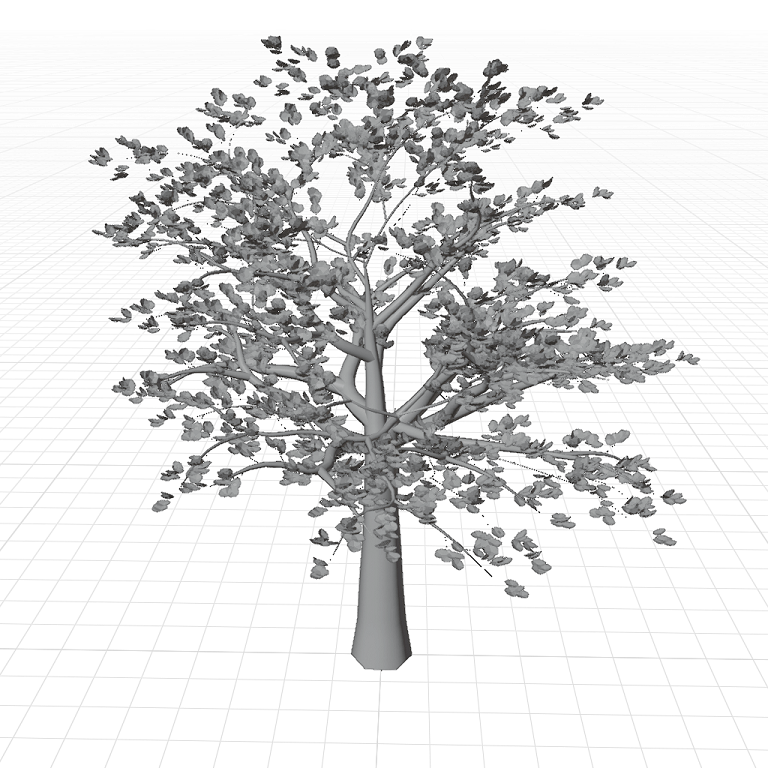}}};
     \node at (3.5,-0.78) {\frame{\includegraphics[width=0.09\textwidth]{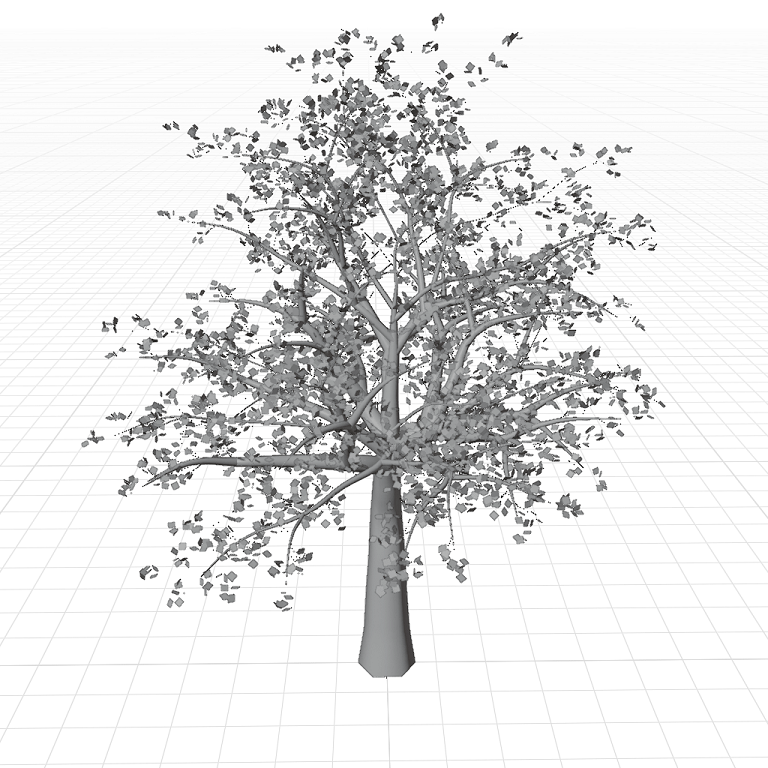}}};
    \node[text=black] at (-1.97,1.7) {\scriptsize Sketch};
    \node[text=black] at (0.4,1.7) {\scriptsize Skeleton};
    \node[text=black] at (1.95,1.7) {\scriptsize Material};
    \node[text=black] at (3.5,1.7) {\scriptsize Mesh};
    
    \node[text=black] at (4.4,0.65) {\rotatebox{-90}{\tiny Ground Truth}};
    \node[text=black] at (4.4,-0.65) {\rotatebox{-90}{\tiny Prediction}};
    \end{tikzpicture}
    \begin{tikzpicture}
     \node at (0.65,0) {\frame{\includegraphics[width=0.185\textwidth]{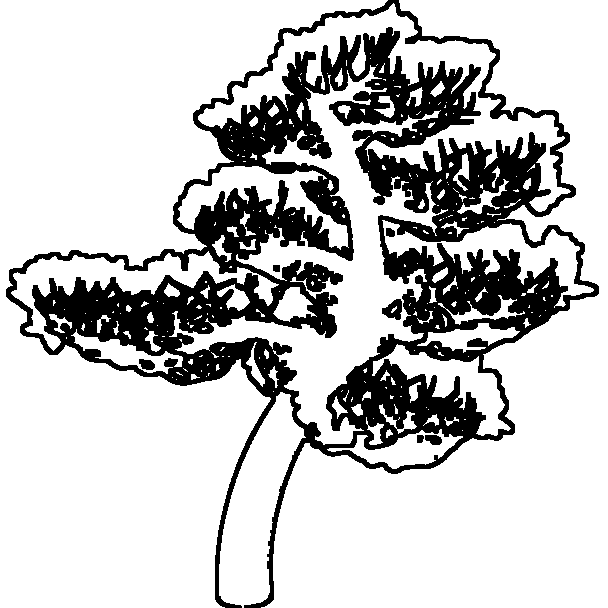}}};
     \node at (3.0,0.78) {\frame{\includegraphics[width=0.09\textwidth]{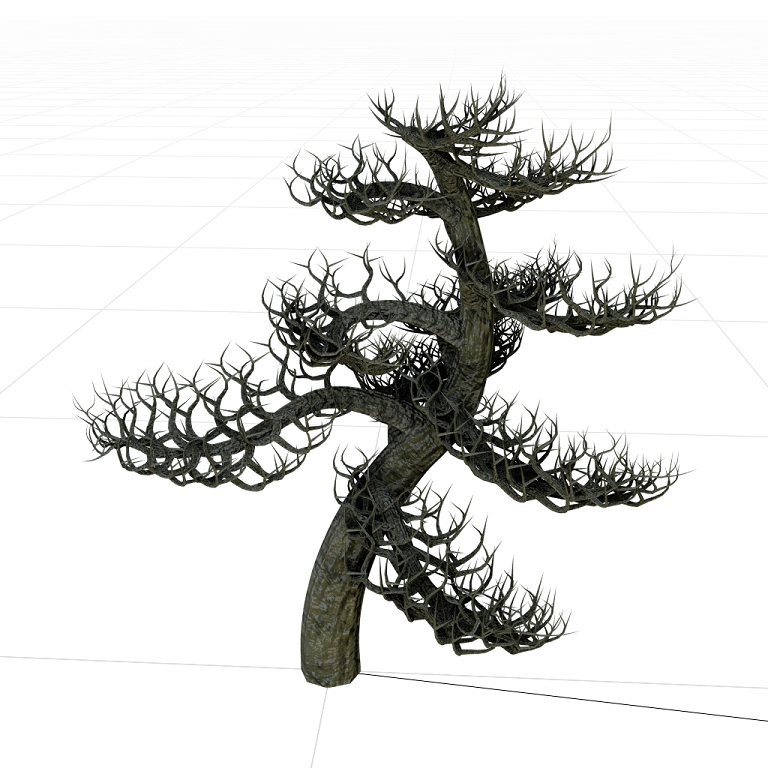}}};
     \node at (3.0,-0.78) {\frame{\includegraphics[width=0.09\textwidth]{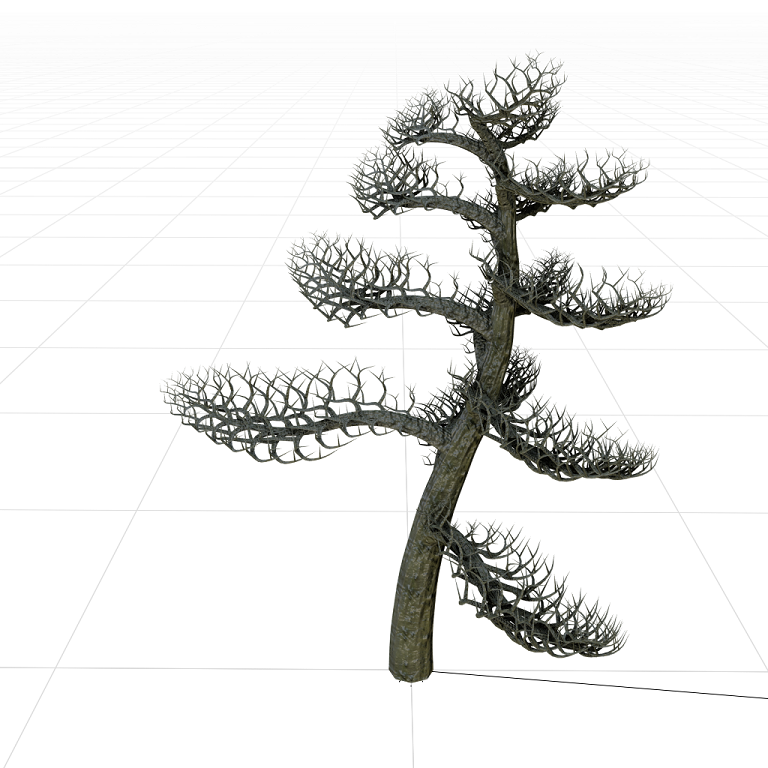}}};
     \node at (4.55,0.78) {\frame{\includegraphics[width=0.09\textwidth]{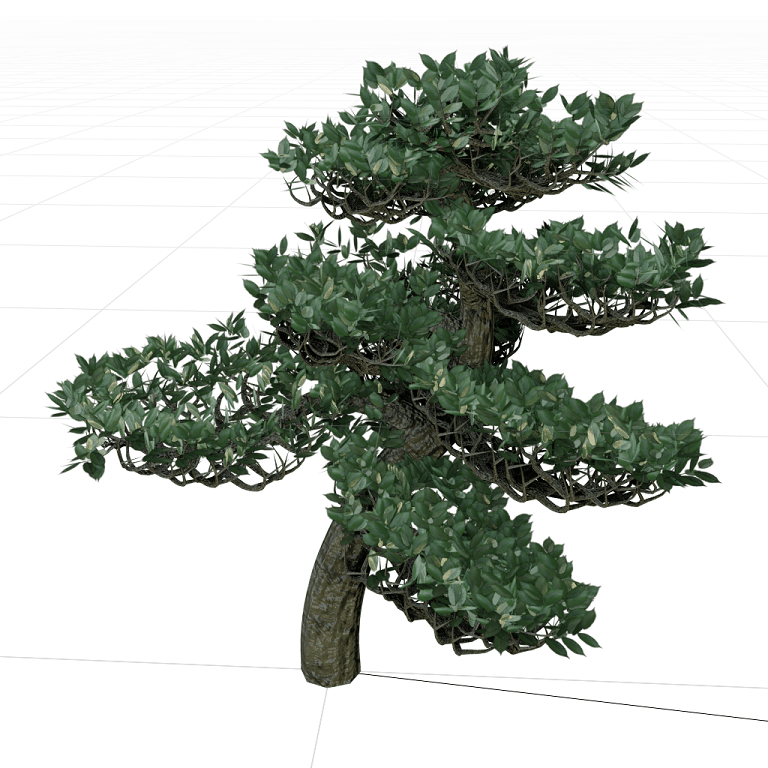}}};
     \node at (4.55,-0.78) {\frame{\includegraphics[width=0.09\textwidth]{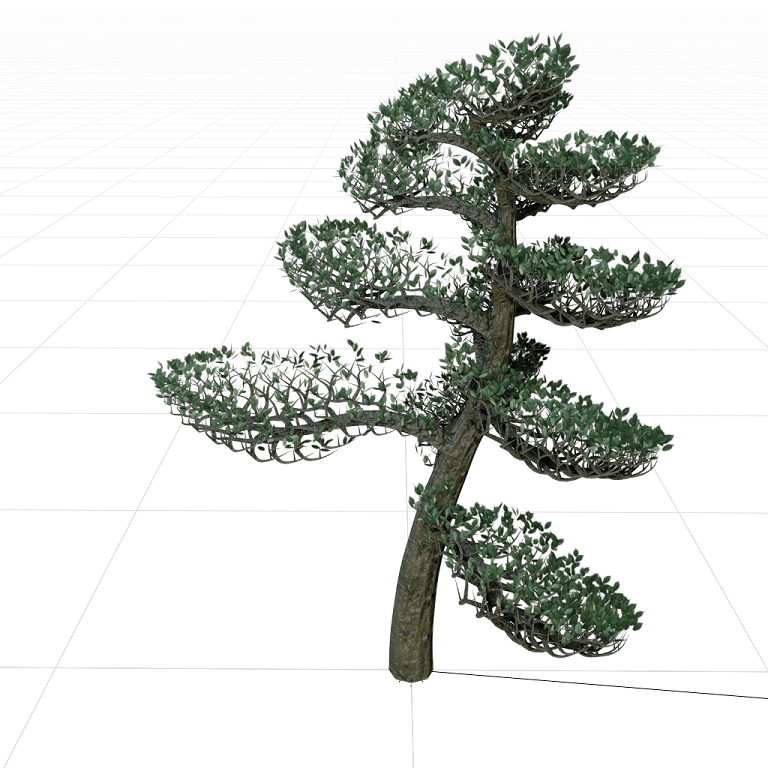}}};
     \node at (6.1,0.78) {\frame{\includegraphics[width=0.09\textwidth]{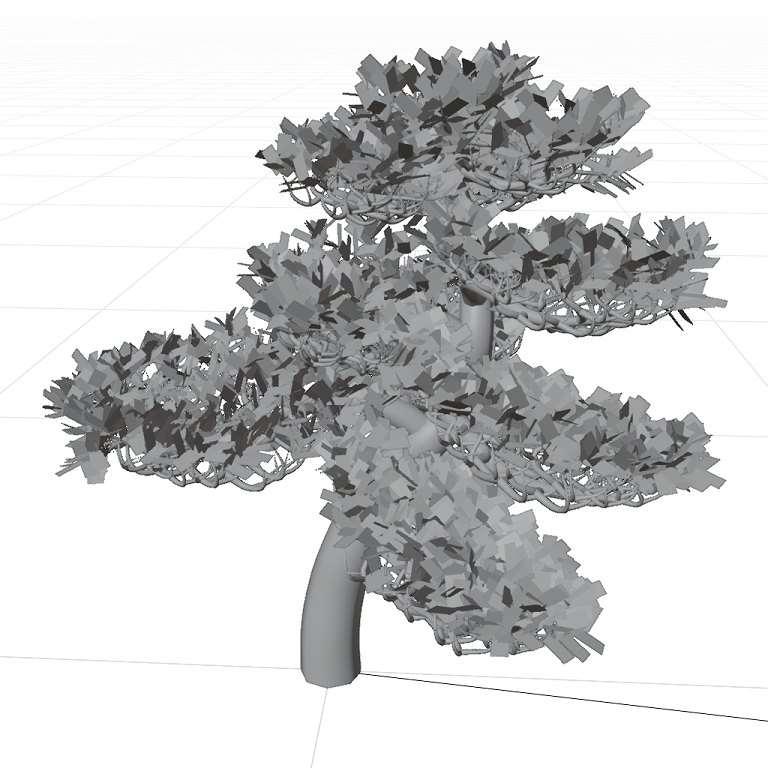}}};
     \node at (6.1,-0.78) {\frame{\includegraphics[width=0.09\textwidth]{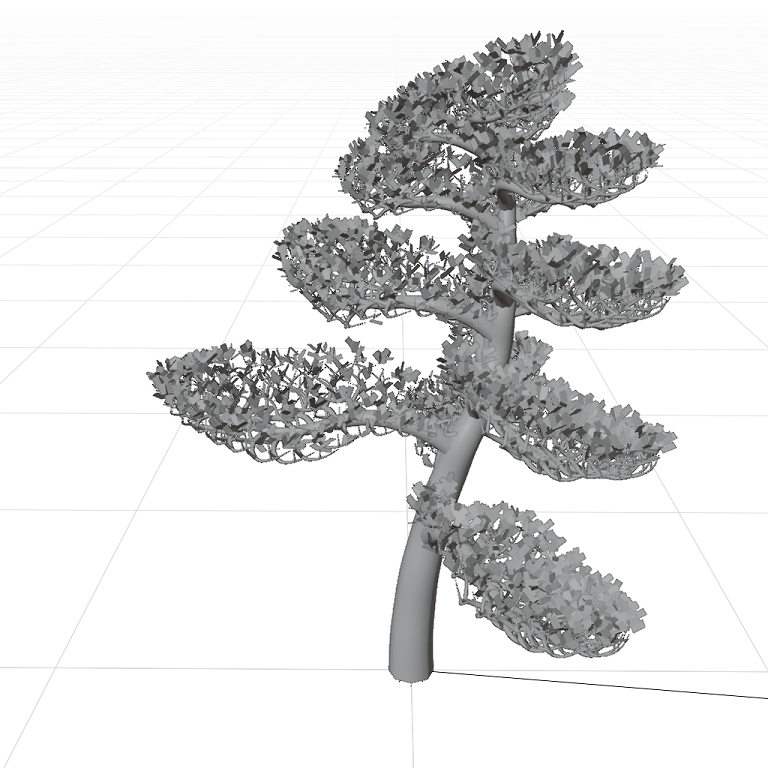}}};
    \node[text=black] at (0.55,1.7) {\scriptsize Sketch};
    \node[text=black] at (3.0,1.7) {\scriptsize Skeleton};
    \node[text=black] at (4.55,1.7) {\scriptsize Material};
    \node[text=black] at (6.1,1.7) {\scriptsize Mesh};
    
    \node[text=black] at (7.0,0.65) {\rotatebox{-90}{\tiny Ground Truth}};
    \node[text=black] at (7.0,-0.65) {\rotatebox{-90}{\tiny Prediction}};
    \end{tikzpicture}
    
    \caption{This figure shows some examples of test sketches, the corresponding GT, and predictions. The skeletons show that the trunk and branches correctly conform with the starting sketch. The reconstructed trees reported in the \quotes{Material} column show the correspondence between the foliage and the shape proposed in the input sketch. The last column shows simple reconstructed meshes without any material.}
    \label{fig:results_4}
\end{figure}
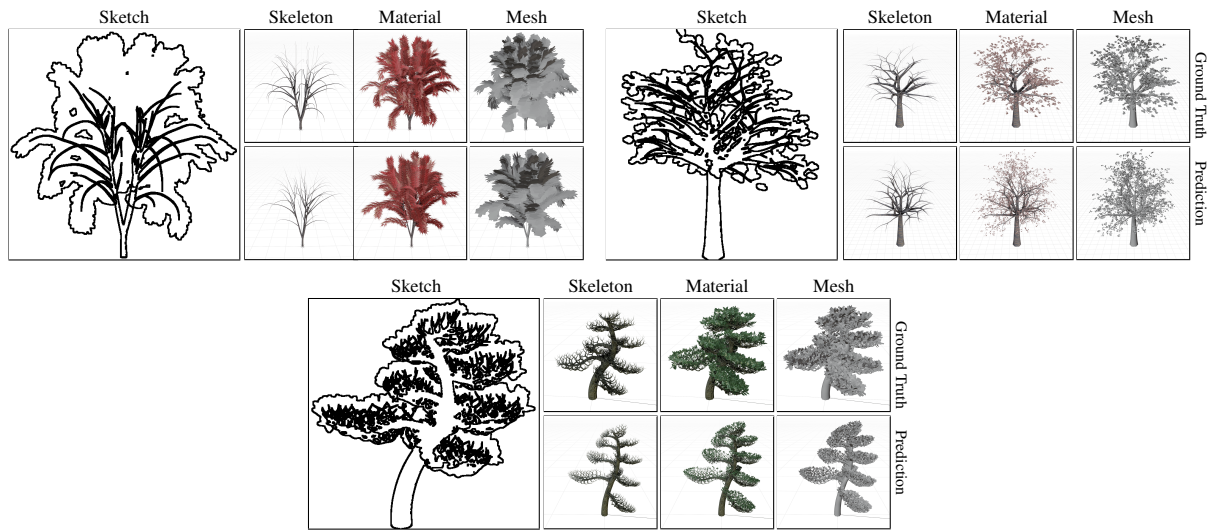

The tree sketch is an input to our TSN, which produces six $4 \times n_t$ sub-matrices containing predicted parameter values (see Sec.~\ref{sub:CNN}). These sub-matrices, which need to be de-normalized, are multiplied from their corresponding normalization matrices stored after the training phase. This operation makes it possible to adopt predicted TSN values as valid Blender inputs. The sub-matrices that are obtained from a single sketch prediction are rearranged as a Blender-like parameters dictionary. In particular, each sub-matrix column is associated with a dictionary key, representing a specific parameter. When the expected parameter is represented by a single value rather than a vector of four elements, only the first value is taken.
The dictionary is also used to identify the \revision{species} of the predicted tree (see Algorithm~\ref{alg:identify}), and assign the correct textures, based on the tree \revision{species}, to the 3D Blender mesh from the same dictionary. Specifically, Algorithm~\ref{alg:identify} takes two data structures as input. The first is a well-designed dictionary-based data structure with characteristic parameters $CP$ that are organized for each tree \revision{species}. Each element in $cp \in CP$ contains three items of information: the parameter name, its value, and the tree \revision{species}. A parameter is defined as $cp$ for a specific tree \revision{species} if it respects two properties: (\textit{i}) it is always present in all samples of that specific tree \revision{species}, and; (\textit{ii}) its value is in a specific range for that tree \revision{species}.
The second input is $P$, which is a dictionary of predicted parameters. We compare each parameter $p \in P$ with the corresponding $cp$ for each tree \revision{species} using the parameter name, and verify that the value of $p$ respects the following condition: 
\begin{equation}
p.value \in [cp.value - \epsilon, cp.value + \epsilon]
\label{eq:condition}
\end{equation}
where $[cp.value - \epsilon, cp.value + \epsilon]$ is the range of acceptable values for that tree \revision{species}. Let $eligibles$ be a dictionary in which the keys are the tree \revision{species}, and the values are counters (one for each tree \revision{species}). If the predicted parameter respects the condition~\eqref{eq:condition}, then this parameter is eligible for that tree \revision{species}, and its counter is increased by one. Finally, Algorithm~\ref{alg:identify} returns the tree \revision{species} of the predicted dictionary based on the highest percentage of counters.
\begin{algorithm}[ht!]
\SetAlgoLined
\SetKwInOut{Input}{inputs}
\SetKwInOut{Output}{outputs}
\Input{$CP$ set of characteristic parameters for each tree-\revision{species}\\
$P$ dictionary of predicted parameters}
\Output{$t$ tree-\revision{species}}

\ForEach{$p \in P$}{
    \ForEach{$cp \in CP$}{
        \If{$p.name == cp.name~\&~p.value == cp.value \pm \epsilon$}{
            $eligibles[cp.species]++$
        }
    }
}
$t = MaxPercentage(eligibles)$
 
 \caption{Detecting the tree \revision{species} from parameters}
 \label{alg:identify}
\end{algorithm}
The rearranged predicted parameters dictionary is then loaded from Blender and interpreted using a load utility method in our RT plugin. This utility makes it possible to generate the 3D tree mesh from the input dictionary, and assigns the correct texture based on the tree \revision{species} detected by Algorithm~\ref{alg:identify}. The texture is chosen from a set of predefined textures for each tree \revision{species}.
To test our approach, an appropriate dataset was defined based on our tree \revision{species}. Figure~\ref{fig:results_1} shows the input \revision{SG} sketches similar to the training set that were generated by our RT add-on, the reconstructed 3D tree mesh, and the GT. As  Figure~\ref{fig:results_1} shows, the tree structure of the test sketches was correctly predicted by our TSN, notably with respect to, for example, the number of splits for each trunk segment, curve angles, the number and distribution of branches, the ratio of the trunk to its height, and the overall height. 
As can be seen from Figures~\ref{fig:results_1} 
, our results are visually consistent with the input sketches, and are consistent with GT parameters with regard to \revision{SG} sketches.
To demonstrate the TSN generalization capability, we created also a test set containing sketch images rendered in different views respect to the $4$ views used for generating the training data. The results of this test, shown in Figures~\ref{fig:images_different_views}, confirms the generalization ability of TSN.
Figure~\ref{fig:results_4} shows some \revision{SG} sketches and the 3D meshes generated by Blender based on the predicted parameters dictionaries. Figure~\ref{fig:results_4} also shows the applied materials as a function of the tree \revision{species}, and the normal mapping of each 3D mesh.
\section{Comparisons}\label{sub:comparison}
To the best of our knowledge, this approach is the first attempt to generate 3D tree shapes from well-defined predicted parameters. Consequently, we try to do our best to compare our results with other approaches. 
Inspired by ~\cite{huang2017}, we assessed and validated our approach in a user study based on HM sketches provided by $15$ different users, as reported in Sec.~\ref{sub:user_study}. In addition, we provide quantitative comparisons using different core nets to validate our choice of EfficientNet-B7 as optimal (see Sec.~\ref{sub:backbone} and Sec.~\ref{sec:h_dist}). Finally, we provide an extensive qualitative analysis of our TSN predicted parameters, as reported in Sec.~\ref{sec:qualitative}.

\subsection{Controlled experiment}\label{sub:user_study}
 
To test our approach in a real-life context, we conducted a controlled experiment with $15$ non-experts in drawing. 
Each person was asked to provide us with a single sketch for each \revision{species} of tree. The goal was to assess the confidence level of our TSN parameters prediction using HM sketches provided by participants. 
Participants were free to choose any software package to draw their trees.
For each participant we randomly selected $5$ \revision{SG} sketches, one for each tree \revision{species}. The participants were asked to look at each \revision{SG} sketch for two minutes and then draw the tree sketch according to their own style. In this way, the participants can understand what \revision{species} of trees the system could handle without being constrained too much on the drawings.

\begin{figure*}[ht!]
 \begin{tabular}{c c c c c c} 
    \rotatebox{90}{\parbox{7em}{\centering \small HM Input}}
    & \frame{\includegraphics[width=0.18\textwidth]{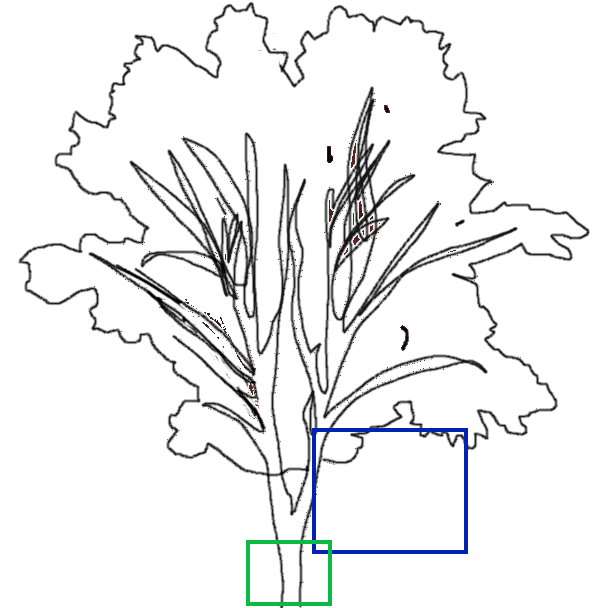}}
    & \frame{\includegraphics[width=0.18\textwidth]{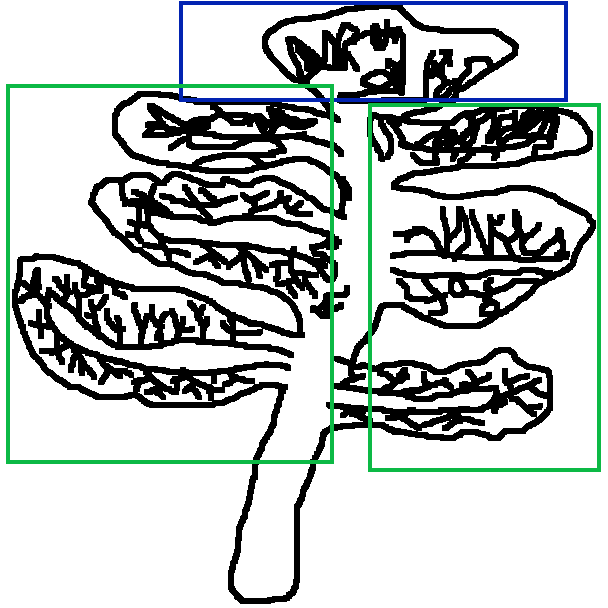}}
    & \frame{\includegraphics[width=0.18\textwidth]{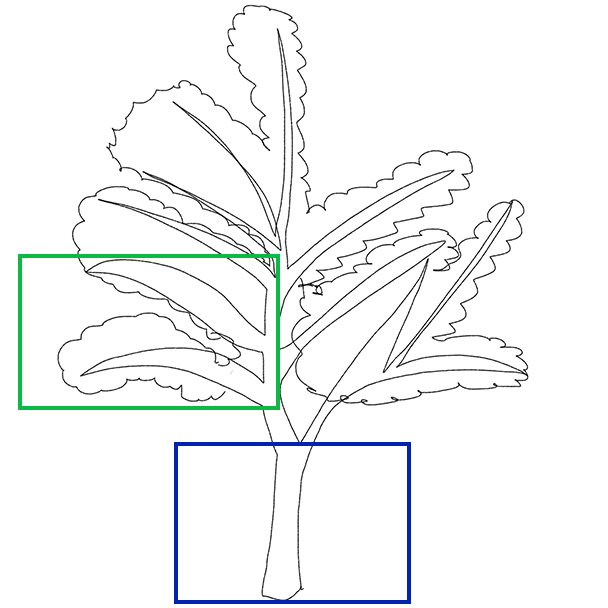}}
    & \frame{\includegraphics[width=0.18\textwidth]{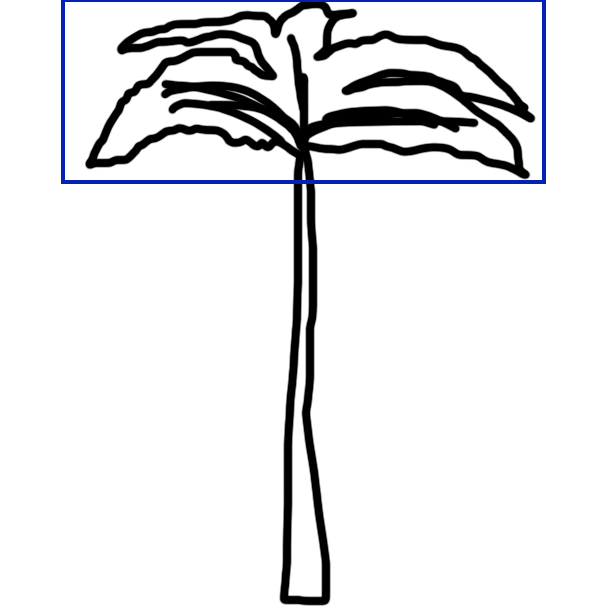}}
    & \frame{\includegraphics[width=0.18\textwidth]{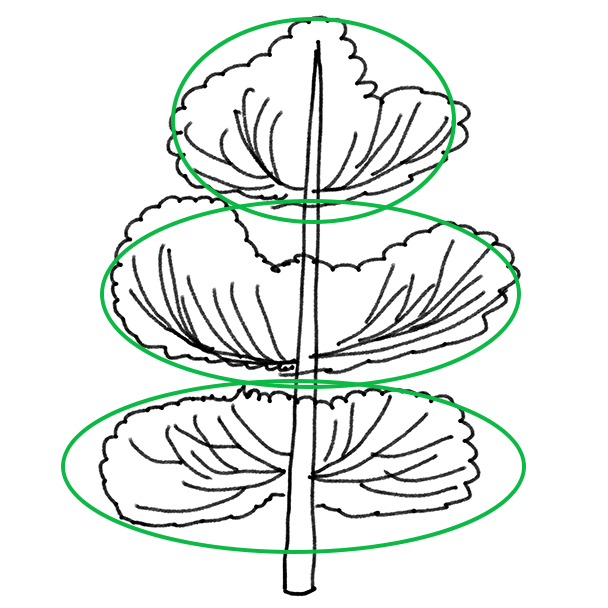}}
    \\ 
    \rotatebox{90}{\parbox{7em}{\centering \small Reconstructed}}
    & \frame{\includegraphics[width=0.18\textwidth]{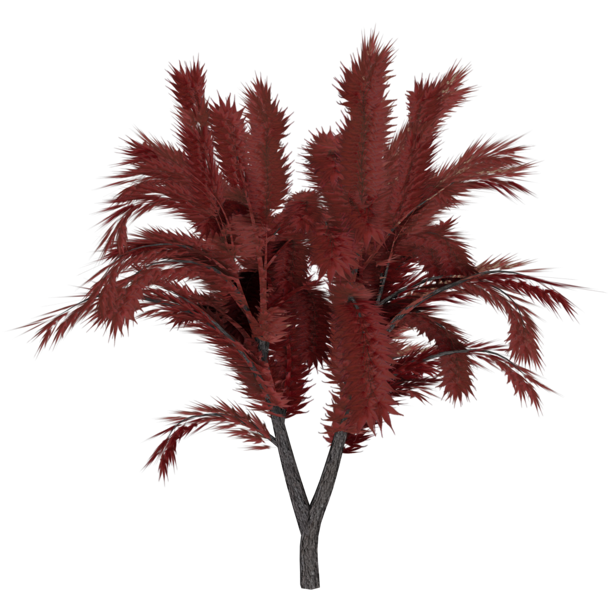}}
    & \frame{\includegraphics[width=0.18\textwidth]{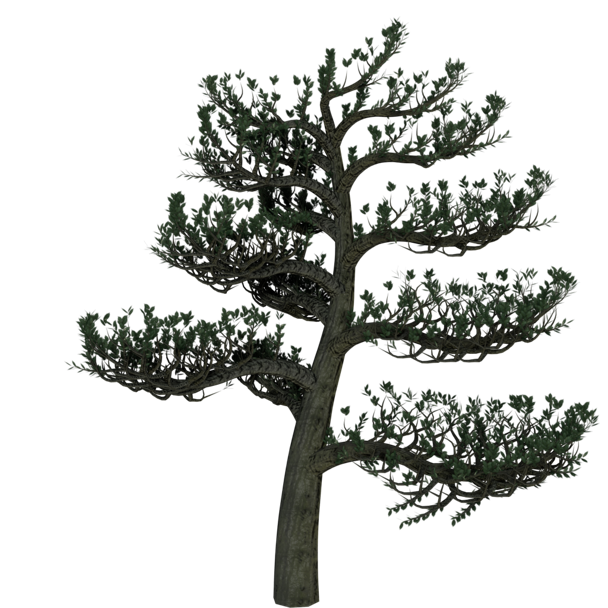}}
    & \frame{\includegraphics[width=0.18\textwidth]{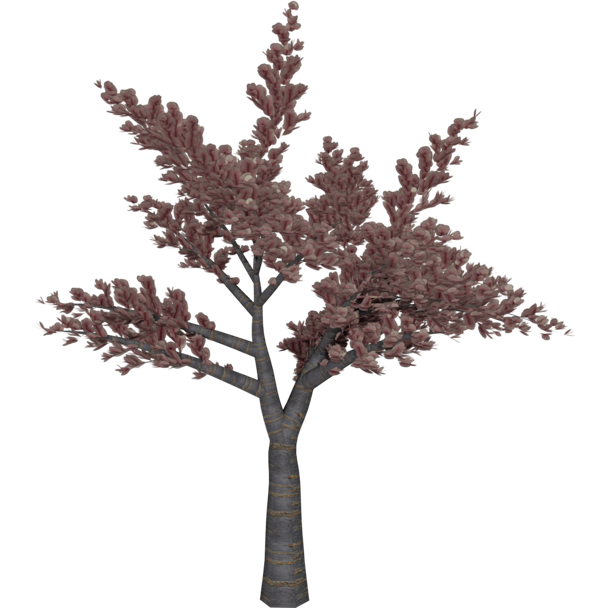}}
    & \frame{\includegraphics[width=0.18\textwidth]{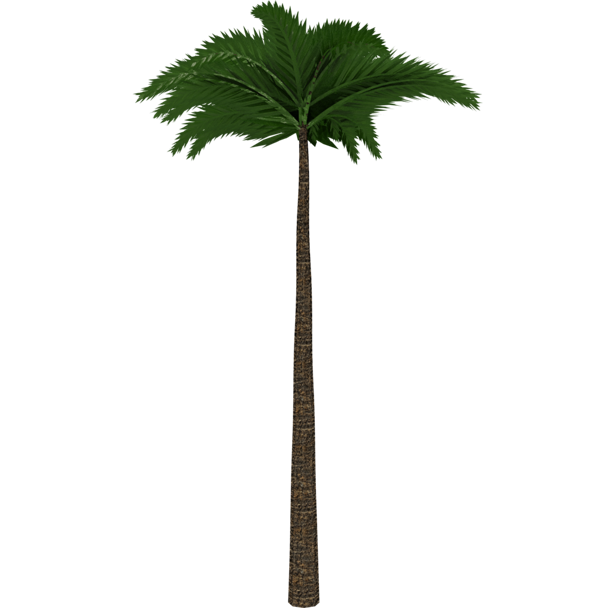}}
    & \frame{\includegraphics[width=0.18\textwidth]{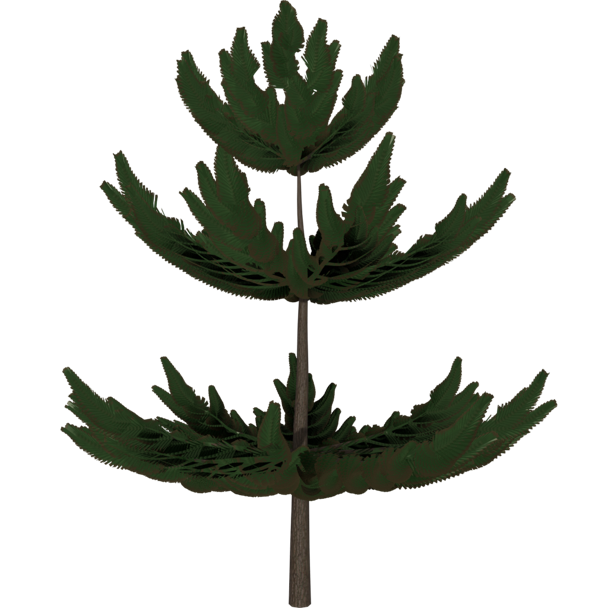}}
    \\ 
    \rotatebox{90}{\parbox{7em}{\centering \small Reconstructed SG}}
    & \frame{\includegraphics[width=0.18\textwidth]{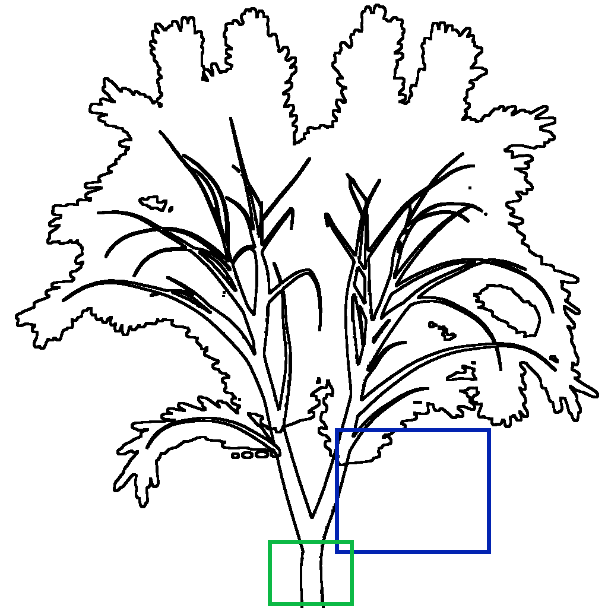}}
    & \frame{\includegraphics[width=0.18\textwidth]{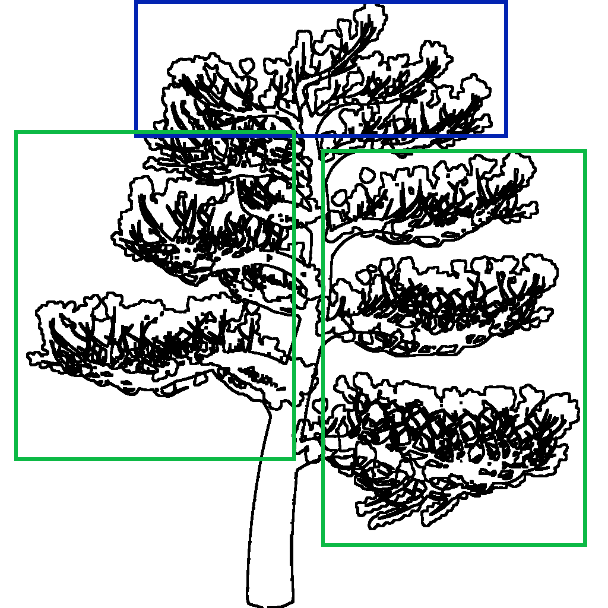}}
    & \frame{\includegraphics[width=0.18\textwidth]{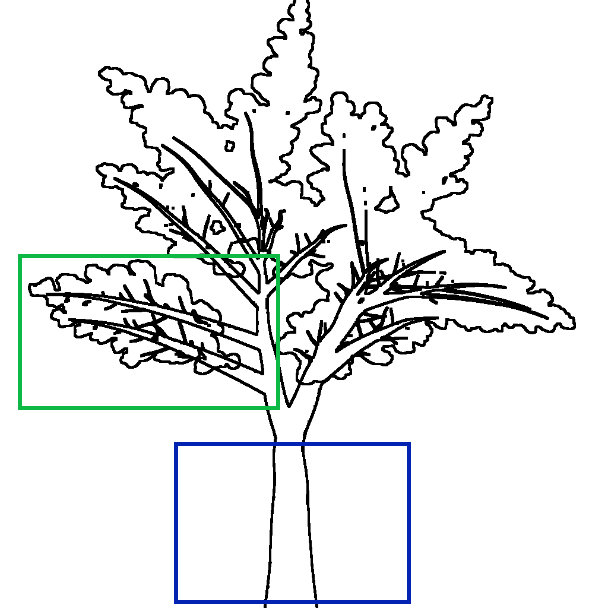}}
    & \frame{\includegraphics[width=0.18\textwidth]{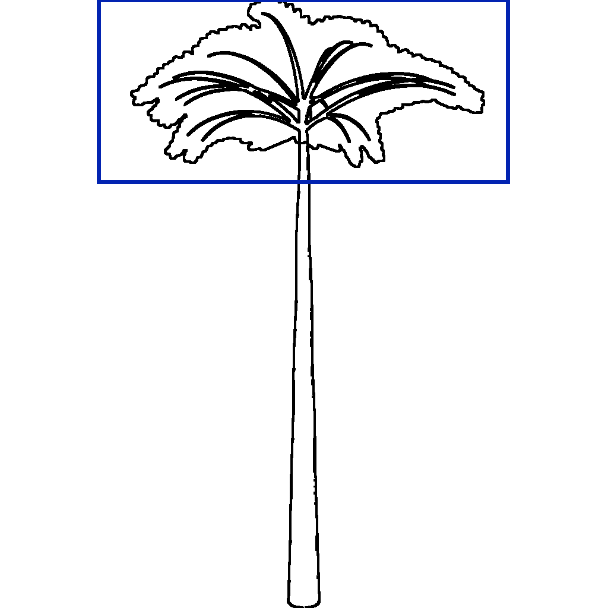}}
    & \frame{\includegraphics[width=0.18\textwidth]{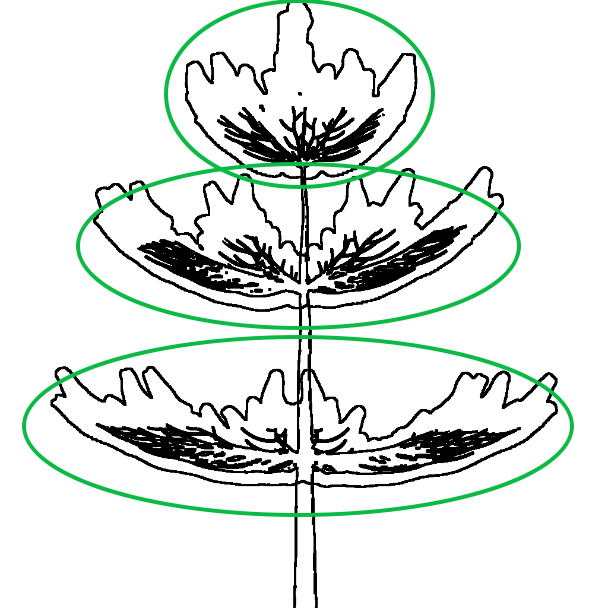}}
    \\
    \rotatebox{90}{\parbox{7em}{\centering \small NNS}}
    & \frame{\includegraphics[width=0.18\textwidth]{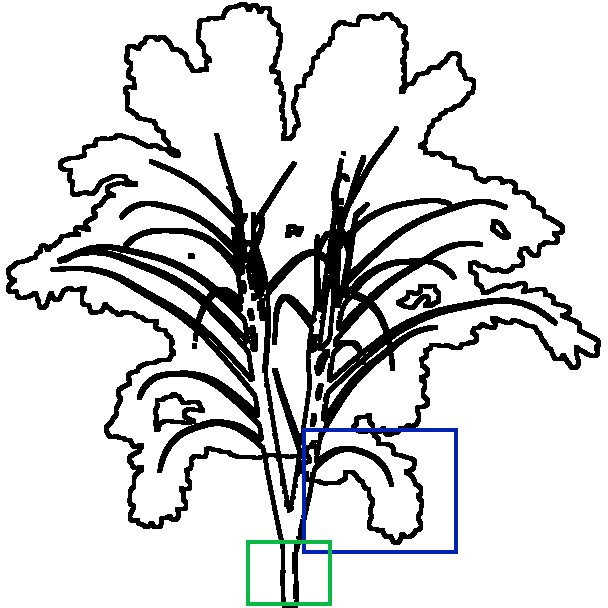}}
    & \frame{\includegraphics[width=0.18\textwidth]{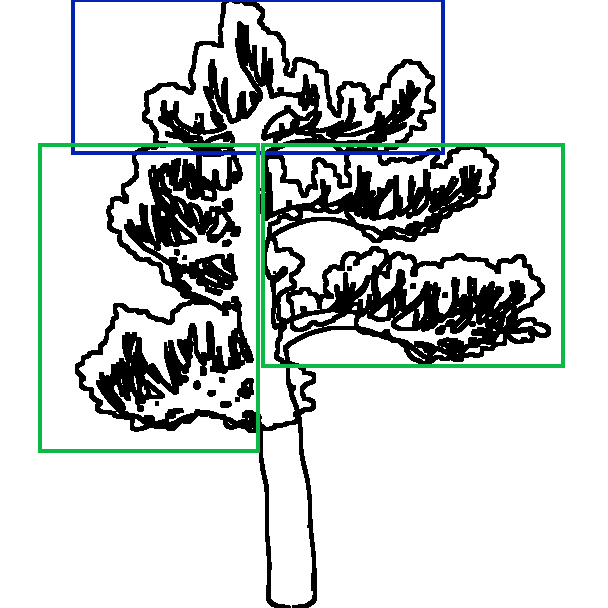}}
    & \frame{\includegraphics[width=0.18\textwidth]{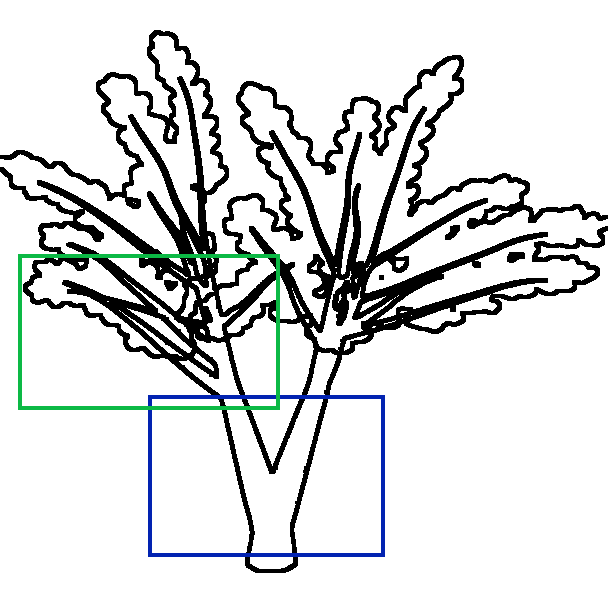}}
    & \frame{\includegraphics[width=0.18\textwidth]{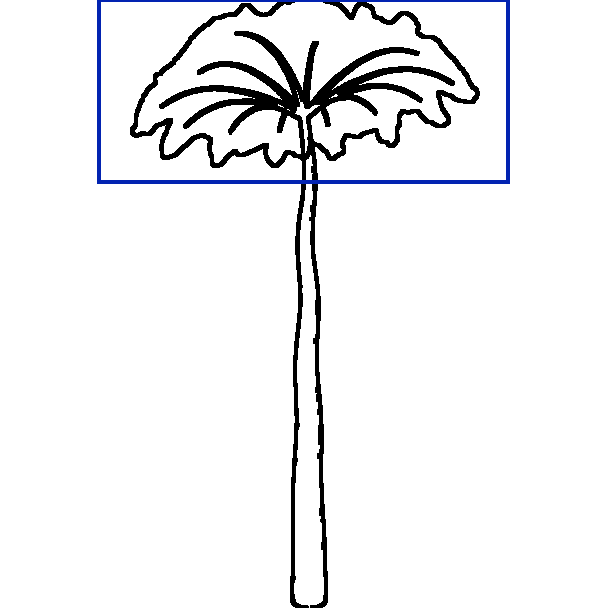}}
    & \frame{\includegraphics[width=0.18\textwidth]{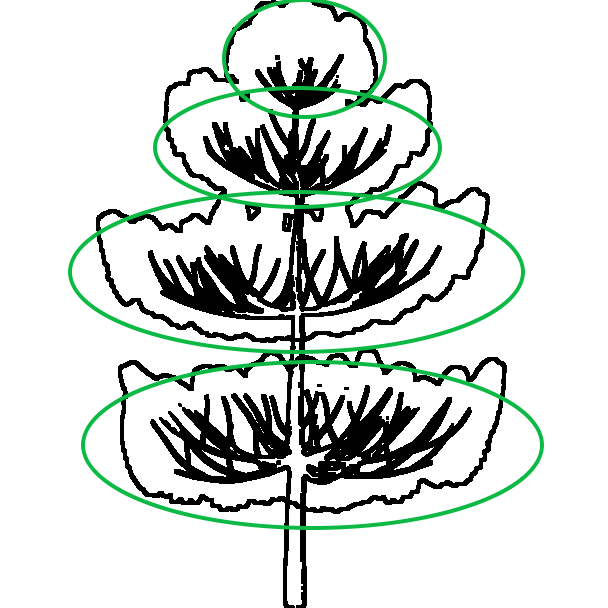}}
    \\
	\end{tabular}
	\caption{\revision{Generalization ability of TSN. The first row shows the HM sketches. The second row show the 3D model generated by the TSN from the HM sketch. The third row contain the SG sketches obtained from the reconstructed 3D model. The last row shows the nearest sketch retrieved from the training set.}}
	\label{fig:nearest_neighbours}
\end{figure*}

\revision{To have another significant proof of the generalization ability of TSN, we adopt the approach of~\cite{garmentdesign_Wang_SA18}, which consists in finding the Nearest Neighbour Sketch (NNS) of the SG sketch extracted from the 3D tree model generated by TSN, given an HM sketch as input.  The NNS is obtained from the training set using the TSN features of sketches. Figure~\ref{fig:nearest_neighbours} shows the results of the comparison between the HM sketch, the reconstructed SG sketch described above, and the NNS. The first column of Figure~\ref{fig:nearest_neighbours} shows that the reconstructed SG resembles the HM sketch more than the NNS. This can be noticed in the width of the trunk (green box) and in the presence of an additional branch (blue box) in NNS that in the HM sketch and in the reconstructed SG is not present. The second column shows an NNS with a different orientation and trunk bending compared to the HM and the SG reconstructed sketch. Furthermore, in the NNS we can count one less side branch (green box), both for the left and the right side, than the HM and the SG reconstructed sketch. The blue box locates the tip of the tree. In the third column, it can be noticed an NNS quite different from the HM sketch. By contrast, the reconstructed SG sketch looks more like the HM one both in the shape of the trunk (blue box) and in the bending (green box) and in the general arrangement of branches. Also in the fourth column the bending and the general arrangement (blue box) of branches is more similar between the HM and the SG reconstructed sketches rather than between the HM sketch and the NNS. In the last column, the NNS has an extra \textit{Branch Whorl} compared to the reconstructed SG and the HM sketch. In addition, the general branch arrangement of the HM sketch is more similar to the reconstructed SG sketch than the NNS.}

\subsection{Core net testing}\label{sub:backbone}
The choice of the EfficientNet-B7 core can greatly influence the outcome of the approach. Therefore, we ran tests with different core models, and assessed their performance.
In particular, for each tested core model, we analyzed the $1-RMSE$ for each specific branch, and the overall average for all branches. The first comparison was between 
% Inception V3 
EfficientNet-B7 and VGG-16 with skip connections (see Sec.~\ref{subsub:over_under}). The second comparison was performed with ResNet and Inception V3, which 
% is seen as the state-of-the-art 
are used as core models in several computer vision tasks~\cite{szegedy2016rethinking}. 
% We 
For ResNet, we used variant $50$ because its depth is sufficient to extract the essential low-level features needed for our task. For the test made with Inception V3 we used the first $2$ convolutional blocks, and the first $5$ Inception Modules, to reduce the DNN complexity and consequently the overfitting problem caused by the overparametrization~\cite{salman2019overfitting}.
% Finally, we
We also compared
% Inception V3
EfficientNet-B7 with AlexNet~\cite{krizhevsky2012imagenet} to evaluate the validity of procedural modelling-based approaches for 3D mesh generation that are proposed in ~\cite{huang2017}. For the sake of completeness, we finally compared EfficientNet-B7 with another SOTA network, called CoAtNet.
Each test was performed using $30$ sketches, of which $15$ were \revision{SG}, and $15$ were HM. For each core model, we trained our architecture for 2k epochs, and the assessment was based on the following factors: \textit{(i)} possible overfitting of branches (low and high) based on the validation curve accuracy with respect to the training curve accuracy; \textit{(ii)} the accuracy of each branch, evaluated by observing the validation curve, and; \textit{(iii)} the generalization level, evaluated by observing the testing curve.

Table~\ref{tab:comp_quant_shm} shows the results of the comparison with the \revision{SG}-based test set. These sketches were obtained using our pipeline, as described in Section~\ref{sub:dataset_creation}. The \revision{SG} tests were run using all parameters and the prediction dictionaries were compared with GT parameters using $1-RMSE$. EfficientNet-B7 performs better with respect to the other core models. However, CoAtNet does perform slightly better on the $[-360, 360]$ branch. Inception V3 also performs well, but there are no branches where Inception V3 outperforms EfficientNet-B7. The overall performance suggest that EfficientNet-B7 perform better than the other networks.

\begin{table*}[htbp]
\centering
\begin{tabular}{lccccccc} 
\toprule
\multirow{2}{*}{Core Nets} & \multicolumn{7}{c}{1-RMSE} \\ 
\cmidrule(lr){2-8}
 & $[-inf, inf]$ & $[-360, 360]$ & $[0, 1]$ & $[0, inf]$ & $[min, max]$ & $[-1 ,1]$ & Overall\\ 
\hline
VGG-16 & 0.8022 & 0.9552 & 0.7988 & 0.9454 & 0.9772 & 0.9899 & 0.9115\\
ResNet-50 & 0.7885 & 0.9568 & 0.7803 & 0.937 & 0.9579 & 0.9646 & 0.8975\\
AlexNet & 0.7791 & 0.9552 & 0.7963 & 0.9688 & 0.9792 & 0.9895 & 0.9113\\
Inception V3 & 0.7957 & 0.959 & 0.802 & 0.9657 & 0.9811 & 0.988 & 0.9152\\
CoAtNet & 0.8073 & \textbf{0.9607} & 0.8047 & 0.9556 & 0.9769 & 0.9848 & 0.915\\
\textbf{EfficientNet-B7 (our)} & \textbf{0.8093} & 0.9595 & \textbf{0.8129} & \textbf{0.9798} & \textbf{0.9936} & \textbf{0.9964} & \textbf{0.9253} \\
\bottomrule
\end{tabular}
\caption{Core net comparisons of \revision{SG} sketches. Observations of overall $1-RMSE$ suggests that EfficientNet-B7 has an higher accuracy. However, CoAtNet performs slightly better on the $[-360, 360]$ branch. The best values are highlighted in bold.}
\label{tab:comp_quant_shm}
\end{table*}

For HM testing, we prepared some hand-made sketches inspired by those generated by the \revision{SG} (see Sec.~\ref{sub:dataset_creation} and Figure~\ref{fig:sketchPipeline}) approach and considered their corresponding parameters as GT. The comparisons reported in Table~\ref{tab:comp_quant_hm} show that 
both SOTA networks (Efficientnet-B7 and CoAtNet) and Inception V3 achieved better results with respect to the other three core models for HM sketches with respect to training sketches, demonstrating a higher level of generalization. In particular, we can notice that EfficientNet-B7 has an higher accuracy respect to CoAtNet and Inception V3.
Furthermore, a comparison of the results reported in Table~\ref{tab:comp_quant_shm} and Table~\ref{tab:comp_quant_hm}, which have similar features, and an analysis of the training report, identified that AlexNet is slightly overfitted on $[0, inf]$ and $[min, max]$. By applying the same analysis to the other core models, we identified that for VGG-16 with skip connections, the branches $[-inf, inf]$ and $[0, inf]$ are slightly overfitted, while for ResNet-50, Inception V3, EfficientNet, and CoAtNet
there is no evidence of overfitting. However, EfficientNet-B7 was selected as the best core model due to its higher accuracy compared to the other DNNs and its generalization ability with HM sketches significantly different from training data. 

\begin{table*}[htbp]
\centering
\begin{tabular}{lccccccc} 
\toprule
\multirow{2}{*}{Core Nets} & \multicolumn{7}{c}{1-RMSE} \\ 
\cmidrule(lr){2-8}
 & $[-inf, inf]$ & $[-360, 360]$ & $[0, 1]$ & $[0, inf]$ & $[min, max]$ & $[-1 ,1]$ & Overall\\ 
\hline
VGG-16 & 0.7577 & 0.9485 & 0.7931 & 0.9176 & 0.931 & 0.9795 & 0.8879\\
ResNet-50 & 0.7585 & 0.9605 & 0.8278 & 0.9356 & 0.952 & 0.9614 & 0.8993\\
AlexNet & 0.75 & 0.955 & 0.8057 & 0.9136 & 0.9124 & 0.975 & 0.8853\\
Inception V3  & \textbf{0.7873} & 0.9618 & 0.8122 & 0.9607 & 0.9746 & 0.9862 & 0.9138\\
CoAtNet & 0.7712 & \textbf{0.9646} & 0.8192 & 0.9455 & 0.9608 & 0.9815 & 0.9071\\
\textbf{EfficientNet-B7 (our)} & 0.7688 & 0.9628 & \textbf{0.8658} & \textbf{0.979} & \textbf{0.9883} & \textbf{0.9957} & \textbf{0.9267} \\
\bottomrule
\end{tabular}
\caption{The results of HM sketches clearly show the difference between the different core nets. Based on overall performance, EfficientNet-B7 has an higher accuracy. Despite the differences between hand-drawn sketches, EfficientNet-B7 is demonstrated to have a greater ability to generalize. The best values are highlighted in bold.}
\label{tab:comp_quant_hm}
\end{table*}
\subsection{Hausdorff Distance}\label{sec:h_dist}
Since in the previous section, we provided a quantitative analysis of the DNNs performances in terms of accuracy, in this section we assess the quality of the reconstructed 3D trees. For this purpose, we used a metric called Hausdorff Distance (HDD)~\cite{metro_meshlab}. 
In particular, we compare the distances calculated for EfficientNet-B7, CoatNet, and Inception V3, because they represent the best performing DNNs, as reported in the previous section. 
The HDD represents the maximum distance among a set of distances between two meshes. This set contains all minimum distances between the vertices of the first and second mesh. In our case, we calculate, for each DNN, the HDD between $15$ tree meshes of different \revision{species} reconstructed from the DNN predictions and their ground-truth meshes. The $15$ ground-truth trees are the same for all DNN to grant the results coherence. Table \ref{tab:hausdorff_distance} shows, for each network, the average of the calculated HDDs. From the results in Table \ref{tab:hausdorff_distance}, it can be seen that EfficientNet-B7 is still the better choice because its HDD tends more to $0$. It means that the meshes resulting from EfficientNet-B7 are more similar to their ground-truth than those resulting from Inception V3 or CoAtNet. Additionally, our approach provides better results in calculating this metric than the SOTA \cite{LIU2021101115}.
\begin{table}[htbp]
\centering
\begin{tabular}{lc} 
\toprule
Core Nets & Hausdorff Distance \\ 
\hline
Inception V3  & 0.047\\
CoAtNet & 0.041\\
\textbf{EfficientNet-B7 (our)} & \textbf{0.031} \\
\bottomrule
\end{tabular}
\caption{Hausdorff distances of the best performing DNNs. Based on the results, the meshes resulting from EfficientNet-B7 are more similar to their ground-truth. The best values are highlighted in bold.}
\label{tab:hausdorff_distance}
\end{table}

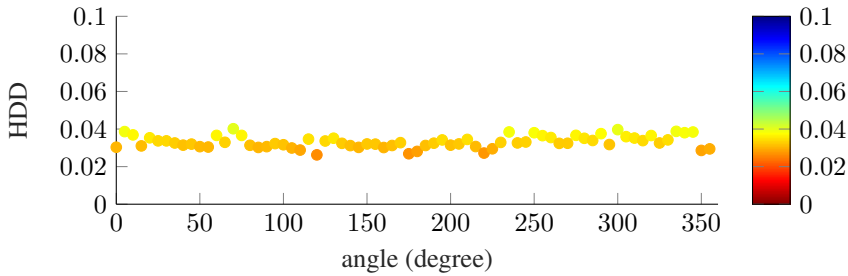
\begin{figure*}[ht!]
\centering
% \documentclass[border={-4pt 0pt 7pt 0pt}]{standalone}
% \usepackage{pgfplots}
% \pgfplotsset{compat=1.18}

% \begin{document}

% \pgfplotsset{
% \pgfplots\colormap={reversejet}{rgb255(0cm)=(128,0,0) rgb255(1cm)=(255,0,0)
% rgb255(3cm)=(255,255,0) rgb255(5cm)=(0,255,255) rgb255(7cm)=(0,0,255)
% rgb255(8cm)=(0,0,128)}
% }

% \begin{figure}
\begin{tikzpicture}
\centering
\begin{axis}[%
width=8cm,
height=2.5cm,
at={(0cm,0cm)},
scale only axis,
point meta min=0,
point meta max=0.1,
colormap={reversejet}{rgb255(0cm)=(128,0,0) rgb255(1cm)=(255,0,0)
            rgb255(3cm)=(255,255,0) rgb255(5cm)=(0,255,255) rgb255(7cm)=(0,0,255)
            rgb255(8cm)=(0,0,128)},
xmin=0,
xmax=360,
xlabel style={font=\color{white!15!black}},
xlabel={angle (degree)},
ymin=0,
ymax=0.1,
ylabel style={font=\color{white!15!black}},
ylabel={HDD},
ylabel style = {yshift = 0.2cm},
yticklabels={,$0$, $0.02$, $0.04$, $0.06$, $0.08$,$0.1$},
axis background/.style={fill=white},
axis x line*=bottom,
axis y line*=left,
colorbar,
colorbar style = {
yticklabels={,$0$, $0.02$, $0.04$, $0.06$, $0.08$,$0.1$}
}
]
\addplot[scatter, only marks, mark=*, mark size=2pt, scatter src=explicit, scatter/use mapped color={mark options={}, draw=mapped color, fill=mapped color}, forget plot] table[row sep=crcr, meta=color]{%
x	y	color\\
0	0.0303289257553697	0.0303289257553697\\
5	0.038640302208101	0.038640302208101\\
10	0.0368833716313771	0.0368833716313771\\
15	0.0310275893237375	0.0310275893237375\\
20	0.0353375979247556	0.0353375979247556\\
25	0.0337116207288027	0.0337116207288027\\
30	0.0337073865694911	0.0337073865694911\\
35	0.0325788685287423	0.0325788685287423\\
40	0.0314336266499908	0.0314336266499908\\
45	0.0319865827566874	0.0319865827566874\\
50	0.030628094210132	0.030628094210132\\
55	0.0304181813971469	0.0304181813971469\\
60	0.0366219859551626	0.0366219859551626\\
65	0.0329464955914981	0.0329464955914981\\
70	0.0401621114460414	0.0401621114460414\\
75	0.0366103010941242	0.0366103010941242\\
80	0.0313611575947112	0.0313611575947112\\
85	0.03021780908359	0.03021780908359\\
90	0.0306393643854646	0.0306393643854646\\
95	0.032256398556789	0.032256398556789\\
100	0.0316395835795984	0.0316395835795984\\
105	0.0299150803337879	0.0299150803337879\\
110	0.0288157976352962	0.0288157976352962\\
115	0.0346408468213861	0.0346408468213861\\
120	0.0262591975106985	0.0262591975106985\\
125	0.0336391816836729	0.0336391816836729\\
130	0.0351199752314776	0.0351199752314776\\
135	0.0324007582899165	0.0324007582899165\\
140	0.0311928743158341	0.0311928743158341\\
145	0.0302452874679884	0.0302452874679884\\
150	0.0320630977257938	0.0320630977257938\\
155	0.0319688522145701	0.0319688522145701\\
160	0.0301658887963097	0.0301658887963097\\
165	0.031249148803025	0.031249148803025\\
170	0.0327604326629859	0.0327604326629859\\
175	0.0268536603827086	0.0268536603827086\\
180	0.0281024836574546	0.0281024836574546\\
185	0.0312468734880341	0.0312468734880341\\
190	0.0324812509679637	0.0324812509679637\\
195	0.0342080486261332	0.0342080486261332\\
200	0.0314288413952018	0.0314288413952018\\
205	0.0320834991712397	0.0320834991712397\\
210	0.03448130195248	0.03448130195248\\
215	0.0306895413558634	0.0306895413558634\\
220	0.0272157874034721	0.0272157874034721\\
225	0.0294982147998622	0.0294982147998622\\
230	0.0328898718952918	0.0328898718952918\\
235	0.038451410140925	0.038451410140925\\
240	0.0326167822606709	0.0326167822606709\\
245	0.0331076882904455	0.0331076882904455\\
250	0.0380347956009791	0.0380347956009791\\
255	0.0364823678256972	0.0364823678256972\\
260	0.0354832289969547	0.0354832289969547\\
265	0.032383417879746	0.032383417879746\\
270	0.0324402489188052	0.0324402489188052\\
275	0.0366520452124404	0.0366520452124404\\
280	0.0350774044699452	0.0350774044699452\\
285	0.0339112182348567	0.0339112182348567\\
290	0.0375537219875441	0.0375537219875441\\
295	0.0317524108299144	0.0317524108299144\\
300	0.039687599142622	0.039687599142622\\
305	0.0359455626359623	0.0359455626359623\\
310	0.0352034089042013	0.0352034089042013\\
315	0.0338357645337737	0.0338357645337737\\
320	0.0365437876894743	0.0365437876894743\\
325	0.0325816567444745	0.0325816567444745\\
330	0.0342333144440357	0.0342333144440357\\
335	0.0387712392196994	0.0387712392196994\\
340	0.0380520214269414	0.0380520214269414\\
345	0.038420366005096	0.038420366005096\\
350	0.028619842270145	0.028619842270145\\
355	0.0294382899871823	0.0294382899871823\\
};
\end{axis}
\end{tikzpicture}%
% \end{figure}
% \end{document}
\caption{\revision{HDD distances for each degree of rotation. Because the HDD values are near to $0$ we narrowed the $y$-axis range to $[0, 0.1]$. In this way, the minimal differences between trees are highlighted.}}
\label{fig:360_hdd_graph}
\end{figure*}

\begin{figure*}[ht!]
\centering
\begin{tikzpicture}
\node (tab1) {
 \begin{tabular}{c c c c c c } 
    &\small SG Input & \small Reconstructed & \small Ground Truth & \small GT HDD & \small HDD Avg
    % \parbox[][][t]{0.18\textwidth}{HDD Mean \\ Value}
    \\
    \rotatebox{90}{\parbox{7em}{\centering \small Worst Case}}
    & \frame{\includegraphics[width=0.18\textwidth]{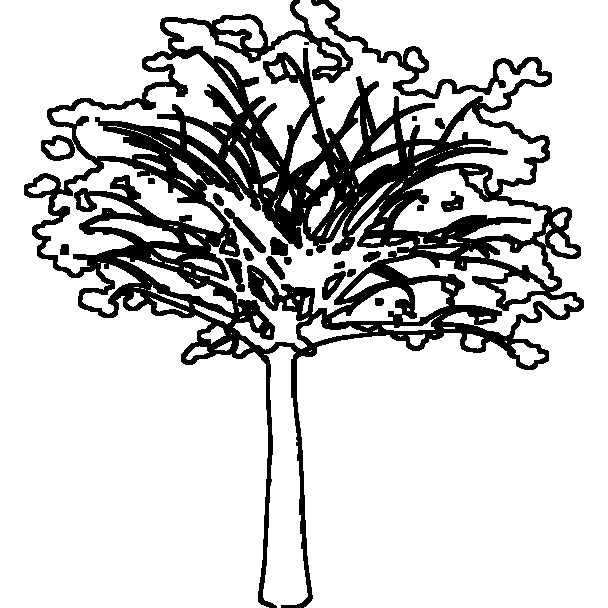}}
    & \frame{\includegraphics[width=0.18\textwidth]{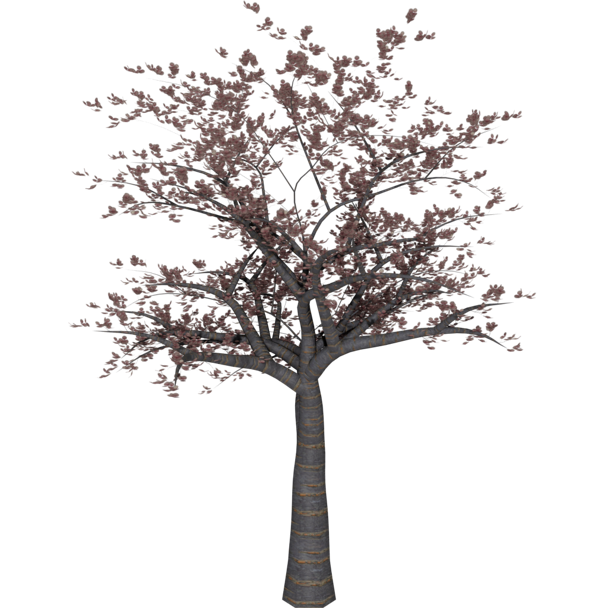}}
    & \frame{\includegraphics[width=0.18\textwidth]{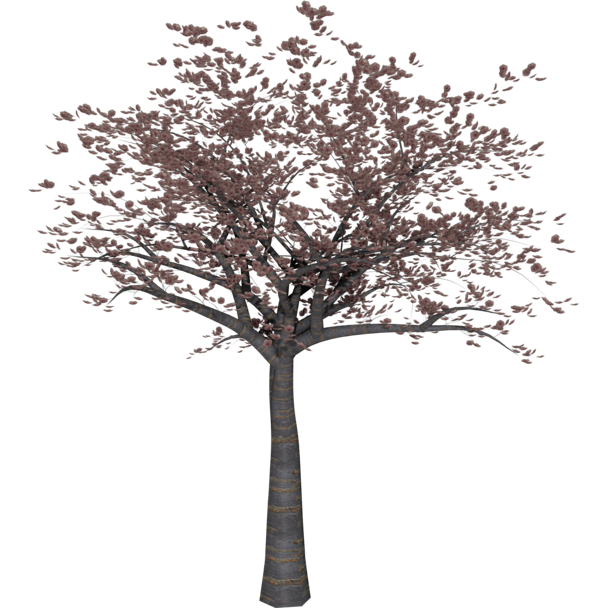}}
    & \frame{\includegraphics[width=0.18\textwidth]{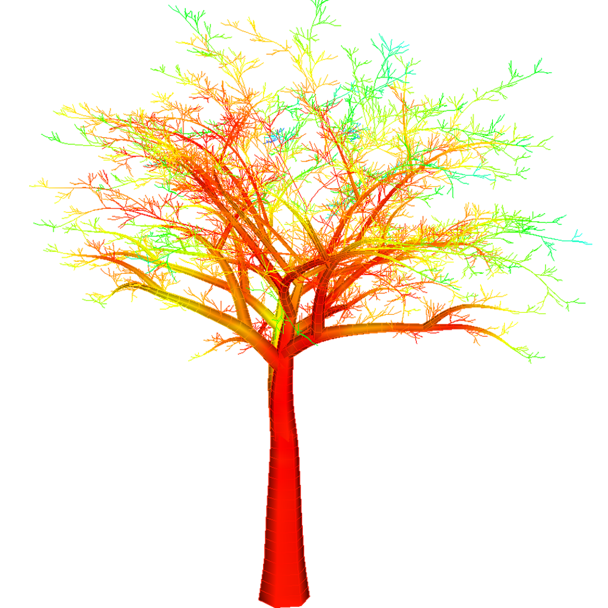}}
    & \frame{\parbox[][0.18\textwidth][c]{3em}{\centering $0.040$}}
    \\
    \rotatebox{90}{\parbox{7em}{\centering \small Median Case}}
    & \frame{\includegraphics[width=0.18\textwidth]{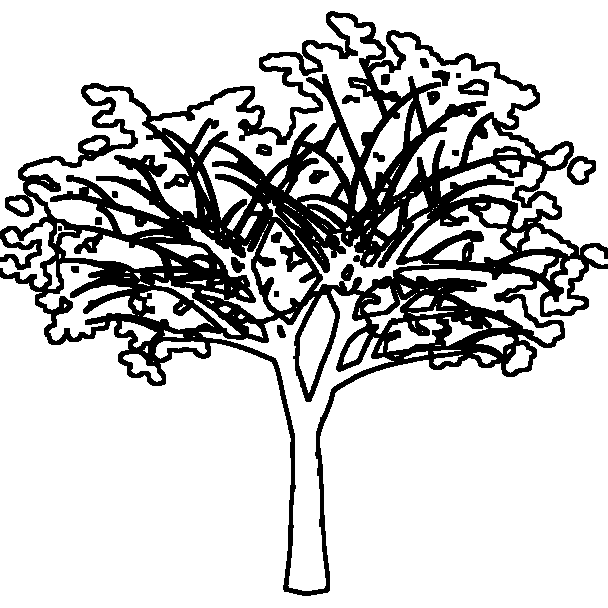}}
    & \frame{\includegraphics[width=0.18\textwidth]{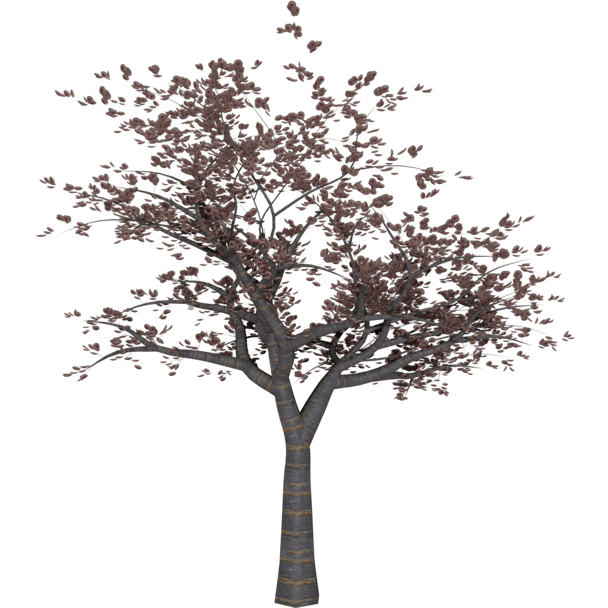}}
    & \frame{\includegraphics[width=0.18\textwidth]{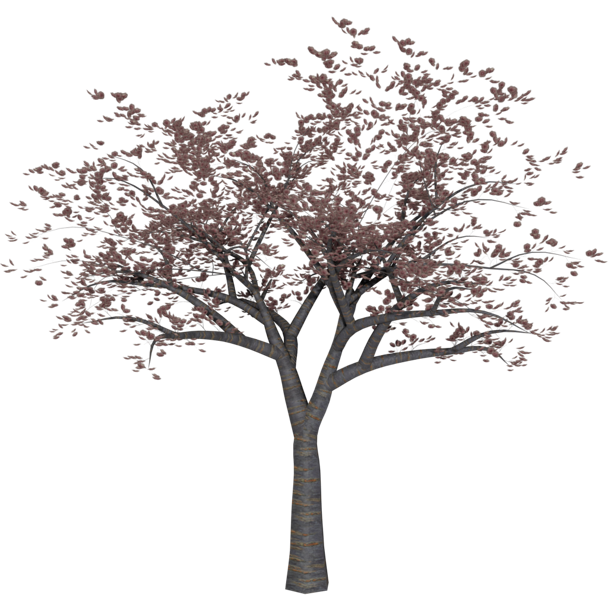}}
    & \frame{\includegraphics[width=0.18\textwidth]{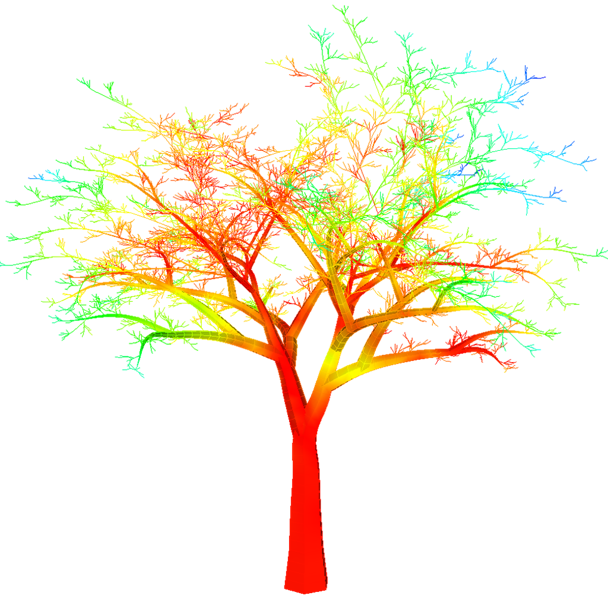}}
    & \frame{\parbox[][0.18\textwidth][c]{3em}{\centering $0.033$}} 
    \\
    \rotatebox{90}{\parbox{7em}{\centering \small Better Case}}
    & \frame{\includegraphics[width=0.18\textwidth]{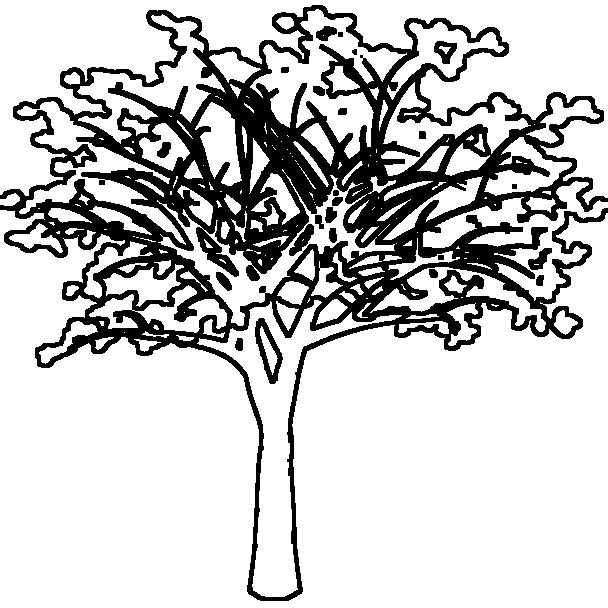}}
    & \frame{\includegraphics[width=0.18\textwidth]{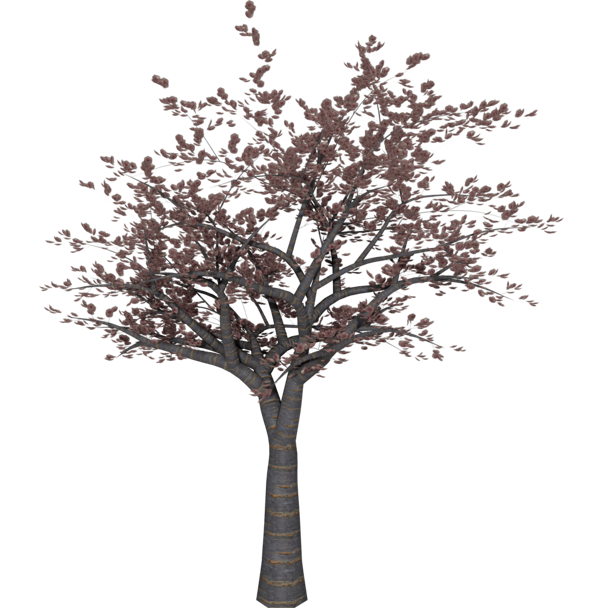}}
    & \frame{\includegraphics[width=0.18\textwidth]{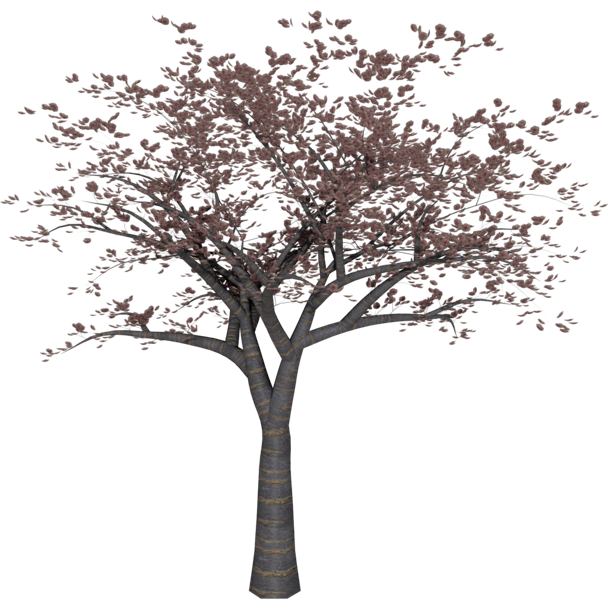}}
    & \frame{\includegraphics[width=0.18\textwidth]{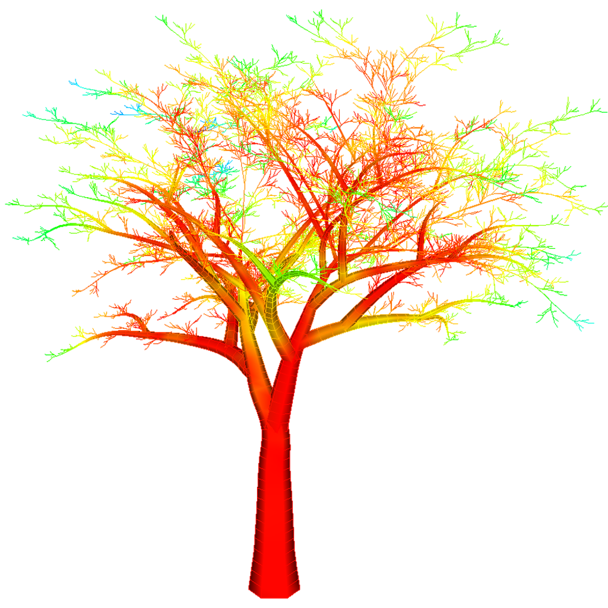}}
    & \frame{\parbox[][0.18\textwidth][c]{3em}{\centering $0.026$}}
    \\
    % \rotatebox{90}{\parbox{4em}{\centering \small HDD Value}}
    % & \parbox[][4em][t]{0.18\textwidth}{\centering $0.040$}
    % & \parbox[][4em][t]{0.18\textwidth}{\centering $0.033$} 
    % & \parbox[][4em][t]{0.18\textwidth}{\centering $0.026$}\\
    % \parbox[][2em][t]{0.05\textwidth}{\small HDD \\ Mean Value}
    % & \parbox[][2em][c]{0.18\textwidth}{\centering $0.040$}
    % & \parbox[][2em][c]{0.18\textwidth}{\centering $0.033$} 
    % & \parbox[][2em][c]{0.18\textwidth}{\centering $0.026$}\\
    \end{tabular}
    };
% 	\begin{tikzpicture}
            \begin{axis}[
                hide axis,
                scale only axis,
                height=0pt,
                width=0pt,
                shift={(6.85cm,4.4cm)},
                colormap={reversejet}{rgb255(0cm)=(128,0,0) rgb255(1cm)=(255,0,0)
            rgb255(3cm)=(255,255,0) rgb255(5cm)=(0,255,255) rgb255(7cm)=(0,0,255)
            rgb255(8cm)=(0,0,128)},
                colorbar horizontal,
                point meta min=0,
                point meta max=0.1,
                colorbar style={
                    width=9.1cm,
                    height = 0.2cm,
                    rotate=90,
                    xtick={0, 0.02, 0.03, 0.04, 0.1},
                    xticklabels={$0$, $0.02$, $0.03$, $0.04$, $0.1$},
                    xticklabel style= {font= \small, xshift=0.4cm, yshift=0.2cm},
                }
              ]
                \addplot [draw=none] coordinates {(0,0) (1,1)};
            \end{axis}
        \end{tikzpicture}
	\caption{\revision{Worst, median, and better case of the experiment with different degrees of rotation. For each case are presented the SG input sketch, the reconstructed 3D model, and the 3D model of the ground truth both with realistic texture (Ground Truth) and with its vertices having the color corresponding to the HDD value (GT HDD). At the bottom of the figure, the HDD Mean Value is shown.}}
	\label{fig:360_hdd}
\end{figure*}

\revision{To further test the TSN accuracy we provide an experiment. It consists of generating the 3D model of a tree that TSN has never seen and rotating it by $5$ degrees until the starting position is reached. For each rotation, a new SG sketch has to be rendered and given as input to the TSN. Finally, the HDD has to be calculated for each TSN prediction. As shown in Figure \ref{fig:360_hdd_graph}, for each degree of rotation the HDD value is approximately $0.03$, in line with the results in Table \ref{tab:hausdorff_distance}. }
\revision{Figure \ref{fig:360_hdd} presents the worst, median and better case of the experiment. Each column contains the SG sketch given as input to our TSN, the 3D model predicted by the TSN, the 3D model of the ground truth with the realistic textures, and the 3D model of the ground truth whose vertices have the color corresponding to their HDD value.}

\subsection{Qualitative analysis}\label{sec:qualitative}
In this section, we provide an empirical analysis of our TSN predicted parameters using HM sketches for which GT parameters were unknown. To do this, we manually extracted some easily-visible parameters for each tree \revision{species} based on the input sketches and compared them to TSN predictions. \revision{Table \ref{tab:params_symbols} describe the symbolism adopted in this section.}
\begin{table}[!ht]
    \centering
    \begin{tabular}{l c | l c}
    \toprule
        \revision{Parameter} & \revision{Symbol} & \revision{Parameter} & \revision{Symbol}\\
        \hline
        \revision{Gravity-bending Strength} & \revision{$GBS$} &
        \revision{Crown Base Height} & \revision{$CBH$} \\
        \revision{Number of Tree Forks} & \revision{$N_{TF}$} &
        \revision{Branch Distribution} & \revision{$BD$} \\
        \revision{Number of Branches} & \revision{$N_B$} &
        \revision{First Half Internodes Branching Angle} & \revision{$\phi_{FIB}$}\\
        \revision{Second Half Internodes Branching Angle} & \revision{$\phi_{SIB}$} &
        \revision{Number of Levels} & \revision{$N_L$} \\
        \revision{Number of Branch Whorls} & \revision{$N_{BW}$}\\
        \bottomrule
    \end{tabular}
    \caption{\revision{Summary of symbolism used in this section.}}
    \label{tab:params_symbols}
\end{table}

% \noindent
% \begin{table}[!ht]
% \begin{tabular}[t]{ll}
% \includegraphics[width=0.4\textwidth,valign=m]{samples/\imageres/handMade_with_tag_efficientnet/1_acer/acer_handMade.png}
% &
%     \resizebox{0.3\textwidth}{!}{%
%          \input{samples/\imageres/handMade_with_tag_efficientnet/1_acer/acer_handMade_params}}
% \end{tabular}
% \caption{Simple visual parameters of the maple tree. The corresponding table reports TSN predicted values. Indices of elements in array-structured parameters represent the branch level.}
% \label{tab:comp_maple}
% \end{table}

% \noindent
% \begin{table}[!ht]
% \begin{tabular}[t]{ll}
% \includegraphics[width=0.4\textwidth,valign=m]{samples/\imageres/handMade_with_tag_efficientnet/1_acer/acer.png}
% &
%     % \resizebox{0.25\textwidth}{!}{%
%          \input{samples/\imageres/handMade_with_tag_efficientnet/1_acer/acer_handMade_params}
%         %  }
% \end{tabular}
% \caption{Simple visual parameters of the maple tree. The corresponding table reports TSN predicted values. Indices of elements in array-structured parameters represent the branch level.}
% \label{tab:comp_maple}
% \end{table}

\begin{figure}[!ht]
    \centering
    \includegraphics[width=0.8\linewidth]{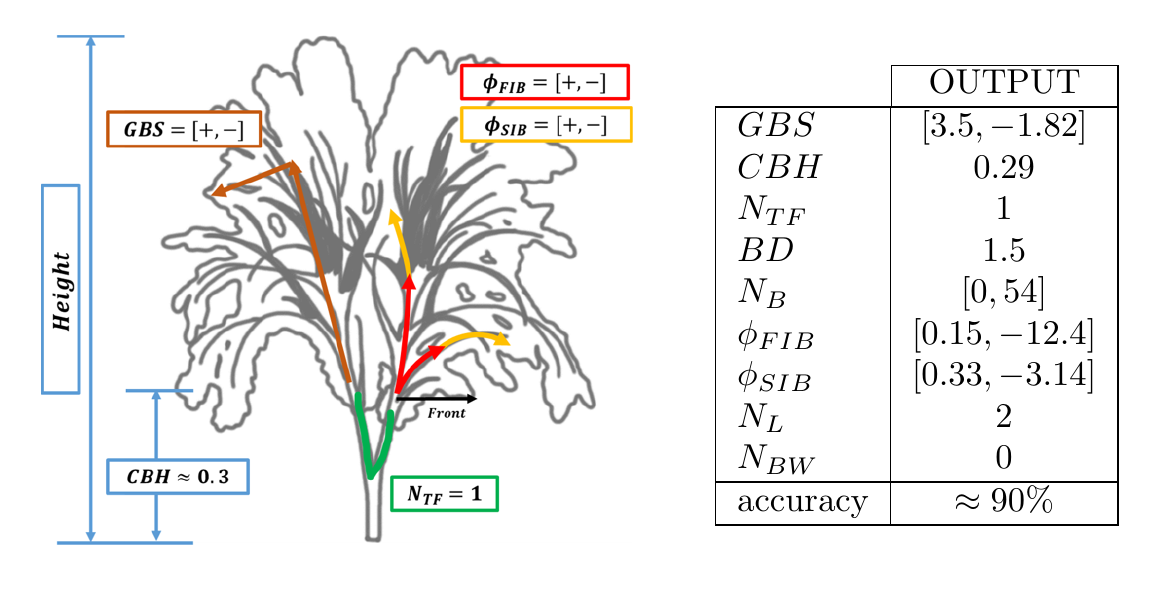}
    \caption{Simple visual parameters of the maple tree. The corresponding table reports TSN predicted values. Indices of elements in array-structured parameters represent the branch level.}
\label{tab:comp_maple}
\end{figure}

Figure~\ref{tab:comp_maple} shows the input HM maple tree sketch, and some of its simple visual parameters. This table shows some of the TSN's predicted parameters and performance accuracy. In particular, the first \revision{$GBS$ value}, highlighted in dark orange, shows that the trunk (the first branch level) is attracted upwards (positive sign) and the branches (second branch level) are attracted downwards (negative sign) with a different attraction coefficient, were correctly predicted by our TSN. Trunk height indicates that about $30\%$ of the maple tree height has no branches. The trunk in the sketch is also split into two parts, confirmed by the predicted \revision{$N_{TF}$} parameter, highlighted in green, which indicates one trunk subdivision. The \revision{$N_B$} parameter indicates the number of branches for each branch level. In the maple tree case, there are $0$ first level branches because the trunk is not considered as a branch, and $54$ second level branches. 
\revision{$\phi_{FIB}$} and \revision{$\phi_{SIB}$} trunk signs are positive, which means that the trunk and its tip are curved backward the front of the tree. Second level branches and their tips are curved in the opposite direction to the trunk as shown in the related sketch.
The remaining parameters are also correctly predicted by our TSN, notably, the \eg \revision{$N_L$} parameters indicate that the sketch is characterized by two branch levels.

% \noindent
% \begin{table}[ht!]
% \begin{tabular}[t]{ll}
% \includegraphics[width=0.4\textwidth,valign=m]{samples/\imageres/handMade_with_tag_efficientnet/2_palm/palm_handMade.png}
% &
%     \resizebox{0.3\textwidth}{!}{%
%          \input{samples/\imageres/handMade_with_tag_efficientnet/2_palm/palm_handMade_params}}
% \end{tabular}
% \caption{An example of a palm tree \revision{SG} sketch with predicted parameters. The correctness of the prediction is mainly demonstrated by the parameters \revision{$BD$} and \revision{$CBH$}, which indicate that the foliage is concentrated towards the top of the tree.}
% \label{tab:comp_palm}
% \end{table}

% \noindent
% \begin{table}[ht!]
% \begin{tabular}[t]{ll}
% \includegraphics[width=0.4\textwidth,valign=m]{samples/\imageres/handMade_with_tag_efficientnet/2_palm/palm.png}
% &
%     % \resizebox{0.3\textwidth}{!}{%
%          \input{samples/\imageres/handMade_with_tag_efficientnet/2_palm/palm_handMade_params}
%         %  }
% \end{tabular}
% \caption{An example of a palm tree \revision{SG} sketch with predicted parameters. The correctness of the prediction is mainly demonstrated by the parameters \revision{$BD$} and \revision{$CBH$}, which indicate that the foliage is concentrated towards the top of the tree.}
% \label{tab:comp_palm}
% \end{table}

\begin{figure}[!ht]
    \centering
    \includegraphics[width=0.8\linewidth]{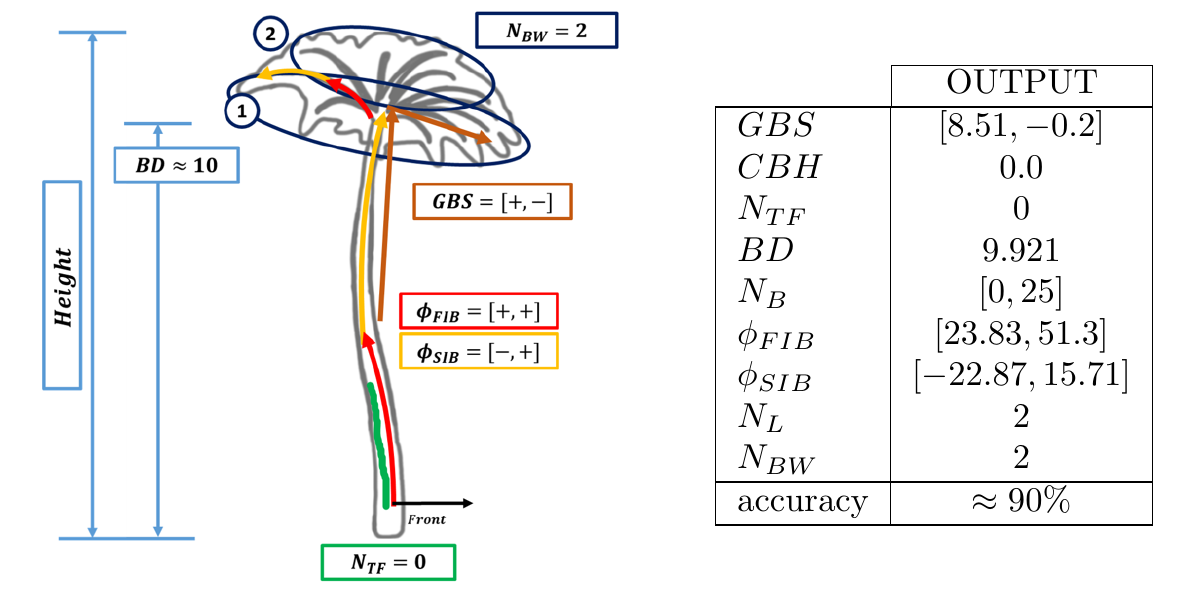}
\caption{An example of a palm tree \revision{SG} sketch with predicted parameters. The correctness of the prediction is mainly demonstrated by the parameters \revision{$BD$} and \revision{$CBH$}, which indicate that the foliage is concentrated towards the top of the tree.}
\label{tab:comp_palm}
\end{figure}

Figure~\ref{tab:comp_palm} is a visual comparison of some simple parameters and their corresponding TSN-predicted values. In this case \revision{$CBH$} indicates that only the $1\%$ of the palm tree has no branches, but the \revision{$BD$} parameter gathers all of the branches towards the top of the tree, and attenuates the \revision{$CBH$} effect. This aspect is also confirmed in the maple tree example~\ref{tab:comp_maple}, where these two parameters are balanced. In the sketch, there are no subdivisions of the trunk, indicated by the \revision{$N_{TF}$} parameter. Since \revision{$\phi_{FIB}$} and \revision{$\phi_{SIB}$} trunk signs are discordant, the trunk is S-shaped and curved backwards (positive sign for \revision{$\phi_{FIB}$}) and its tip (negative sign for \revision{$\phi_{SIB}$}) is curved forwards. The first-level branches follow the curvature of the trunk. Finally, this example shows the \revision{$N_{BW}$} parameter, which indicates how many rings the branches are distributed on around the trunk.

% \noindent
% \begin{table}[ht!]
% \begin{tabular}[t]{ll}
% \includegraphics[width=0.4\textwidth,valign=m]{samples/\imageres/handMade_with_tag_efficientnet/3_pine/pine_handMade.png}
% &
%     \resizebox{0.3\textwidth}{!}{%
%          \input{samples/\imageres/handMade_with_tag_efficientnet/3_pine/pine_handMade_params}}
% \end{tabular}
% \caption{This figure and the corresponding table show that the number of \revision{$N_{BW}$} is correctly predicted by our system, as are other parameters.}
% \label{tab:comp_pine}
% \end{table}

\begin{figure}[!ht]
    \centering
    \includegraphics[width=0.8\linewidth]{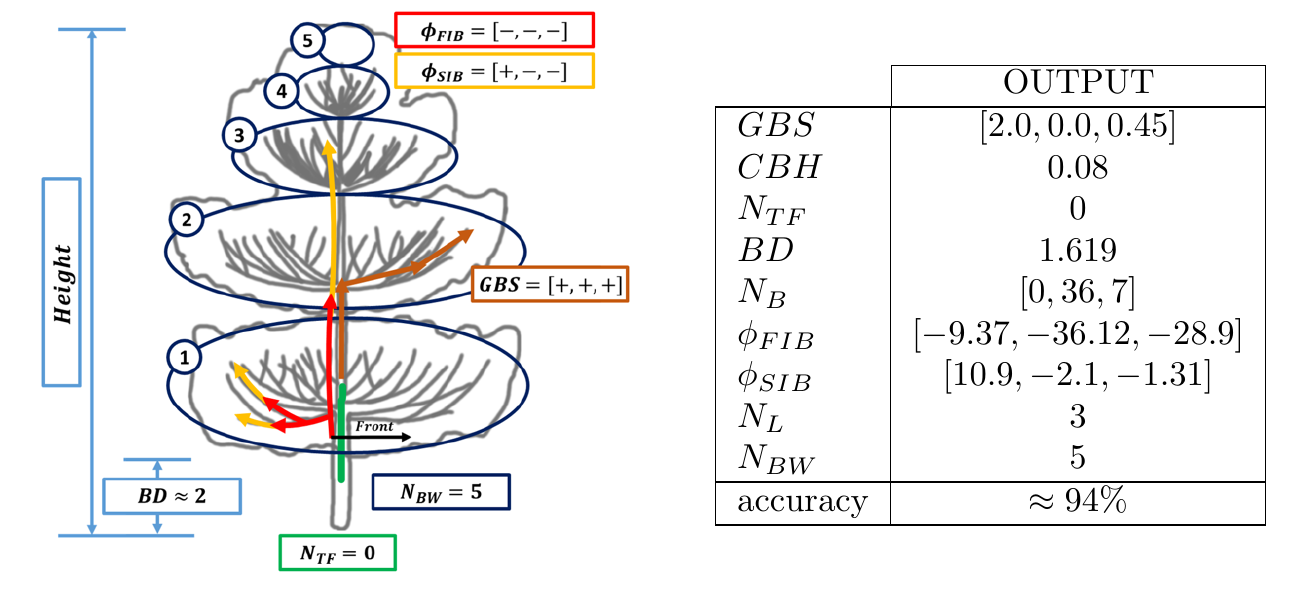}
\caption{This figure and the corresponding table show that the number of \revision{$N_{BW}$} is correctly predicted by our system, as are other parameters.}
\label{tab:comp_pine}
\end{figure}

The most evident parameter in Figure~\ref{tab:comp_pine} is \revision{$N_{BW}$} corresponding table, notably the sign and number of  \revision{$GBS$}, \revision{$\phi_{FIB}$}, and \revision{$\phi_{SIB}$} parameters. The red arrows show the sign of the \revision{$\phi_{FIB}$} parameter, and indicates that all branch levels are curved forwards. The tip of the last two branches levels follows the same direction of curvature as their base, and the trunk tip is curved backwards with respect to its base, as can be seen from the \revision{$\phi_{SIB}$} parameter. In this case, the branches of all levels are attracted upward as correctly predicted by our TSN.

% \noindent
% \begin{table}[ht!]
% \begin{tabular}[t]{ll}
% \includegraphics[width=0.4\textwidth,valign=m]{samples/\imageres/handMade_with_tag_efficientnet/4_cherry/cherry_handMade.png}
% &
%     \resizebox{0.3\textwidth}{!}{%
%          \input{samples/\imageres/handMade_with_tag_efficientnet/4_cherry/cherry_handMade_params}}
% \end{tabular}
% \caption{An example of a cherry tree. In this case, the \revision{$N_{TF}$} parameter is $1$, indicating that the main trunk is divided into two sub-trunks.}
% \label{tab:comp_cherry}
% \end{table}

\begin{figure}[!ht]
    \centering
    \includegraphics[width=0.8\linewidth]{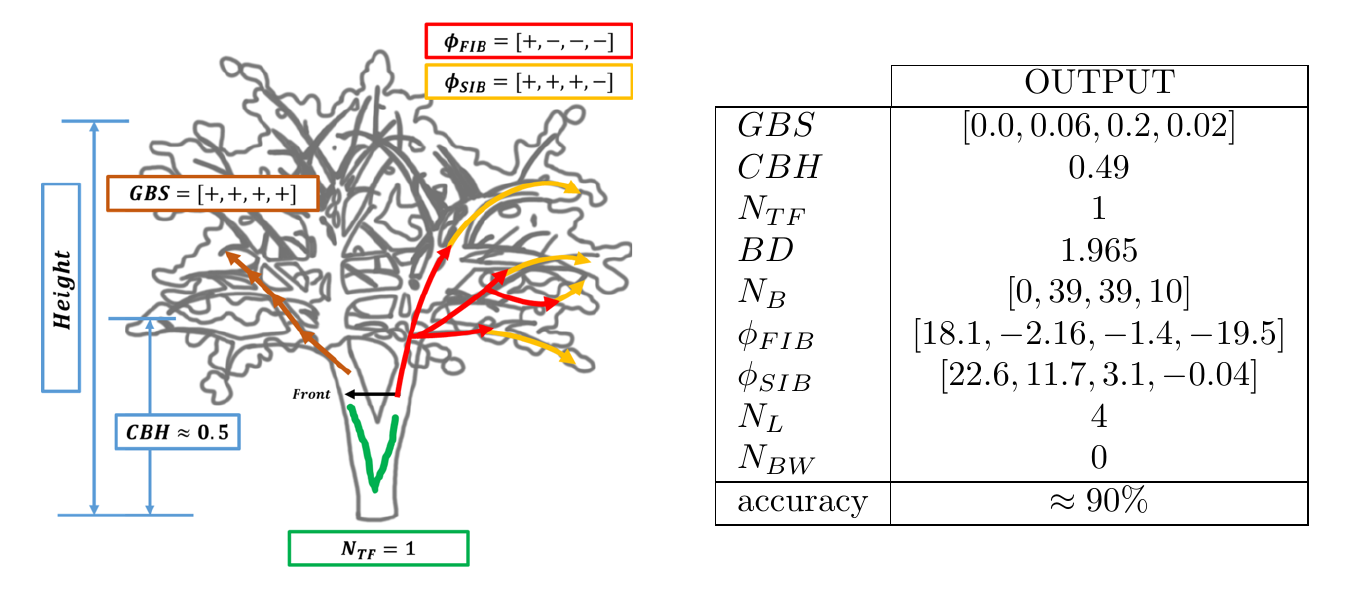}
\caption{An example of a cherry tree. In this case, the \revision{$N_{TF}$} parameter is $1$, indicating that the main trunk is divided into two sub-trunks.}
\label{tab:comp_cherry}
\end{figure}

The cherry tree shown in Figure~\ref{tab:comp_cherry} is characterized by four branch levels (\revision{$N_L = 4$}). The coefficients of \revision{$\phi_{FIB}$} and \revision{$\phi_{SIB}$} indicate the steepness of the curve. In this case, the first is curved backwards, and the other branch levels are curved in the opposite direction. The  curvature of the tip of the first three branch levels follows the curvature of the base of the trunk, and the tips of the last branch levels follow their base curvature. In this case, the trunk is split into two, and the branch attraction points upward for all levels. Finally, the \revision{$CBH$} indicates that about the $50\%$ of the cherry tree has no branches, which is consistent with the \revision{$BD$} parameter.

% \noindent
% \begin{table}[ht!]
% \begin{tabular}[t]{ll}
% \includegraphics[width=0.4\textwidth,valign=m]{samples/\imageres/handMade_with_tag_efficientnet/5_bonsai/bonsai_handMade.png}
% &
%     \resizebox{0.3\textwidth}{!}{%
%          \input{samples/\imageres/handMade_with_tag_efficientnet/5_bonsai/bonsai_handMade_params}}
% \end{tabular}
% \caption{An example of the bonsai tree \revision{SG} sketch. The S-shape of the trunk and first-level branches is indicated by the discordant signs of \revision{$\phi_{FIB}$} and \revision{$\phi_{SIB}$} parameters.}
% \label{tab:comp_bonsai}
% \end{table}

\begin{figure}[!ht]
    \centering
    \includegraphics[width=0.8\linewidth]{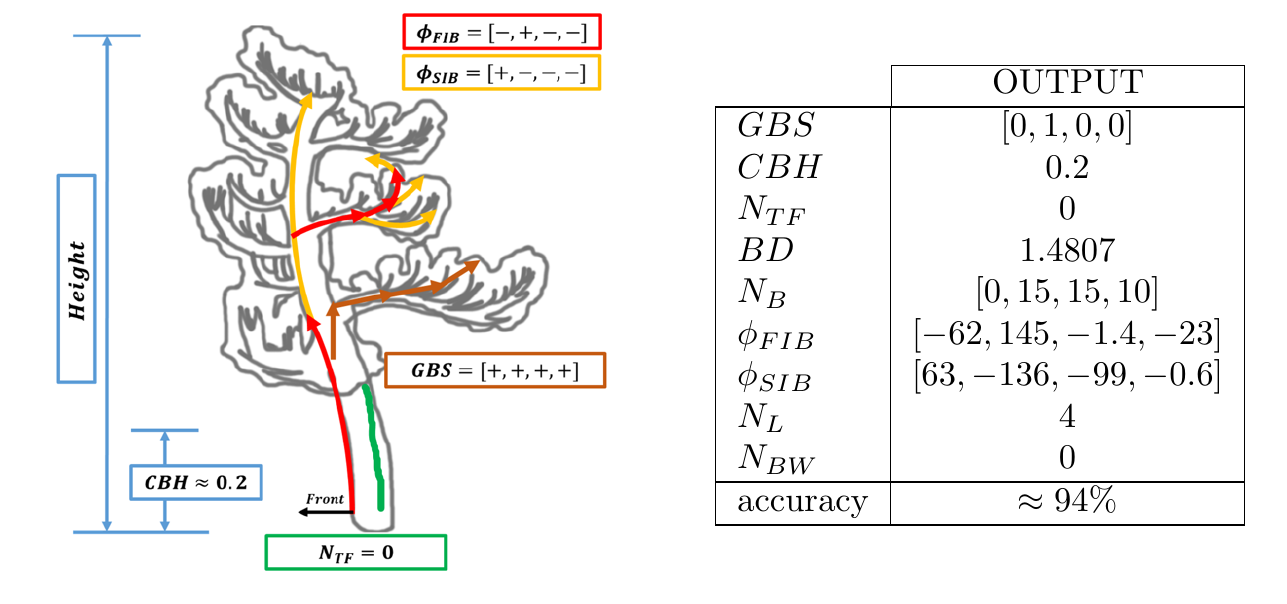}
\caption{An example of the bonsai tree \revision{SG} sketch. The S-shape of the trunk and first-level branches is indicated by the discordant signs of \revision{$\phi_{FIB}$} and \revision{$\phi_{SIB}$} parameters.}
\label{tab:comp_bonsai}
\end{figure}

The bonsai tree shown in Figure~\ref{tab:comp_bonsai} also have four branch levels (\revision{$N_L = 4$}). The trunk of the bonsai has no splits, as indicated by the \revision{$N_{TF}$} parameter, and is also S-shaped, as showed by the discordant signs of the \revision{$\phi_{FIB}$} and \revision{$\phi_{SIB}$} parameters. The first-level branches are also S-shaped but are specular in their orientation to the trunk. Indeed, the \revision{$\phi_{FIB}$} and \revision{$\phi_{SIB}$} of the other branches levels are all negative, so the bases and the tips of these branches are curved in the forward direction. In addition, the $20\%$ of the cherry tree has no branches and all the branches levels are attracted upward, as confirmed by the \revision{$CBH$} and \revision{$GBS$} parameters respectively.
\section{Conclusion and Future Work}\label{sec:conclusions}
Our proposed approach can predict parameters for 3D tree mesh generation using a DNN method. The main goal is to automate procedural modeling by introducing a broker system between the modeler and the modeling software used to build 3D trees. The core of our broker consists of a DNN based on convolutions, that we call TreeSketchNet, which is trained using supervised methods to learn the mapping between well-known Weber-Penn tree parameters and 2D sketches of trees. As a large amount of data is needed for training and validation datasets, we developed a dedicated RT Blender add-on. This add-on makes it possible to automate the generation of realistic \revision{SG} sketches starting from 3D tree meshes that are generated a priori using a set of fixed and unfixed randomly-controlled parameters. This approach is implemented to overcome the problem of generating an expensive HM dataset of drawings. For our experimental purpose, we consider $5$ tree \revision{species} (maple, pine, bonsai, palm, and cherry) and create corresponding sketches from front, back, left, and right camera angles. 
We assessed our system and the obtained results using a controlled experiment. Specifically, we asked participants to provide HM sketches based on reference \revision{SG} examples shown for two minutes, and used them to test the system. The results were promising, and we believe may be a starting point for future research. Our experiment also highlighted the high level of generalization, and validated the accuracy of our approach. Furthermore, we experimented with several other core nets to identify the most suitable and performant option. Notably, we tested AlexNet, which is widely-used in computer vision tasks, and won the ILSVRC competition~\cite{ILSVRC15}, becoming established as the state-of-the-art in the deep learning field. AlexNet is the only DNN that has been used in a similar sketch-to-mesh parameters approach~\cite{huang2017}. However, it did not perform as well as Efficientnet-B7, and was therefore discarded. Finally, we provide a qualitative analysis of our results, specifically, a visual comparison of the predicted parameters with their corresponding input sketches. Our results show that our procedural modelling-based approach is better compared to image-to-mesh or voxel, with respect to rough and smooth surfaces, artifacts, holes, and deformed or unnecessary polygons. 
As our results are promising, we plan to continue to investigate the sketch-to-3D mesh approach based on procedural modelling and deep learning. \revision{Indeed, our approach is a baseline for the generation of a 3D tree model from a hand-made sketch. We proved that by giving the user some guidelines for drawing the sketch, the TSN obtain results consistent with the input sketch. Obviously, as affirmed also by \cite{unlu2022interactive}, the approach has limitations once the user draws a tree very different from the sketches in the training set. We foreseen to explore the use of different style of sketches to make our approach more generalizable.} 
In future work, we could expand the number of tree \revision{species} and/ or generate new 3D meshes by adapting our approach to other contexts where procedural modelling can be used, such as vases, chairs, buildings, furniture, \etc~\cite{huang2017}. 
Also in future work, we would like to define a method for texture generation, which would avoid the use of the tree \revision{species} identification algorithm~\ref{alg:identify} that is used to select the most suitable texture from a predefined set.
A possible method could be to consider colored input sketches, and extract the texture or color from them. \revision{Furthermore, this approach could be used as a starting point for implementing an application for generating 3D trees or 3D environments in real time. Another future work could use our approach to generate a 3D model from a sketch and use it to produce a video in cartoon style, making a mesh-to-image synthesis.}

\subsection{Discussion and Limitations}\label{sub:limitations}
\begin{figure}[ht!]
\centering
 \begin{tabular}{c c c} 
    Input & Reconstructed & Ground Truth\\
	\frame{\includegraphics[width=0.15\textwidth]{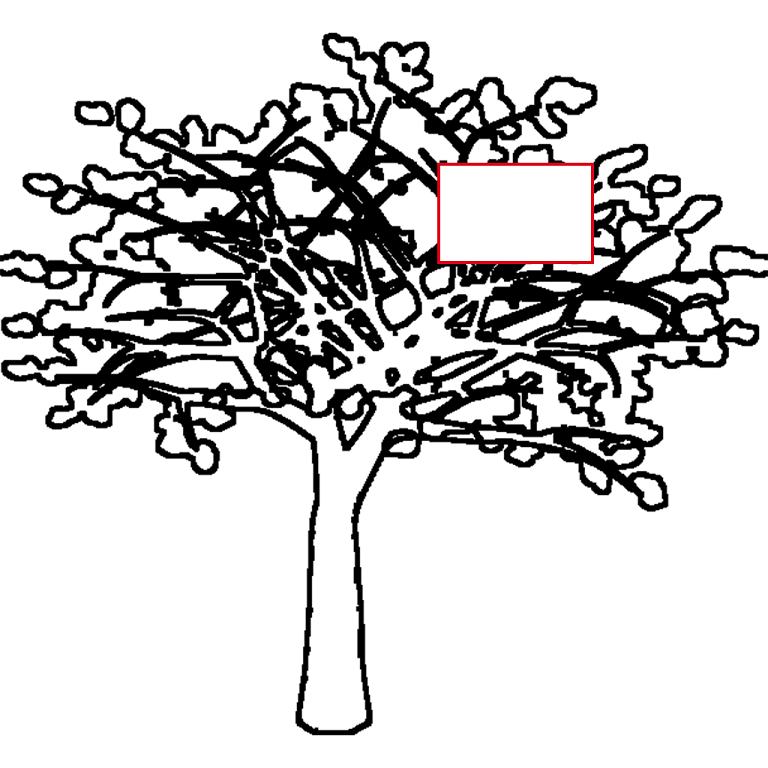}}
	& \frame{\includegraphics[width=0.15\textwidth]{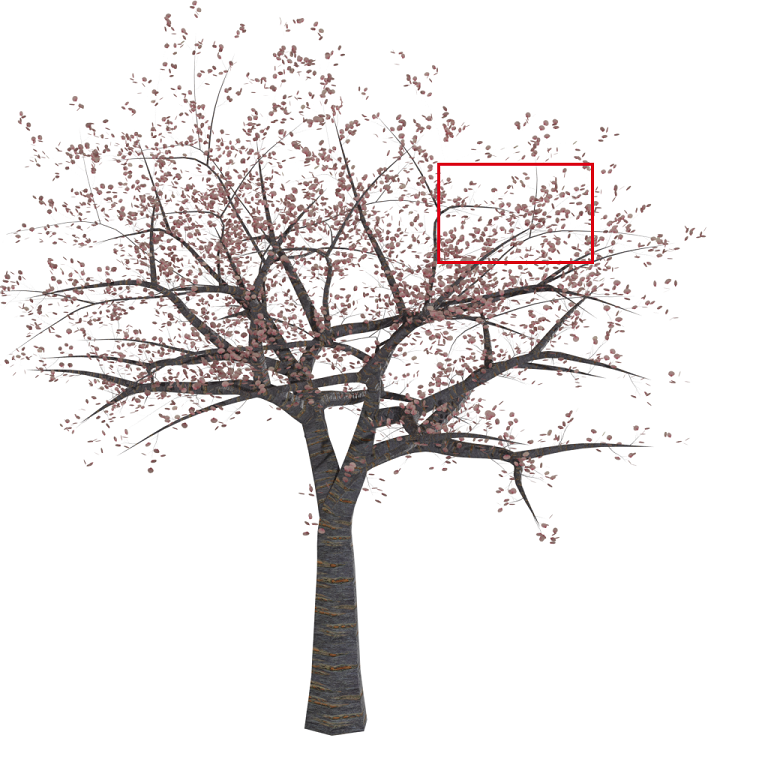}} 
	& \frame{\includegraphics[width=0.15\textwidth]{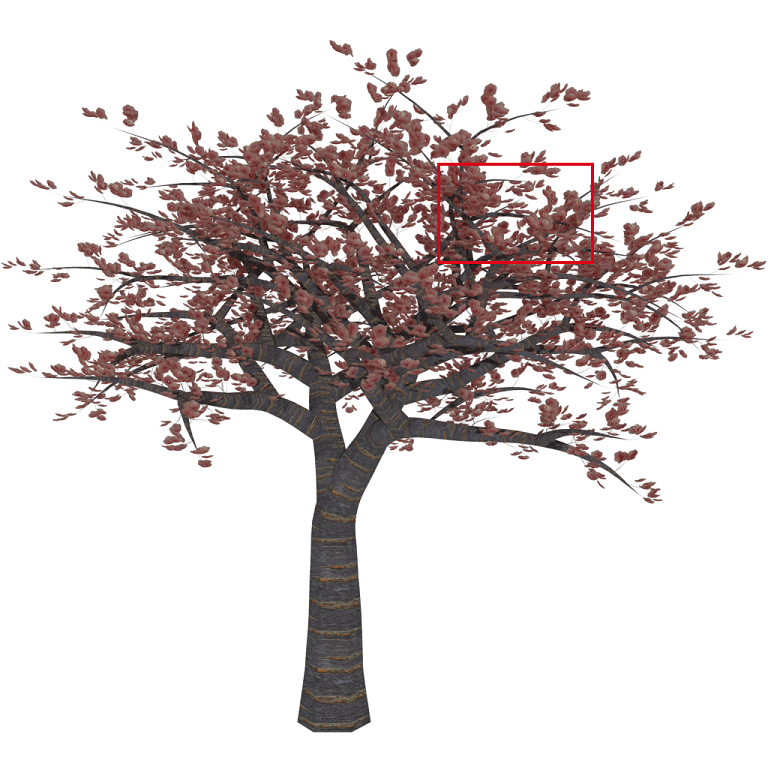}}
	\\
	\frame{\includegraphics[width=0.15\textwidth]{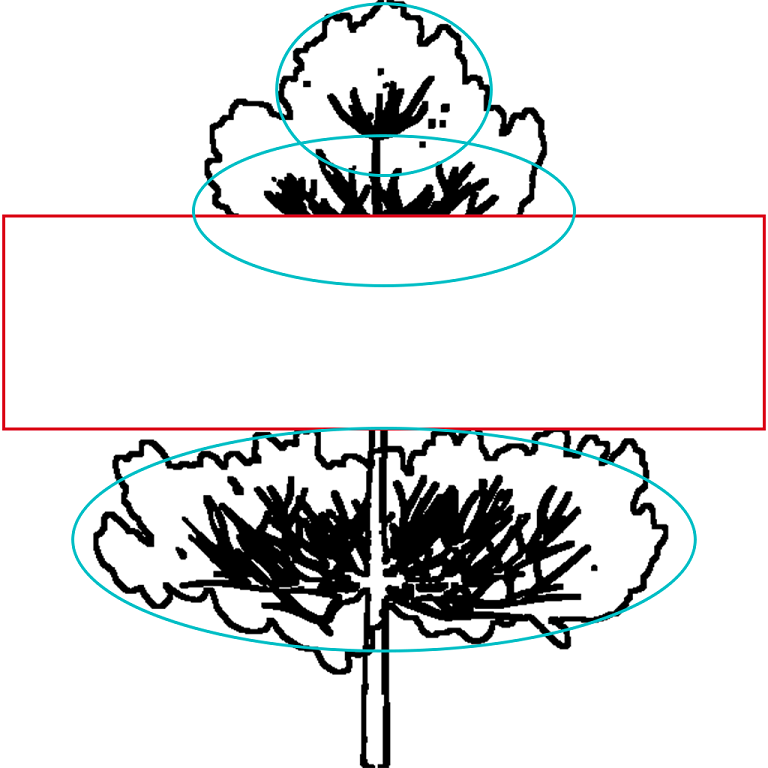}}
	& \frame{\includegraphics[width=0.15\textwidth]{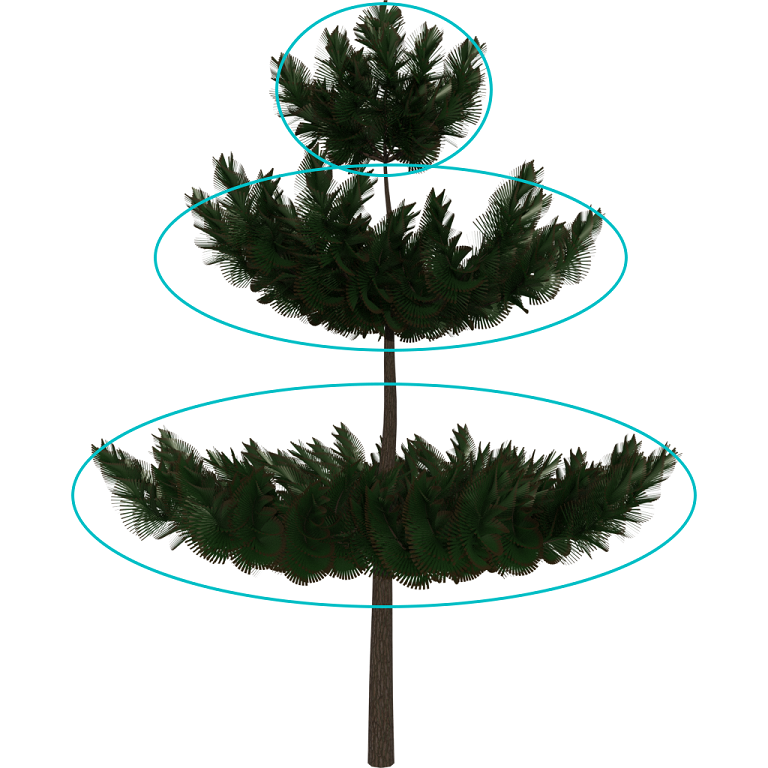}} 
	& \frame{\includegraphics[width=0.15\textwidth]{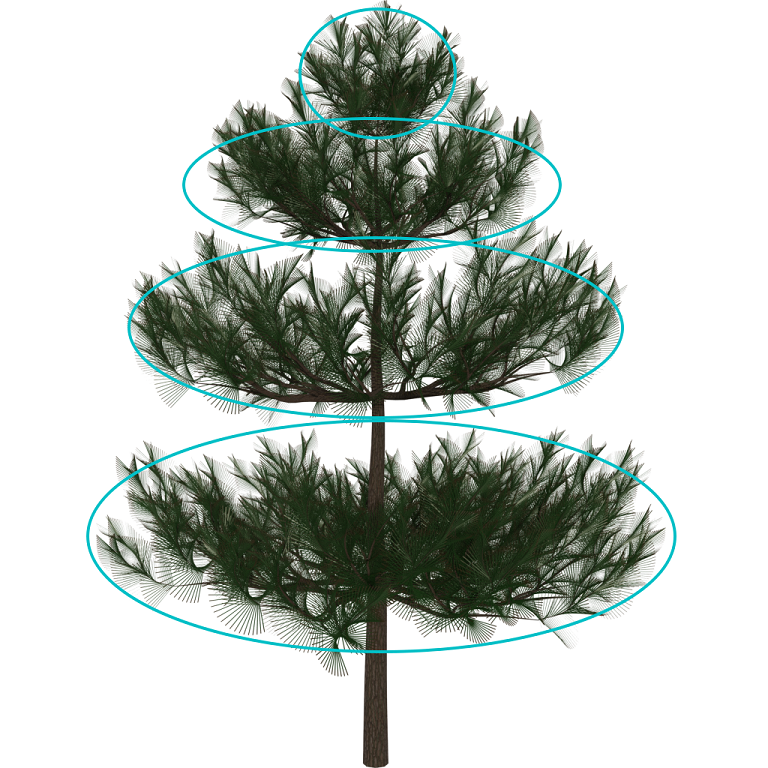}}
	\\
	\end{tabular}
	\caption{Results of TSN given sketches with missing regions.}
	\label{fig:cropped_images}
\end{figure}

\begin{figure}[ht!]
\centering
 \begin{tabular}{c c c c} 
    Input HM & 3D Mesh & Input HM & 3D Mesh\\
	\frame{\includegraphics[width=0.15\textwidth]{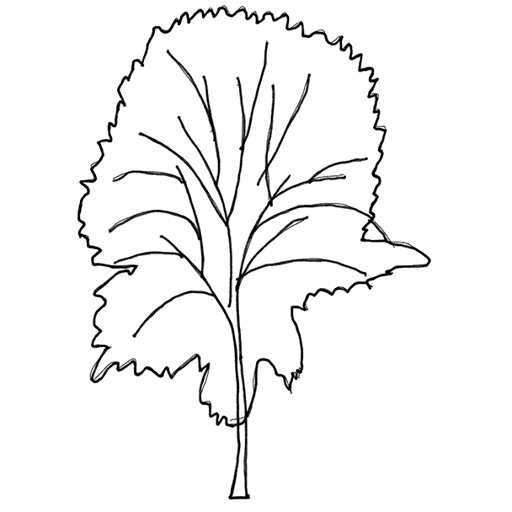}}
	& \frame{\includegraphics[width=0.15\textwidth]{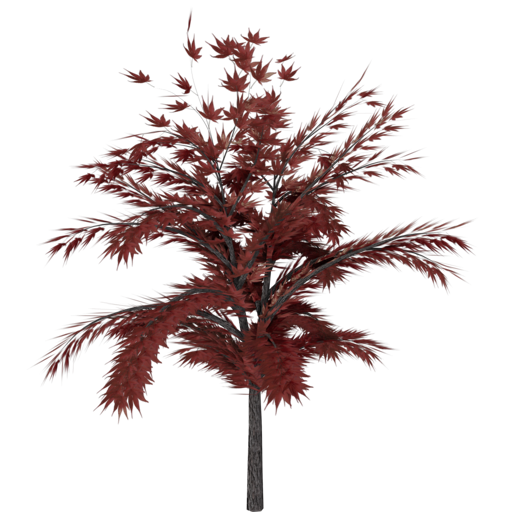}} &
	\frame{\includegraphics[width=0.15\textwidth]{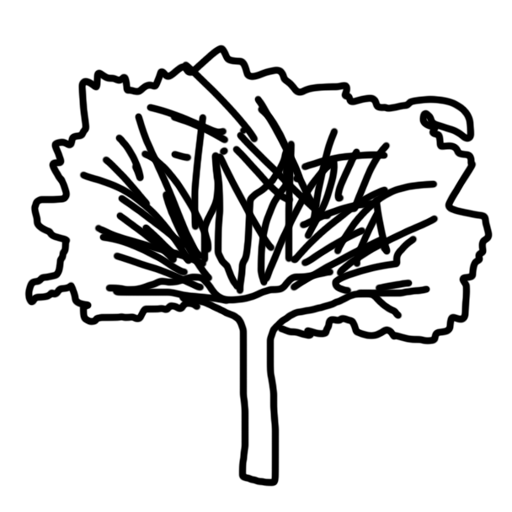}}
	& \frame{\includegraphics[width=0.15\textwidth]{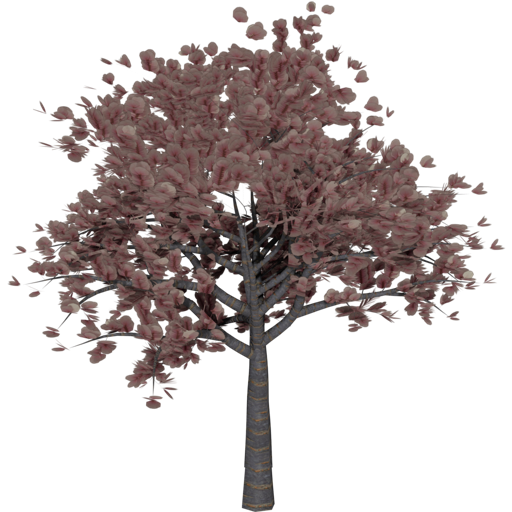}}\\
	\end{tabular}
	\caption{Examples of outliers in the controlled experiment.}
	\label{fig:limitations}
\end{figure}

\revision{There are several reasons why our approach, which automates procedural modeling by predicting parameters from sketches, proves to be the best choice for generating complex 3D models, such as trees.} \revision{One of the reasons is that} approaches that directly predict mesh using images as DNN input often present qualitative problems~\cite{NIPS2019_8340, xie2019pix2vox, gkioxari2019mesh}, such as rough and smooth surfaces, artifacts, holes, and deformed and unnecessary polygons, \revision{especially for thin and layered structure like those of tree trunk and branches}. Furthermore, the 3D meshes that are generated from direct methods often poorly resemble the input RGB images or sketches. Procedural modeling approaches for the generation of 3D meshes can overcome these problems~\cite{garmentdesign_Wang_SA18,2014Smelik,liu2021deep}. Another advantage is that the 3D mesh is always correct, as our robust RT Blender add-on correctly interprets the parameters and uses them, avoiding artifacts and error generation. The latter is demonstrated by the predictive accuracy of the parameters of our TSN, even in difficult conditions, as reported in Section~\ref{sub:user_study}. Although procedural modelling has only recently been explored in the field of deep learning and artificial intelligence ~\cite{liu2021deep, park2019, yumer2015}, there are, as yet, few specific approaches that examine prediction parameters for 3D non-linear content generation, either plants, in general, or trees~\cite{LIU2021101115,li2021}, in particular. \revision{We compared} our baseline (see Section~\ref{sub:backbone}) with better-known core nets, to assess the effectiveness of our work. \revision{To further assess the performance of TSN in term of accuracy we provided a quantitative analysis using the HDD metric.} We also evaluated the coherence of the predicted parameters with sketches provided as TSN input, through a visual qualitative analysis (see Section~\ref{sec:qualitative}) of some parameters that are easier to understand and visually identify. In addition, we compared the 3D tree meshes generated from predicted parameters with ground-truth meshes using the Hausdorff distance, as reported in Section \ref{sec:h_dist}.
To test the robustness of our method, we conducted some extra experiments of challenging cases. In detail, we tested our TSN with sketch images containing broken segments. Figure~\ref{fig:cropped_images} shows two examples of tree sketches with deleted parts. In the first row of Figure~\ref{fig:cropped_images} it can be seen that the reconstructed tree has less branches in the cropped region of the sketch than the ground-truth. The row of Figure~\ref{fig:cropped_images} shows the sketch of a pine with a missing ring. For this reason, TSN predicts a pine with one less ring than the ground-truth.
Figure~\ref{fig:limitations} reports two outliers caused by the poor TSN accuracy concerning the provided input sketches drawn by the experiment participants (see Section~\ref{sub:user_study}). The first \revision{two images} of Figure~\ref{fig:limitations} \revision{represent} a HM sketch of a maple tree with the relative 3D model. The sketch represents a tree with few branches drawn with a very thin line. As a result, the TSN predicts a maple tree with few branches, placing leaves on them. For this reason, the crown of the output tree is barer than sketch one, while the predicted branches are similar to the sketch ones.
Another interesting behavior is shown in the \revision{last two images} of Figure~\ref{fig:limitations}. Here, the tree branches are not well drawn, so the TSN tries to predict the tree shape based on the information provided by the trunk and the crown. The result is a 3D mesh not very similar to the input sketch but consistent with the tree \revision{species} and the main visible elements.

\begin{figure*}[ht!]
\centering
 \begin{tabular}{c c c c c c} 
    \rotatebox{90}{\parbox{6em}{\centering \small HM Input}}
    & \frame{\includegraphics[width=0.15\textwidth]{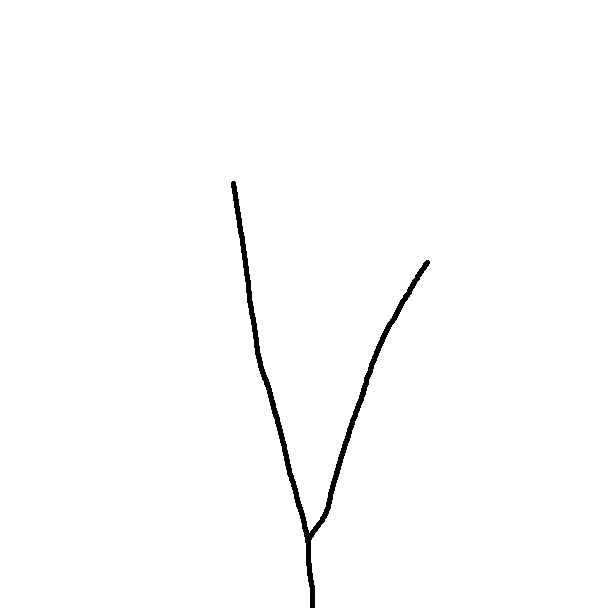}}
    & \frame{\includegraphics[width=0.15\textwidth]{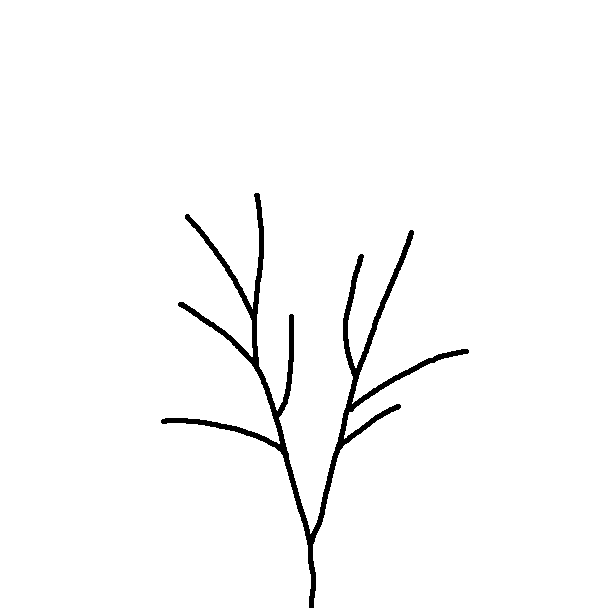}}
    & \frame{\includegraphics[width=0.15\textwidth]{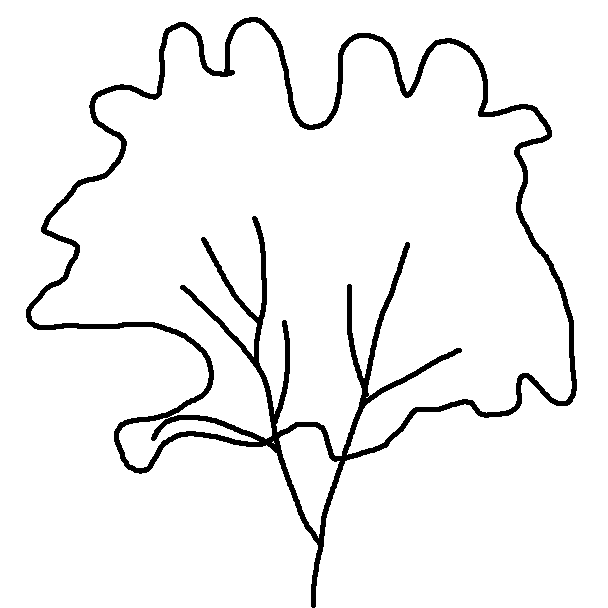}}
    & \frame{\includegraphics[width=0.15\textwidth]{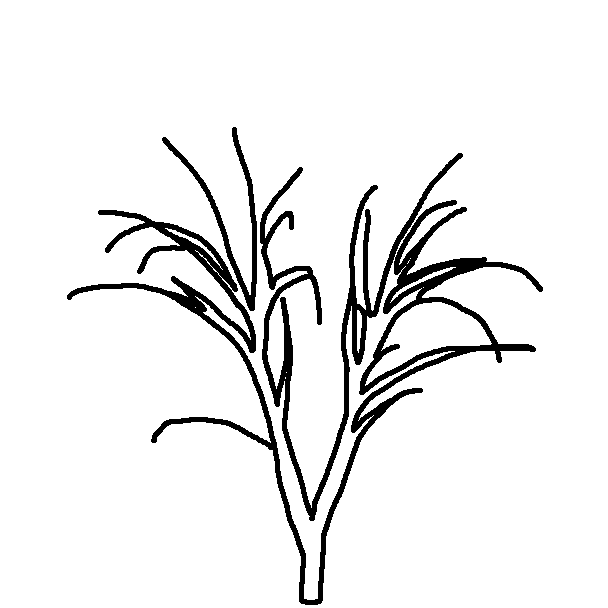}}
    & \frame{\includegraphics[width=0.15\textwidth]{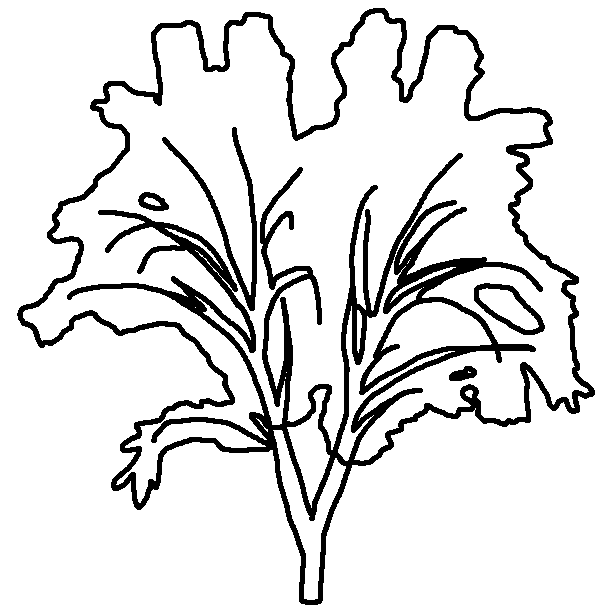}}
    
    \\ 
    \rotatebox{90}{\parbox{6em}{\centering \small 3D Mesh}}
    & \frame{\includegraphics[width=0.15\textwidth]{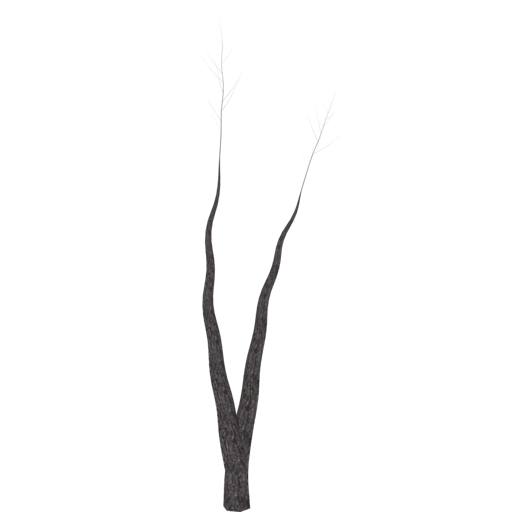}}
    & \frame{\includegraphics[width=0.15\textwidth]{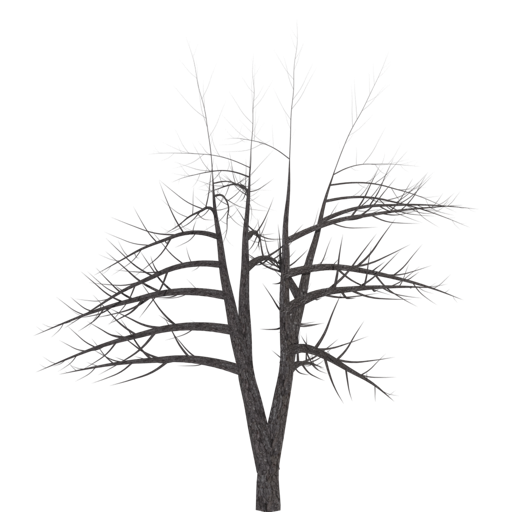}}
    & \frame{\includegraphics[width=0.15\textwidth]{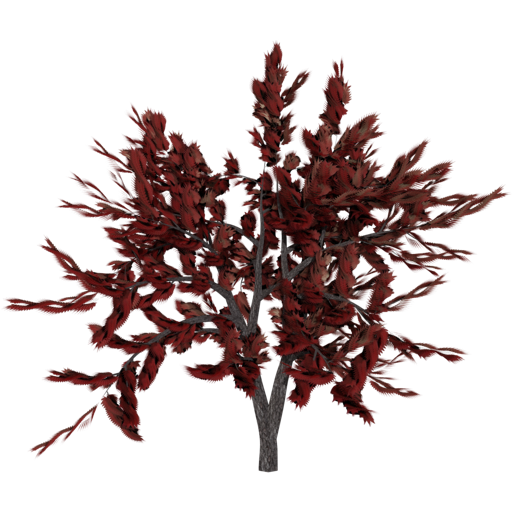}}
    & \frame{\includegraphics[width=0.15\textwidth]{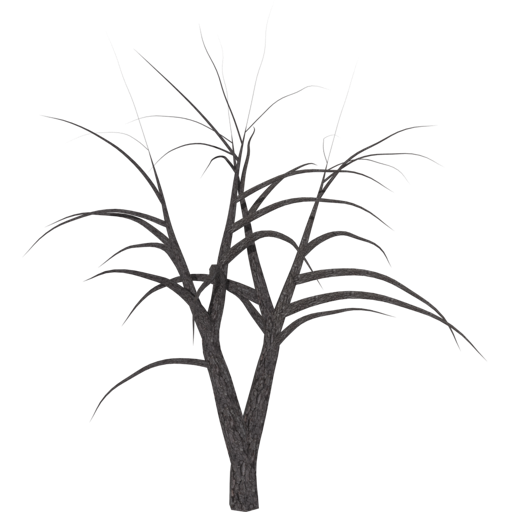}}
    & \frame{\includegraphics[width=0.15\textwidth]{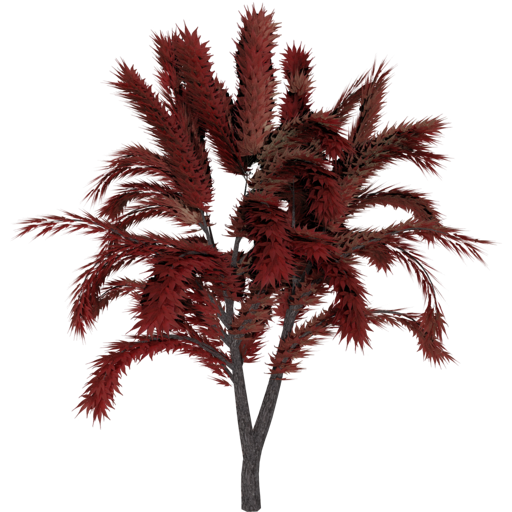}}
    \\
	\end{tabular}
	\caption{Experiment with HM sketches having different LOC.}
	\label{fig:LOD}
\end{figure*}

\revision{To get more visible indication on the details of sketches that the TSN considers mandatory, we propose another experiment. This consists of testing the TSN robustness in relation to HM sketches with various Levels Of Completeness (LOC). To further test our TSN, we decided to draw five maples with different LOCs, because with this species the TSN returns less accurate results. Figure~\ref{fig:LOD} shows the results of this experiment. Comparing the first two columns, TSN has more difficulties predicting the second-level branches if the crown is not present in the sketch and if the trunk and the branches have no thickness (second column). This is because the training set contains only sketches of crown trees whose branches are represented with different thicknesses in relation to their level. For this reason, when the sketch in the second column is given as input to TSN, this does not understand at what level the branches belong. On the other hand, the sketches in the third and fourth columns have more information than the sketch in the second column. In particular, the sketch in the third column contains also the crown information, and the sketch in the fourth column has information about the thickness of the trunk and the branches. As a result, the TSN generates trees similar to the input sketch. From all of this, it follows that TSN is able to obtain better results with sketches having the same drawing style as the sketch in the last column. Observing the results in Figure~\ref{fig:LOD} it can be noticed that if the sketch contains a branch much more bent than the others, the TSN tends to even the curvature of the other branches.}

 \revision{Through the information obtained from the previous experiments it is clear that it is necessary to provide participants with some guidelines for drawing HM sketches.} Although our work provides a pipeline for the generation of 3D trees based on well-defined parameters, one of the limitations is the small number of tree \revision{species} considered. However, this limitation could be easily overcome by considering more \revision{species} in future scenarios. \revision{The proposed approach is based on Weber-Penn method because is stable and was implemented as a Blender add-on. Despite this, it could be adapted to other newer procedural modelling methods \cite{stava2014inverse}, thank to the structure of the last part of TSN. In fact, it would be enough to sort each parameter of the new methodology into the branch of the last layer of the network corresponding to the relative range of values and then retrain the network using the previously learned weights as starting point for the new training.}

\section{Online Resources}\label{resources}
%We shared our the RT Blender plugin, DNN source code, and the trained weights through a GitHub repository:~\url{https://github.com/Unibas3D/TreeSketchNet}. Furthermore it is possible to ask the authors to share the training dataset by sending an email to them. Finally we attached an illustrative video in the GitHub page. 
We have shared our the RT Blender plugin, DNN source code, and the trained weights through a GitHub repository:~\url{https://github.com/Unibas3D/TreeSketchNet}. Furthermore, the authors are happy to share the training dataset upon request by sending an email to them. Finally, we have included an illustrative video into the GitHub page.

%\appendix
%\input{sections/appendix}

%Bibliography
\bibliographystyle{unsrt}  
\bibliography{references}

\end{document}